\documentclass[10pt]{article}

\newif\ifpreprint
\preprinttrue
\usepackage[preprint]{tmlr}

\usepackage[utf8]{inputenc}
\usepackage{microtype}
\usepackage{graphicx}
\usepackage{subcaption}
\usepackage{xcolor}
\usepackage{hyperref}
\usepackage{url}
\hypersetup{colorlinks=true, linkcolor=blue!50!black, citecolor=blue!50!black, urlcolor=blue!50!black}

\usepackage{amsmath}
\usepackage{amssymb}
\usepackage{booktabs}
\usepackage{array}
\usepackage[inline]{enumitem}
\usepackage{listings}
\usepackage[most]{tcolorbox}
\usepackage{placeins}
\usepackage{orcidlink}

\newcommand{\pos}[1]{\textcolor{red!60!black}{#1}}   %
\newcommand{\negative}[1]{\textcolor{green!50!black}{#1}} %

\title{Benchmarking and Adapting LLMs for English--Swedish Medical Question Answering}

\author{
    \hspace{-8pt} \name Seyed Alireza Molavi\textsuperscript{1,*}\,\orcidlink{0009-0000-7675-9783} \email seyed.alireza.molavi@hh.se \\
    \name Zhan Su\textsuperscript{1}\,\orcidlink{0000-0001-5189-9165} \email zhan.su@hh.se \\
    \name Stefan Byttner\textsuperscript{1} \email stefan.byttner@hh.se \\
    \name Yan Hu\textsuperscript{2}\,\orcidlink{0000-0003-2974-4073} \email huyan@cuhk.edu.cn \\
    \name Peyman Sheikholharam Mashhadi\textsuperscript{1}\,\orcidlink{0000-0002-0051-0954} \email peyman.mashhadi@hh.se \\
    \name Prayag Tiwari\textsuperscript{1,*}\,\orcidlink{0000-0002-2851-4260} \email prayag.tiwari@hh.se \\
    \addr \textsuperscript{1}Halmstad University, Halmstad, Sweden \\
    \addr \textsuperscript{2}The Chinese University of Hong Kong, Shenzhen, China \\
    \textnormal{\footnotesize \textsuperscript{*}Corresponding authors}
}

\newcommand{\codeurl}{\url{https://github.com/sar-molavi/SweMedMCQ-bench}}
\newcommand{\modelurl}{\url{https://huggingface.co/Anonymiserad}}

\def\month{MM}
\def\year{YYYY}
\def\openreview{\url{https://openreview.net/forum?id=XXXX}}

\begin{document} 
\maketitle

\begin{abstract}
Large language models (LLMs) perform well on English medical exams, but an English exam score does not tell us how well they answer the same questions in Swedish, the language of Swedish health care. We measure how accuracy changes between these languages and investigate how to improve it in Swedish. First, we build a paired English--Swedish benchmark from ten medical multiple-choice datasets, so each model answers the same questions in both languages. We check the translations through a conservative process grounded in Swedish medical sources and judgments from multiple LLMs. Smaller benchmark models lose up to 12.4 percentage points in Swedish when answering directly. When the models generate reasoning before answering, none loses more than 2.7 points. We then investigate whether existing English medical data can improve Swedish accuracy in compact LLMs that can run on local hardware. We translate the training questions and train two small LLMs on correct answers alone or on worked solutions generated by larger open LLMs. Including Swedish questions with these worked solutions brings English and Swedish accuracy close together while improving both. The best 1.2-billion-parameter model rises from 23.5\% to 52.0\% in Swedish. Across three training runs, its accuracy is comparable to that of a 20-billion-parameter open LLM answering in its fast mode, while requiring one-sixth as much memory for its weights. The gains also extend to two exam sets originally written in Swedish. We release the benchmark, its edit history, and the code\footnote{Code and data: \codeurl. Models: \modelurl.}.
\end{abstract}

\section{Introduction}
\label{sec:intro}

Large language models (LLMs) pass English medical licensing exams~\citep{singhal2023large,nori2023capabilities}, but less is known about how much of this accuracy remains when the same questions are asked in Swedish. This matters for Swedish health care, which works in Swedish.
Swedish is the principal language of Sweden, and patient records must be written in Swedish\footnote{\raggedright Language Act (SFS 2009:600), \S\,4, \url{https://www.riksdagen.se/sv/dokument-och-lagar/dokument/svensk-forfattningssamling/spraklag-2009600_sfs-2009-600/}; Patient Data Act (SFS 2008:355), ch.~3, \S\,13, \url{https://www.riksdagen.se/sv/dokument-och-lagar/dokument/svensk-forfattningssamling/patientdatalag-2008355_sfs-2008-355/}.}, so a model that supports clinicians must read and write Swedish medical text, and Swedish clinical text is sensitive and hard to share~\citep{vakili2025sweclineval}.
Health data fall under health-care secrecy and are a special category of personal data under the GDPR\footnote{\raggedright Public Access to Information and Secrecy Act (SFS 2009:400), ch.~25, \S\,1, \url{https://www.riksdagen.se/sv/dokument-och-lagar/dokument/svensk-forfattningssamling/offentlighets-och-sekretesslag-2009400_sfs-2009-400/}; Regulation (EU) 2016/679 (GDPR), Art.~9, \url{https://eur-lex.europa.eu/eli/reg/2016/679/oj}.}, and in many countries hospitals do not allow clinical data to be sent to external LLM services; running open models on the hospital's own hardware is the alternative~\citep{wiest2024privacy}. Open models of 27B parameters or more, such as Gemma-4-31B~\citep{gemmateam2026gemma4}, need 54--72 GB for their weights alone, while a model of 1--2B parameters needs 2.3--3.3 GB and fits on a laptop GPU.

\begin{figure}[tbp]
    \centering
    \includegraphics[width=\textwidth]{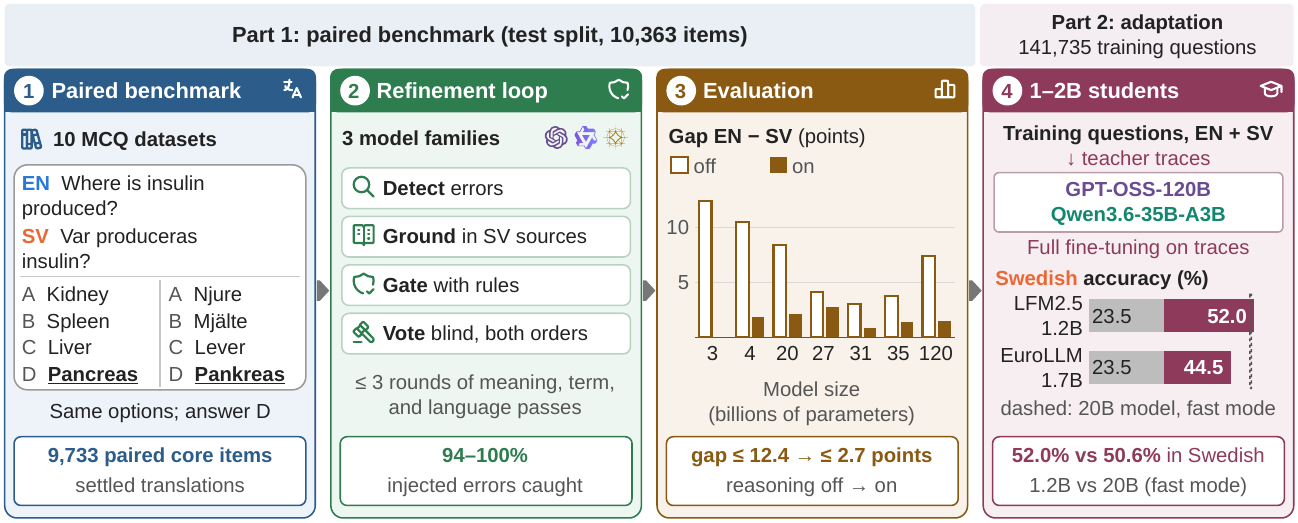}
    \caption{Overview of the study. The circled numbers mark the four contributions, in reading order. Part~1 (contributions 1--3) uses the test splits: \textcircled{\scriptsize 1} a paired English--Swedish benchmark, shown with a real item; \textcircled{\scriptsize 2} the refinement loop of Section~\ref{sec:refinement}; \textcircled{\scriptsize 3} the English--Swedish gap of the open models by size, with reasoning off (hollow) and on (filled). Part~2 (contribution 4) uses the training splits: teachers write reasoning traces for the translated training questions, the 1.2B and 1.7B students are fine-tuned on them, and the bars give Swedish accuracy before training (gray) and for the best student; the dashed line marks a 20B open model in its fast mode (50.6\%).}
    \label{fig:overview}
\end{figure}

Swedish is spoken by about 10 million people\footnote{\raggedright Swedish Institute and the Language Council of Sweden, \textit{Svenska språket} (2022), \url{https://si.se/app/uploads/2023/10/svenska-spraket.pdf}.}, but it makes up 0.15\% of the pretraining documents of Llama~2~\citep{touvron2023llama} and 0.61\% of the tokens of the multilingual corpus CulturaX~\citep{nguyen2024culturax}, against 89.7\% and 45.1\% for English. Swedish groups have therefore built their own models for what they call a smaller language~\citep{ekgren2024gptsw3}. Multilingual medical studies such as \textit{MedExpQA}~\citep{alonso2024medexpqa} and \textit{MMedBench}~\citep{qiu2024towards}, and studies of health questions asked in other languages~\citep{jin2024better}, find that accuracy in English does not carry over to other languages. In medicine, a change in one term, a negation, or a dose can change the meaning of a question, so the size of this loss in Swedish decides whether a model can be used there.

Measuring the loss needs the same questions in both languages. Swedish medical benchmarks exist~\citep{hertzberg2024medqa,moell2025swedish,vakili2025sweclineval}, but apart from a translated PubMedQA, their questions have no English counterparts, so their results mix the effect of language with differences in topic, difficulty, and style. We therefore ask two questions. How much accuracy do LLMs, and models of at most 4B parameters in particular, lose when a medical question is asked in Swedish instead of English? And can the English data and the open models that already exist bring a 1--2B model, which runs on local hardware, to the Swedish accuracy of open models ten or more times its size? We answer them in two connected parts, which Figure~\ref{fig:overview} shows.

\paragraph{Part 1: a paired benchmark.} We translate the test sets of ten English multiple-choice datasets into Swedish. Each Swedish item keeps the options, their order, and the gold answer of its English source, so a model answers the same question in both languages. Because the comparison is only as fair as the translation, a conservative loop refines it: open models from three families find errors, check them against a composite of Swedish medical term bases, repair them, and keep a model-written edit only if blind judges vote for it. We evaluate only on the 93.9\% of items that the loop leaves without an unresolved finding, and we break the results down by refinement category. The two proprietary models, tested earlier on the unrefined translation, lose 0.3 and 1.4 points in Swedish, and no model loses more than 2.7 points when it reasons. Without reasoning, the loss reaches 10.5 points for a 4B and 12.4 points for a 3B open model.

\paragraph{Part 2: adapting 1--2B models.} We translate the training sets, let two open teacher models write reasoning traces, and fine-tune two students of 1.2B and 1.7B parameters, LFM2.5-1.2B~\citep{amini2025lfm2} and EuroLLM-1.7B~\citep{martins2025eurollm}, on the gold answer alone or on the teacher's reasoning, in English, Swedish, or both. Traces for the Swedish training questions alone bring the English--Swedish gap to 0.3 points for LFM2.5-1.2B and $-0.2$ points for EuroLLM-1.7B and raise accuracy in both languages; gold answers alone raise Swedish accuracy by at most 9.4 points. The best student answers 52.0\% of the Swedish items, against 50.6\% for a 20B open model, GPT-OSS-20B, in its fast mode without reasoning, with 2.3 instead of 14 GB of weights and 3.4 instead of 5.1 TFLOPs per item. On two sets of exam questions written in Swedish, the best student of each base model gains 28 to 44 points, and in the reported runs the trace data rank as on the translated items.

This paper makes four contributions, numbered as in Figure~\ref{fig:overview}:
\begin{enumerate}[leftmargin=*,itemsep=1pt,topsep=3pt]
    \item \textbf{A paired benchmark.} Each of the 9,733 test items has the same options and answer in English and Swedish, and the training questions have a matched Swedish translation.
    \item \textbf{A conservative refinement loop.} Open models find and repair errors, code checks them against Swedish medical sources, and blind judges from three model families must vote for a model-written edit before it is kept. We log every edit and release rule-based and model-written edits separately.
    \item \textbf{An English--Swedish evaluation.} Without reasoning, open models of 27--35B parameters lose 3.0--4.1 points in Swedish, far less than the 3--4B models. For the six models with both modes, reasoning lowers the mean gap from 6.2 to 1.7 points.
    \item \textbf{Students of 1--2B parameters.} Training on teacher reasoning traces that include Swedish raises Swedish accuracy by 28.5 points for LFM2.5-1.2B and 21.0 points for EuroLLM-1.7B, and the best student is 1.0 to 1.3 points above a 20B open model in its fast mode in Swedish over three training seeds. The gains carry over to two exam sets written in Swedish.
\end{enumerate}
We release the benchmark, its edit history, the code, and the students.

\section{Related Work}
\label{sec:related}

\paragraph{Multilingual and Swedish medical evaluation.} \textit{MedExpQA} evaluates LLMs on the same Spanish exam questions in four languages~\citep{alonso2024medexpqa}, and \textit{MMedBench} adds multilingual training data and models~\citep{qiu2024towards}. Both find that accuracy depends on the language of the question. For Swedish, \textit{MedQA-SWE} collects questions from Swedish medical exams~\citep{hertzberg2024medqa}. The \textit{Swedish Medical LLM Benchmark} combines a translated PubMedQA with Swedish exam questions and clinical cases~\citep{moell2025swedish}. We use MedQA-SWE and the exam questions of this benchmark to test our students on questions written in Swedish. \textit{SweClinEval} uses real Swedish clinical text for classification and entity recognition~\citep{vakili2025sweclineval}. We add a paired English--Swedish benchmark over ten datasets, together with matched training data.

\paragraph{Translated benchmarks.} Translation is a common way to make benchmarks comparable across languages. \textit{Global MMLU} studies the cultural and linguistic problems this causes~\citep{singh2025global}, and \textit{MMLU-ProX} combines machine translation with expert review~\citep{xuan2025mmlu}. Both show that translation quality affects the comparison, and even a correct translation can change measured cross-lingual results through its artifacts~\citep{artetxe2020translation}.
Several methods improve a first translation: \textit{TEaR} translates, estimates errors, and refines~\citep{feng2025tear}, and compact open models can estimate translation quality and suggest fixes~\citep{guttmann2026compactqe}. We follow this idea, with error categories in the style of MQM~\citep{lommel2014multidimensional}, but split the work into short, separately checked steps and ground terminology in authoritative Swedish sources. LLM judges prefer certain answer positions and sometimes their own outputs~\citep{zheng2023judging}. We therefore judge in both orders and pool votes from three model families.

\paragraph{Distillation and multilingual tuning.} Rationales from a larger teacher can train a smaller student better than labels alone~\citep{hsieh2023distilling,li2023symbolic}, and medical reasoning models are often trained on traces from stronger models~\citep{chen2024huatuogpt}. Multilingual instruction tuning can make models more robust across languages, but the effect depends on the model and the data mix~\citep{chen2024monolingual}. %
Closest to our Part~2, \citet{ferrazzi2026multilingual} translate MedQA and MedMCQA into Italian and Spanish and fine-tune models of up to 8B parameters on multilingual medical reasoning traces. We compare supervision with the gold answer alone and with reasoning traces on the same source questions, in English, Swedish, and bilingual settings, for 1--2B students. Low-rank adaptation (LoRA)~\citep{hu2022lora} is a cheaper alternative to full fine-tuning, and Appendix~\ref{sec:appendix-model-selection} compares the two for LFM2.5-1.2B.

\section{Method}
\label{sec:method}

The two parts of the study share the same source datasets and the same evaluation: the test splits of English medical multiple-choice datasets become the paired benchmark, and their training splits become the supervision data of the students. Every open model, from the largest to the smallest student, is tested on the same English and Swedish test items with the same protocol, so the results of both parts can be read on one scale.

\subsection{Building the Paired Benchmark}
\label{sec:benchmark-method}

\paragraph{Initial translation.} We translate the question and the options of every test item into Swedish with GPT-5.4\footnote{\texttt{gpt-5.4-2026-03-05}, \url{https://platform.openai.com/docs/models}.}, with a prompt that has three worked examples and asks the model to keep the medical meaning, negation, certainty, numbers, terminology, and the order of the options; Appendix~\ref{sec:appendix-prompts} gives the prompt. Identifiers and answer labels are copied unchanged, so each Swedish item stays paired with its English source. We call this translation the \emph{draft}.

\subsubsection{Refinement Loop}
\label{sec:refinement}

GPT-5.4 still makes translation errors: our loop refines 1,063 of its 10,363 drafts and leaves 624 disputed, with a confirmed error but no accepted edit, and in medicine a single wrong term can change the answer. A model that checks its own work, however, tends to confirm its own habits: it flags correct Swedish and pulls terms toward English. We therefore designed a conservative loop, shown in Figure~\ref{fig:refinement_loop} and fully specified in Appendix~\ref{sec:appendix-refinement}. An edit is kept only if it fixes a confirmed error, passes deterministic checks, and wins a vote of independent judges; only missing units, which code restores from the English source, need no vote. If any step fails, the current text stays.

\begin{figure}[tbp]
    \centering
    \includegraphics[width=\linewidth]{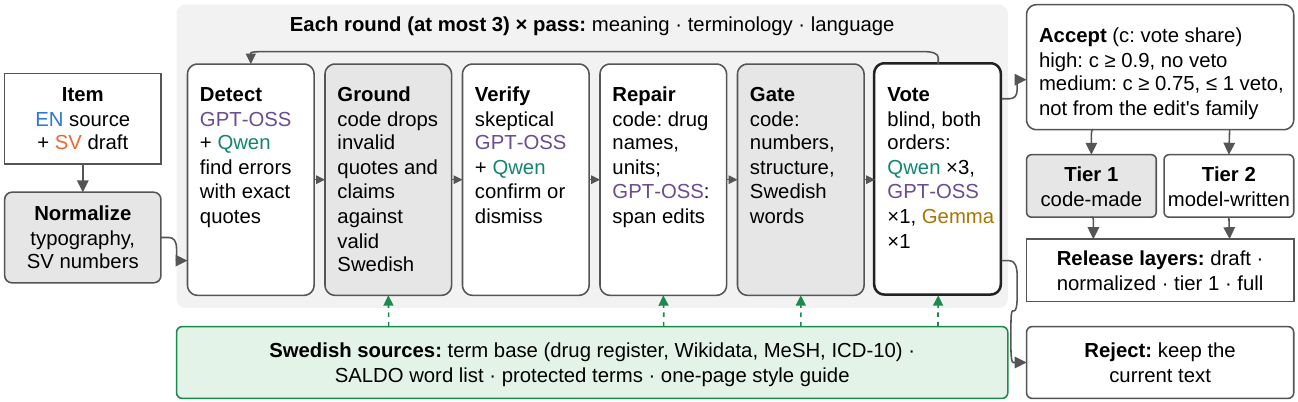}
    \caption{The refinement loop for one item: up to three rounds of a meaning, a terminology, and a language pass, each followed by deterministic gates and, where required, a blind vote. Accepted edits form tier~1, made by rules, or tier~2, written by a model. Gray boxes are code; dashed arrows show where the Swedish sources are used. Figure~\ref{fig:refinement_loop_full} gives a more detailed version.}
    \label{fig:refinement_loop}
\end{figure}

\paragraph{Three model families.} GPT-OSS-120B~\citep{openai2025gptoss120bgptoss20bmodel} is the worker: it finds, checks, and repairs errors. Qwen3.8-27B~\citep{qwen38} is a second detector and checker for meaning errors and the main judge, and Gemma-4-31B is a third judge. None of them wrote the draft, and each candidate receives votes from all three families.

\paragraph{Grounding in Swedish sources.} A composite term base maps English medical terms to accepted Swedish forms. It joins the Swedish drug register with Wikidata~\citep{vrandevcic2014wikidata} for drug names, and uses Swedish MeSH and the Swedish primary-care version of ICD-10 for diseases, anatomy, and procedures. A Swedish word list, mostly from SALDO~\citep{borin2013saldo}, decides whether a word is valid Swedish, and 33 \emph{protected terms} keep established clinical expressions from being replaced. A one-page style guide in every prompt states the main conventions, including the rule that no change is better than a speculative change.

\paragraph{Steps.} Code first normalizes typography and Swedish number style. Each item then runs up to three rounds of three short passes, for \emph{meaning}, \emph{terminology}, and \emph{language}, so that each prompt checks one kind of error instead of a long checklist. In each pass, the models \textbf{detect} errors, code \textbf{grounds} them in the Swedish sources and drops claims against valid Swedish, a skeptical prompt \textbf{verifies} them, and code or the worker \textbf{repairs} them. A deterministic \textbf{gate} then rejects edits that lose a number or add a non-Swedish word, and the judges \textbf{vote}.

\paragraph{Acceptance by vote.} A judge sees the English item, the Swedish context, and two unlabeled versions of the field, the current text and the candidate, but not the claimed error. It answers which version is better, \emph{equal}, or \emph{both wrong}, and gives its reason first. Each candidate except a restored unit gets $n=10$ votes, five in each presentation order: three Qwen samples and one each from GPT-OSS and Gemma. With $n_+$ votes for the candidate, the confidence score is
\begin{equation}
    c = n_+ / n
    \label{eq:confidence}
\end{equation}
A candidate has \emph{high} confidence if $c \geq 0.9$ and no judge prefers the current text or answers \emph{both wrong}, and \emph{medium} confidence if $c \geq 0.75$ with at most one such veto, which must not come from the model family that wrote the edit. All other candidates are rejected. The rule asks for more than a majority but not for unanimity, so one stray vote from another model family cannot block a medium-confidence edit.

\paragraph{Release in tiers.} \emph{Code-made} edits are made by deterministic rules: they put in Swedish drug names from the drug register and restore missing units from the English source, and they form \emph{tier~1}. Edits written by a model form \emph{tier~2}. The benchmark uses the tier-1 text, and Section~\ref{sec:translation-results} explains why. We release every layer and log every finding, edit, and vote, so every released character can be traced.

\subsection{Evaluation Protocol}
\label{sec:protocol}

\paragraph{Evaluation on settled translations.} We evaluate only on the 9,733 of 10,363 items (93.9\%) that the loop leaves without an unresolved finding; the other 630 are the 624 disputed items and six items with a corrupt draft field or an unresolved critical finding. These items are set aside in both languages, so English and Swedish always cover the same items. Section~\ref{sec:benchmark-results} checks whether the gap depends on the translation category, including these items.

\paragraph{Prompts and scoring.} Every core test item is asked in its own language, with the same instruction in English or Swedish. The \emph{explanation prompt} asks for a short explanation and the chosen option; the \emph{answer-only prompt} asks for the option alone; Appendix~\ref{sec:appendix-repro} gives both prompts. Decoding is greedy, the output must match a JSON schema, and reasoning models are run with reasoning on and off. An unfinished, malformed, or impossible answer counts as wrong. We report accuracy weighted by items and the gap $\Delta=\mathrm{EN}-\mathrm{SV}$, which is positive when Swedish is worse; Appendix~\ref{sec:appendix-benchmark} gives the bootstrap intervals, the McNemar tests, and all settings.

\subsection{Adapting 1--2B Models}
\label{sec:adaptation}

The right side of Figure~\ref{fig:overview} shows how we build the students, and Figure~\ref{fig:part2_detail} in Appendix~\ref{sec:appendix-distillation} gives the full pipeline with all counts.

\paragraph{Swedish training data.} We translate the training split with GPT-OSS-120B. Besides the translation, the model returns warnings, such as a risk that the translation changes the answer; we remove the questions flagged this way to keep the gold labels trustworthy. The training translations are not refined; only the test set, on which every reported score depends, is.

\paragraph{Supervision.} \emph{Answer-level} supervision uses the gold option alone, for example \texttt{\{"answer\_index": "B"\}}, with the answer-only prompt; no teacher is needed. For \emph{reasoning-trace} supervision, two open reasoning models answer every training question in both languages: GPT-OSS-120B three times (greedy and twice at temperature~0.7) and Qwen3.6-35B-A3B~\citep{qwen36_35b_a3b} twice (greedy and at temperature~0.7). We keep a response only if its answer is correct and its explanation is not empty, and drop length outliers, so a question can have up to five correct solutions. GPT-OSS traces are kept as a reasoning block before the answer. Qwen reasoning traces are longer (median 754--821 words against 121--147 for GPT-OSS), so we keep only its explanation and answer, which keeps the training sequences short.

\paragraph{Swedish reasoning traces.} GPT-OSS-120B reasons in English even when the question is in Swedish; only its explanation is in Swedish. We call these the Swedish traces, after the language of the question. To test whether the language of the reasoning matters, we build a second Swedish set (\emph{SV-rewritten}) in which GPT-OSS-120B rewrites each trace into Swedish, keeping every medical step and adding no new facts.

\paragraph{Training.} We fine-tune all parameters of each student, including the embeddings, on the response with the standard next-token loss for up to two epochs, and keep the checkpoint with the best validation loss. Each supervision type is trained on English, Swedish, and bilingual (EN+SV) data built from the same questions, as well as SV-rewritten data for reasoning traces. This design tells us whether Swedish data is needed, whether English training transfers, and whether reasoning traces help more than answers. Reasoning-trace students are tested with \emph{reasoning} decoding, which reasons before the answer, and with \emph{direct} decoding, which answers at once. Appendix~\ref{sec:appendix-training} gives the settings and shows that frozen and trained embeddings reach 80.9\% and 80.8\% on English held-out questions, and 77.7\% and 78.7\% on Swedish ones.

\section{Experiments}
\label{sec:experiments}

\subsection{Data and Models}
\label{sec:datasets}

We use the medical question-answering collection of \textit{MedicalAgentsBench}~\citep{shao2026medicalagentsbenchcomplexmedicalreasoning}, which puts established multiple-choice benchmarks into one format with stable item identifiers and fixed splits. Its ten dataset configurations are \textit{MedExQA}~\citep{kim2024medexqa}, the health part of \textit{MMLU-Pro}~\citep{wang2024mmlu}, \textit{AfriMedQA}~\citep{nimo2025afrimed}, the medical and biology parts of \textit{MMLU}~\citep{hendrycks2020measuring}, \textit{MedQA}~\citep{jin2021disease}, \textit{MedMCQA}~\citep{pal2022medmcqa}, \textit{PubMedQA}~\citep{jin2019pubmedqa}, \textit{MedBullets}~\citep{chen2025benchmarking}, and the reasoning and understanding parts of \textit{MedXpertQA}~\citep{zuo2025medxpertqa}. Together, they cover licensing exams, specialist exams, biomedical research abstracts, and expert-level clinical reasoning, and using a shared, published collection keeps our splits comparable with other work. The paired benchmark has 10,363 test items in each language, of which we report the 9,733 core items. The collection also marks a \emph{hard} subset of questions that fewer than half of the models tested by its authors answered correctly; 840 of them are core items, and we call the other 8,893 core items the \emph{remainder}. The training splits have 141,735 questions, 120,765 of them from MedMCQA, and give 549,162 English and 549,754 Swedish reasoning-trace examples; we hold out 5\% of the training questions for model selection and never use the test splits for training or checkpoint selection. Table~\ref{tab:dataset_statistics} gives the sizes of every dataset, and Appendix~\ref{sec:appendix-distillation} those of the training data.

\begin{table}[tbp]
    \centering
    \small
\begin{tabular}{llrrrr}
\toprule
\textbf{Dataset} & \textbf{Content} & \textbf{Train} & \textbf{Test} & \textbf{Core} & \textbf{Hard} \\
\midrule
MedExQA & five allied health professions & 20 & 935 & 921 & 100 \\
MMLU-Pro & health questions, ten options & 5 & 818 & 783 & 100 \\
AfriMedQA & African medical exams & 124 & 174 & 166 & 32 \\
MMLU & medicine and biology & 5 & 1,089 & 1,060 & 73 \\
MedQA & US licensing exam (USMLE) & 10,178 & 1,273 & 1,179 & 100 \\
MedXpertQA-U & expert-level knowledge & 5 & 589 & 544 & 100 \\
MedMCQA & Indian medical entrance exams & 120,765 & 2,816 & 2,666 & 100 \\
MedXpertQA-R & expert-level clinical reasoning & 5 & 1,861 & 1,665 & 100 \\
PubMedQA & research abstracts (yes / no / maybe) & 450 & 500 & 466 & 100 \\
MedBullets & USMLE-style clinical cases & 10,178$^\ast$ & 308 & 283 & 89 \\
\midrule
\textbf{Total} & & \textbf{141,735} & \textbf{10,363} & \textbf{9,733} & \textbf{894} \\
\bottomrule
\end{tabular}

    \caption{The ten datasets. Train and test are the splits of the \textit{MedicalAgentsBench} collection; hard is its released hard subset of the test split. Core: the test items on which we report results, all except the 630 items set aside in Section~\ref{sec:protocol}; Appendix~\ref{sec:appendix-categories} describes them. $^\ast$The MedBullets training file is a copy of the MedQA training file.}
    \label{tab:dataset_statistics}
\end{table}

Table~\ref{tab:models} in Appendix~\ref{sec:appendix-benchmark} lists all models and their roles. In Part~1 we evaluate two proprietary models, GPT-5.4 and Claude Sonnet 4.6\footnote{\url{https://www.anthropic.com/claude/sonnet}.}, which we tested in an earlier run on the unrefined translation, and eight open models, six of which can reason or answer directly; the three loop models are among them. Appendix~\ref{sec:appendix-benchmark} also reports Qwen3.5-35B-A3B, the previous version of Qwen3.6-35B-A3B, which we tested in Swedish only on the unrefined translation. In Part~2, the students are LFM2.5-1.2B-Instruct, a compact general model, and EuroLLM-1.7B-Instruct, a multilingual model pretrained with Swedish among other European languages, which shows whether a model that already knows Swedish adapts differently.

\subsection{Translation Quality}
\label{sec:translation-results}

\begin{figure}[tbp]
    \centering
    \includegraphics[width=\linewidth]{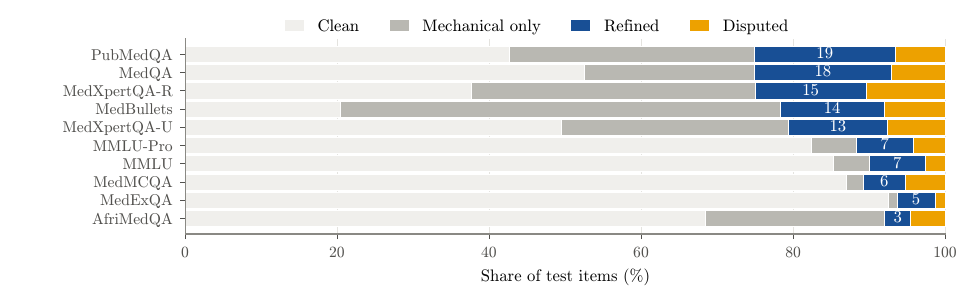}
    \caption{Outcome of the refinement per dataset for the 10,363 test items (refined share in \%). Appendix~\ref{sec:appendix-translation-quality} gives the counts and the review of the edits.}
    \label{fig:refine_overview}
\end{figure}

\paragraph{The loop refines 10.3\% of the items.} Figure~\ref{fig:refine_overview} shows the outcome per dataset. The loop leaves 67.2\% of the drafts untouched. In another 16.5\% only the Swedish number and typography conventions were applied. It refines 1,063 items (10.3\%) with 1,252 accepted edits, and the edits are short (median 6 characters per changed field). Edits concentrate on clinical vignettes and research abstracts: the loop refines 18.0\% of the MedQA items, 13.1--14.6\% of the MedXpertQA items, and 18.6\% of the PubMedQA items, against 5.0--7.5\% of the knowledge questions of MedExQA, MedMCQA, MMLU, and MMLU-Pro. A further 624 items stay disputed and \textit{are set aside from the reported results}.

\paragraph{Accepted edits.} Table~\ref{tab:refine_edits} lists the accepted edits. Tier~1 holds 294 edits made by code, which replace English drug names by the Swedish form of their International Nonproprietary Name (\textit{vancomycin}~$\rightarrow$~\textit{vankomycin}) and restore missing secondary units from the source. Tier~2 holds 958 edits written by a model and accepted by the judges, from grammar edits to meaning repairs such as a reversed finding (\textit{ömmande}~$\rightarrow$~\textit{oöm} for ``non-tender'').

\begin{table}[t]
    \centering
    \footnotesize
    \setlength{\tabcolsep}{2pt}
\begin{tabular}{@{}lrrrrr@{}}
\toprule
& & \multicolumn{2}{c}{\textbf{Confidence}} & \multicolumn{2}{c}{\textbf{Check (\%)}} \\
\cmidrule(lr){3-4}\cmidrule(l){5-6}
\textbf{Edit class} & \textbf{Edits} & \textbf{High} & \textbf{Med.} & \textbf{Better} & \textbf{Worse} \\
\midrule
\multicolumn{6}{@{}l}{\textit{Tier 1 (deterministic rules)}} \\
Drug name $\rightarrow$ Swedish & 156 & 156 & 0 & 100 & 0 \\
Missing unit & 138 & \multicolumn{2}{c}{no vote} & 50 & 0 \\
\textit{Total (weighted)} & \textit{294} & 156 & 0 & \textit{77} & \textit{0} \\
\midrule
\multicolumn{6}{@{}l}{\textit{Tier 2 (model)}} \\
Grammar & 377 & 271 & 106 & 47 & 10 \\
English left in Swedish & 209 & 136 & 73 & 50 & 9 \\
Meaning & 190 & 141 & 49 & 45 & 18 \\
Term-base terminology & 159 & 140 & 19 & 64 & 32 \\
Drug name, reworded & 16 & 15 & 1 & 100 & 0 \\
Spelling/gender & 7 & 4 & 3 & 100 & 0 \\
\textit{Total (weighted)} & \textit{958} & 707 & 251 & \textit{51} & \textit{15} \\
\bottomrule
\end{tabular}
    \caption{Accepted edits by class, tier, and confidence level, with a blinded per-edit check of 129 sampled edits by Claude Opus 5.5 with \emph{xhigh} reasoning effort: \% better or worse than the text just before the edit, the rest equal. Missing units are restored by a deterministic rule without a vote, so they have no confidence level. Totals are weighted by the number of edits per class. Appendix~\ref{sec:appendix-translation-quality} gives the full table with the number of reviewed edits per class.}
    \label{tab:refine_edits}
\end{table}

\paragraph{Checks and blinded review.} The judges catch 94--100\% of injected errors, such as a removed negation or a changed number, and verification accepts none of 2,824 planted false meaning claims. Edits touch the correct option and the incorrect options at similar rates, 0.80\% and 0.74\%. In a blinded item-level review fixed in advance, two separate runs of Claude Opus 5.5 with \emph{xhigh} reasoning effort, a model family outside the loop, compared 200 sampled items with their text before the edits (agreement 85\%, $\kappa=0.78$). None of the 40 sampled items with only code-made edits was worse. In this item-level review, model-written edits were better 4.3 times as often as worse, but the upper 95\% bound of their harm rate, 17.1\%, is above the 10\% bar we set, so the benchmark uses only the code-made edits. Appendix~\ref{sec:appendix-expert} gives the protocol and the results per confidence level.

\subsection{Benchmark Results}
\label{sec:benchmark-results}

\begin{figure}[tbp]
    \centering
    \begin{subfigure}[t]{0.49\linewidth}
        \centering
        \includegraphics[width=\linewidth]{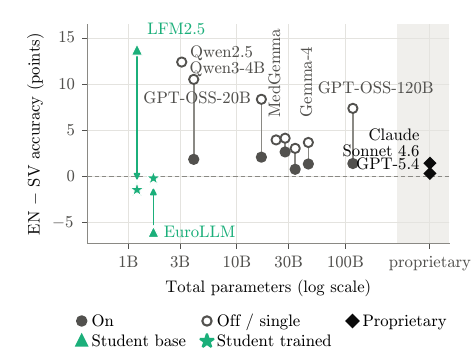}
        \caption{Gap by model size}
        \label{fig:gap_vs_size}
    \end{subfigure}\hfill
    \begin{subfigure}[t]{0.49\linewidth}
        \centering
        \includegraphics[width=\linewidth]{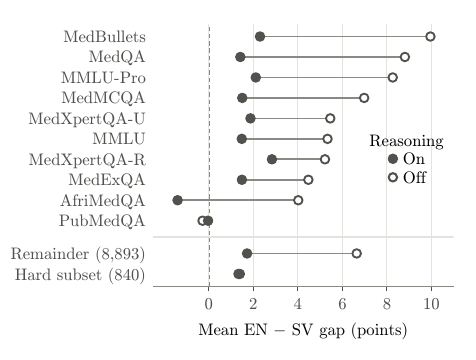}
        \caption{Gap by dataset}
        \label{fig:dataset_gap}
    \end{subfigure}
    \caption{(a) The English--Swedish gap is smaller for the larger models we tested and with reasoning. Open models are shown with reasoning on (filled) and off (hollow); proprietary models, whose sizes are not published, sit on the right. Triangles mark the students before training, and arrows lead to their best Swedish model after training on reasoning traces (stars): LFM2.5-1.2B on English and Swedish traces, and EuroLLM-1.7B on Swedish traces. (b) Mean English--Swedish gap per dataset over the six open models with both modes, with reasoning off (hollow) and on (filled). The bottom rows split the core items into the hard subset and the remainder.}
    \label{fig:gap}
\end{figure}

\paragraph{The gap is at most 2.7 points with reasoning and up to 12.4 points without.} Table~\ref{tab:benchmark_main} shows that every benchmark model is less accurate in Swedish. With reasoning, no model loses more than 2.7 points. On the unrefined translation, the proprietary models lose 0.3 (GPT-5.4) and 1.4 points (Claude Sonnet 4.6), and the best open model, Gemma-4-31B with reasoning, reaches 78.0\% in Swedish and loses 0.8 points. Models that answer directly lose 3.0 to 12.4 points, the 3--4B models the most: Qwen3-4B 10.5 points and Qwen2.5-3B 12.4. Figure~\ref{fig:gap_vs_size} shows this pattern against model size. Of the students we adapt in Part~2, LFM2.5-1.2B starts at the far end, losing 13.8 points in Swedish before training, while EuroLLM-1.7B is 6.0 points more accurate in Swedish than in English, where many of its outputs are invalid.

\begin{table}[tbp]
    \centering
    \begin{minipage}[t]{0.485\linewidth}
    \centering
    \footnotesize
    \setlength{\tabcolsep}{3.5pt}
\begin{tabular}{lrrrrrr}
\toprule
& \multicolumn{3}{c}{\textbf{Reasoning on}} & \multicolumn{3}{c}{\textbf{Reasoning off}} \\
\cmidrule(lr){2-4}\cmidrule(lr){5-7}
\textbf{Model} & \textbf{EN} & \textbf{SV} & $\boldsymbol{\Delta}$ & \textbf{EN} & \textbf{SV} & $\boldsymbol{\Delta}$ \\
\midrule
\multicolumn{7}{l}{\textit{Proprietary}} \\
GPT-5.4 & 81.3 & 81.0 & \pos{0.3} & -- & -- & -- \\
Claude Sonnet 4.6 & 77.7 & 76.3 & \pos{1.4} & -- & -- & -- \\
\midrule
\multicolumn{7}{l}{\textit{Open-weight}} \\
Gemma-4-31B & 78.7 & 78.0 & \pos{0.8} & 77.1 & 74.0 & \pos{3.0} \\
Qwen3.6-35B-A3B & 74.7 & 73.3 & \pos{1.3} & 72.2 & 68.5 & \pos{3.7} \\
Qwen3.8-27B & 74.6 & 72.0 & \pos{2.7} & 70.6 & 66.5 & \pos{4.1} \\
GPT-OSS-120B & 71.8 & 70.4 & \pos{1.4} & 64.3 & 56.9 & \pos{7.4} \\
GPT-OSS-20B & 65.8 & 63.7 & \pos{2.1} & 59.0 & 50.6 & \pos{8.4} \\
MedGemma-27B & -- & -- & -- & 64.5 & 60.5 & \pos{4.0} \\
Qwen3-4B-Thinking & 59.3 & 57.4 & \pos{1.8} & 53.9 & 43.3 & \pos{10.5} \\
Qwen2.5-3B & -- & -- & -- & 44.9 & 32.5 & \pos{12.4} \\
\bottomrule
\end{tabular}

    \caption{Paired benchmark accuracy (\%, weighted by items) on the 9,733 core test items. $\Delta=\mathrm{EN}-\mathrm{SV}$, computed before rounding (red: Swedish lower). Models without a reasoning mode are listed under ``reasoning off''; proprietary models use medium reasoning and were tested in an earlier run on the unrefined translation of all items, so their rows are per-dataset accuracies weighted by the core items, without item-level results. Sizes are in Table~\ref{tab:models}, per-dataset results in Appendix~\ref{sec:appendix-benchmark}.}
    \label{tab:benchmark_main}
    \end{minipage}\hfill
    \begin{minipage}[t]{0.485\linewidth}
    \centering
    \footnotesize
    \setlength{\tabcolsep}{4pt}
\begin{tabular}{lrrrrrr}
\toprule
& \multicolumn{3}{c}{\textbf{LFM2.5-1.2B}} & \multicolumn{3}{c}{\textbf{EuroLLM-1.7B}} \\
\cmidrule(lr){2-4}\cmidrule(lr){5-7}
& \multicolumn{2}{c}{\textit{Tested on}} & & \multicolumn{2}{c}{\textit{Tested on}} & \\
\textbf{Trained on} & \textbf{EN} & \textbf{SV} & $\boldsymbol{\Delta}$ & \textbf{EN} & \textbf{SV} & $\boldsymbol{\Delta}$ \\
\midrule
\multicolumn{7}{l}{\textit{Answer-level}} \\
None (base model) & 36.3 & 23.8 & 12.4 & 14.7 & 21.3 & -6.6 \\
EN & 35.2 & 24.8 & 10.4 & 33.7 & 30.0 & 3.6 \\
SV & 27.4 & 27.8 & -0.4 & 30.9 & \textbf{30.7} & 0.1 \\
EN+SV & 33.2 & \textbf{30.7} & 2.5 & 31.6 & 29.4 & 2.2 \\
\midrule
\multicolumn{7}{l}{\textit{Reasoning trace}} \\
None (base model) & 37.2 & 23.5 & 13.8 & 17.5 & 23.5 & -6.0 \\
EN & 49.0 & 14.5 & 34.5 & 43.2 & 39.4 & 3.8 \\
SV & 50.1 & 49.8 & 0.3 & 44.3 & \textbf{44.5} & -0.2 \\
SV (rewritten) & 44.0 & 44.3 & -0.3 & 39.4 & 39.0 & 0.5 \\
EN+SV & 50.5 & \textbf{52.0} & -1.5 & 41.9 & 44.5 & -2.6 \\
\bottomrule
\end{tabular}

    \caption{Accuracy (\%) of the students on the English and Swedish test items, by the data they were trained on. Answer-level models and the base model above them use the answer-only prompt; reasoning-trace models use the explanation prompt and reason before answering. $\Delta=\mathrm{EN}-\mathrm{SV}$, computed before rounding. Bold: best Swedish accuracy per student and supervision. SV (rewritten): Swedish data with the reasoning rewritten into Swedish. Chance is 20.7\%; the base EuroLLM-1.7B is below it in English because 29--47\% of its outputs are invalid. Appendix~\ref{sec:appendix-results} gives more conditions, the share of valid outputs, and paired tests.}
    \label{tab:main_adaptation_results}
    \end{minipage}
\end{table}

\paragraph{Reasoning narrows the gap.} For the six models with both modes, the average gap falls from 6.2 points with reasoning off to 1.7 points with reasoning on, and reasoning helps Swedish more than English for every one of them, by 1.5 to 8.7 points with paired 95\% intervals that exclude zero. The gap is significant for all 14 open configurations under the exact McNemar test with the Holm correction, and Appendix~\ref{sec:appendix-benchmark} gives the intervals. By a rule-based count, five of the six models restate the Swedish question in English when they reason, as Section~\ref{sec:analysis-language} shows.

\paragraph{The gap is not concentrated in the errors the loop found.} The 11 open configurations of 20B or more lose 3.4 points on clean items, 3.6 on items with a model-written edit, and 3.4 on the disputed items that we set aside, and 3.2 to 4.1 points on every category when each is reweighted to the dataset mix of the core. Evaluating on all 10,363 items instead of the core changes the gap of each configuration by at most 0.32 points. Errors that the loop missed may still contribute. Appendix~\ref{sec:appendix-categories} gives these checks.

\paragraph{Datasets.} Figure~\ref{fig:dataset_gap} shows the gap per dataset for the six open models with both modes. Without reasoning, the gap is largest on MedBullets, MedQA, and MMLU-Pro, with 10.0, 8.8, and 8.3 points. With reasoning, it stays below 3 points on every dataset. PubMedQA shows a gap of $-0.3$ points without and 0.0 with reasoning, possibly because its yes/no/maybe answers depend more on reading an abstract than on medical terms. Appendix~\ref{sec:appendix-benchmark} gives the accuracy of every model per dataset.

\subsection{Adaptation Results}
\label{sec:adaptation-results}

\begin{figure}[tbp]
    \centering
    \includegraphics[width=\textwidth]{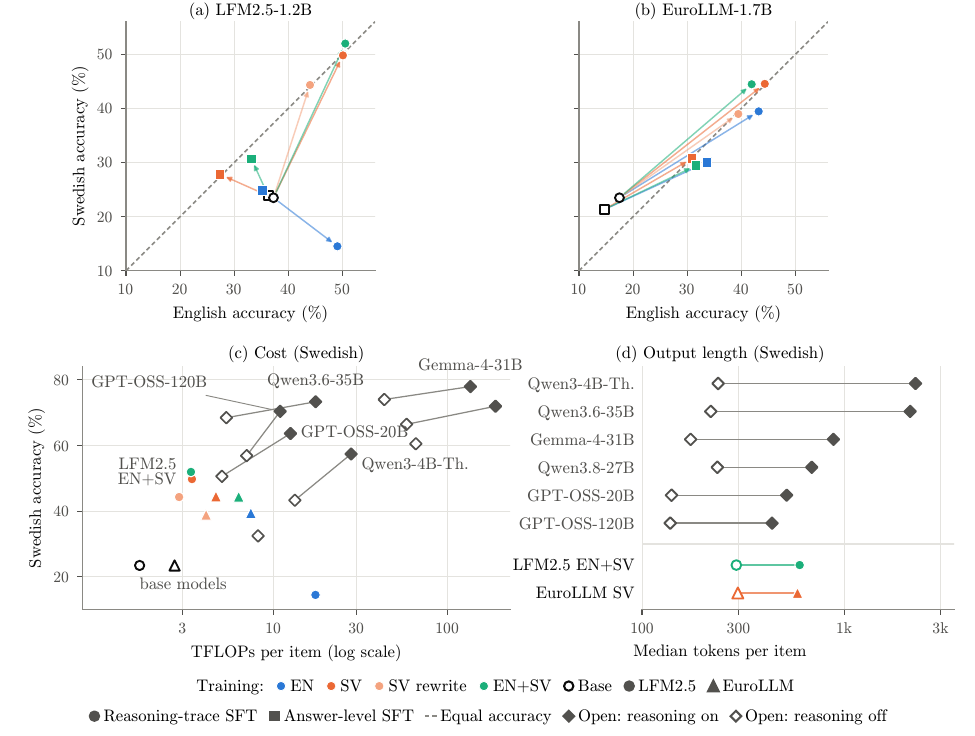}
    \caption{(a, b) Where training moves each student. Each point is one trained model at its accuracy on English (x) and Swedish (y) test items. Points on the dashed line answer both languages equally well, and arrows start at the base model. (c) Swedish accuracy against inference compute per item ($2N_\mathrm{active}$ FLOPs per processed token, log scale). Open models are gray diamonds, with reasoning on filled and off hollow. Students with reasoning are colored by training data: circles are LFM2.5, triangles EuroLLM, and hollow black markers the base models. (d) Median generated tokens per Swedish item with reasoning off (hollow) and on (filled), for the open models and the best Swedish model of each student. SFT: supervised fine-tuning.}
    \label{fig:train_test_shift}
\end{figure}

We now ask whether training can close the gap of 1--2B models, and at what cost. Table~\ref{tab:main_adaptation_results} lists the accuracy of each student by its training data, and Figure~\ref{fig:train_test_shift} shows the same results as movements away from the base model, next to their inference cost.

\paragraph{Swedish data gives the best Swedish models.} English traces raise LFM2.5-1.2B in English, from 37.2\% to 49.0\%, but lower it in Swedish, from 23.5\% to 14.5\%, because more than half of its Swedish reasoning outputs do not finish, as Section~\ref{sec:analysis-english} shows. Swedish traces give 50.1\% in English and 49.8\% in Swedish, so they also lift English, although the model saw no English question in training, only English reasoning. Bilingual traces are best in Swedish, with 52.0\%, 2.2 points above Swedish traces, and within 0.4 points of them in English; with two further training seeds, these leads are 2.5 and 1.9 points in Swedish, all significant, and 1.2 and 0.6 points in English, none significant after the Holm correction, as Appendix~\ref{sec:appendix-seeds} shows. The bilingual data, however, also has twice the examples and training steps. EuroLLM-1.7B, pretrained with Swedish, gains in both languages from every kind of trace: English traces lift its Swedish accuracy from 23.5\% to 39.4\%, and Swedish traces give its best model, with 44.3\% in English and 44.5\% in Swedish; bilingual traces also reach 44.5\% in Swedish, but only 41.9\% in English.

\paragraph{Gold answers alone add at most 9.4 points in Swedish.} Bilingual answer-level training raises LFM2.5-1.2B in Swedish from 23.8\%, its base accuracy with the answer-only prompt, to 30.7\%, 6.8 points, against 28.5 points with bilingual traces, which have about four times as many examples, as Table~\ref{tab:cost_pipeline} shows; the best EuroLLM answer-level model gains 9.4 points in Swedish, from 21.3\% to 30.7\%. For LFM2.5-1.2B, gold answers alone close the gap mainly by lowering English: answer-level training on Swedish lowers it in English from 36.3\% to 27.4\% and raises it in Swedish only from 23.8\% to 27.8\%, while bilingual reasoning traces raise both languages.

\paragraph{What the students cost.} The best LFM2.5-1.2B student needs about 3.4 TFLOPs per Swedish item and 2.3 GB for its weights; the best EuroLLM-1.7B student needs 4.7 TFLOPs and 3.3 GB. GPT-OSS-20B in its fast mode, without reasoning, is 1.3 points below the LFM2.5-1.2B student in Swedish, with a paired 95\% interval of 0.3 to 2.4 points, as Table~\ref{tab:headline_tests} shows; with two further training seeds, it is 1.2 and 1.0 points below, and no seed's lead is significant after the Holm correction over the seed comparisons of Appendix~\ref{sec:appendix-seeds}. GPT-OSS-20B needs 1.5 times that student's compute and six times its memory; with reasoning, it reaches 63.7\%. The open models of 27B or more remain above every student, at 56.9\% or more in Swedish, and need 54--72 GB. The students also write shorter outputs: a median of 603 (LFM2.5) and 588 tokens (EuroLLM) per Swedish item, against 2,124 for Qwen3.6-35B-A3B and 2,258 for Qwen3-4B-Thinking, which is 5.5 points more accurate than LFM2.5 at 8.3 times its compute. Building the students is a one-time cost of about $28\times10^{18}$ FLOPs to generate the training data and $62\times10^{18}$ to train all students, as Table~\ref{tab:cost_pipeline} shows. Figure~\ref{fig:train_test_shift}c and d and Appendix~\ref{sec:appendix-cost} give the costs of every model.

\subsection{Questions Written in Swedish}
\label{sec:native}

Our Swedish training and test questions are both machine-translated, so the gains above may hold only for translated Swedish. We therefore test every student on two sets written in Swedish, from the theoretical part of the Swedish licensing test for physicians trained outside the EU and EEA: \textit{MedQA-SWE}~\citep{hertzberg2024medqa}, with 3,172 questions, and the Swedish Medical Doctors Test (SMDT) of the Swedish Medical LLM Benchmark~\citep{moell2025swedish}, with the 159 questions that do not repeat MedQA-SWE. Every question has five options, so chance is 20\%, and prompts, decoding, and scoring are those of the benchmark. We used neither set for training or for choosing a student, but we did not search our training questions for similar ones, and the sets are public, so the base models and the teachers may have seen them in pretraining, and teacher exposure could reach the students through the traces. The sets have no English versions, and we tested only the students and their base models on them, so they measure Swedish accuracy, not the gap or the comparison with larger models.

\paragraph{The best trace students gain 28 to 44 points.} Table~\ref{tab:native_swedish} shows large gains on both sets over each base model with the same prompt: bilingual traces lift LFM2.5-1.2B from 21.3\% to 64.8\% on MedQA-SWE and from 22.6\% to 54.1\% on SMDT, and Swedish traces lift EuroLLM-1.7B from 21.0\% to 56.7\% and from 17.0\% to 50.9\%. The sets differ from our benchmark in difficulty, question mix, and the number of options, so accuracies are comparable within a set, not across sets. In the reported runs, the order of the eight trace students with reasoning on the core Swedish items carries over, with a Spearman correlation of 0.98 on both sets, and for each student the trace data with the highest core accuracy also has the highest accuracy on both sets. Only one of these four leads over the next-best trace data is significant after the Holm correction: bilingual traces are 3.2 points above Swedish traces for LFM2.5-1.2B on MedQA-SWE, with a paired 95\% interval of 1.6 to 4.9 points. With seeds 14 and 15, this lead is 2.8 and 0.5 points, and the bilingual student gains 28.3 and 30.2 points on SMDT, where Swedish traces lead with seed 14. With 159 questions, a single SMDT accuracy has a 95\% interval of up to about 8 points either way, and a follow-up question can name the answer to the question before it, which makes all intervals somewhat optimistic. As on the benchmark, LFM2.5-1.2B trained on English traces leaves most of its Swedish reasoning outputs unfinished, 53.0\% on MedQA-SWE and 62.3\% on SMDT, and falls below its base model; its median output has more than 11,000 tokens on both sets, against 454 to 661 for the other trained trace students, as on the core items.

\begin{table}[tbp]
    \centering
    \small
    \setlength{\tabcolsep}{4pt}
\begin{tabular}{lrrrrrrrrrrrr}
\toprule
& \multicolumn{6}{c}{\textbf{LFM2.5-1.2B}} & \multicolumn{6}{c}{\textbf{EuroLLM-1.7B}} \\
\cmidrule(lr){2-7}\cmidrule(lr){8-13}
& \multicolumn{2}{c}{Core SV} & \multicolumn{2}{c}{MedQA-SWE} & \multicolumn{2}{c}{SMDT} & \multicolumn{2}{c}{Core SV} & \multicolumn{2}{c}{MedQA-SWE} & \multicolumn{2}{c}{SMDT} \\
\cmidrule(lr){2-3}\cmidrule(lr){4-5}\cmidrule(lr){6-7}\cmidrule(lr){8-9}\cmidrule(lr){10-11}\cmidrule(lr){12-13}
\textbf{Trained on} & Acc. & Tok. & Acc. & Tok. & Acc. & Tok. & Acc. & Tok. & Acc. & Tok. & Acc. & Tok. \\
\midrule
\multicolumn{13}{l}{\textit{Answer-level}} \\
None (base model) & 23.8 & 11 & 24.5 & 11 & 23.9 & 11 & 21.3 & 15 & 19.8 & 15 & 19.5 & 15 \\
EN & 24.8 & 8 & 23.7 & 8 & \textbf{25.8} & 8 & 30.0 & 12 & \textbf{28.3} & 12 & \textbf{28.9} & 12 \\
SV & 27.8 & 8 & 22.9 & 8 & 24.5 & 8 & \textbf{30.7} & 12 & 28.1 & 12 & 24.5 & 12 \\
EN+SV & \textbf{30.7} & 8 & \textbf{25.5} & 8 & 17.0 & 8 & 29.4 & 12 & 26.5 & 12 & 18.2 & 12 \\
\midrule
\multicolumn{13}{l}{\textit{Reasoning trace}} \\
None (base model) & 23.5 & 143 & 21.3 & 142 & 22.6 & 156 & 23.5 & 48 & 21.0 & 57 & 17.0 & 67 \\
EN & 14.5 & 11,375 & 16.0 & 11,404 & 8.2 & 11,161 & 39.4 & 505 & 48.3 & 454 & 44.0 & 542 \\
SV & 49.8 & 595 & 61.5 & 567 & 51.6 & 632 & \textbf{44.5} & 588 & \textbf{56.7} & 567 & \textbf{50.9} & 631 \\
SV (rewritten) & 44.3 & 590 & 54.3 & 605 & 49.7 & 661 & 39.0 & 568 & 48.4 & 572 & 37.1 & 634 \\
EN+SV & \textbf{52.0} & 603 & \textbf{64.8} & 579 & \textbf{54.1} & 628 & 44.5 & 585 & 55.1 & 571 & 45.3 & 602 \\
\bottomrule
\end{tabular}

    \caption{Swedish accuracy (\%) of the students on the core Swedish items of our benchmark (Core SV), which are translated, and on two sets written in Swedish: MedQA-SWE (3,172 questions) and SMDT (159 questions that do not repeat MedQA-SWE). Prompts and decoding are those of Table~\ref{tab:main_adaptation_results}; reasoning-trace models reason before answering. Acc.: accuracy (\%); Tok.: median number of generated tokens per question. Chance is 20\% on both native sets. Bold: best accuracy per student, set, and supervision.}
    \label{tab:native_swedish}
\end{table}

\paragraph{Answer-level training helps much less.} LFM2.5-1.2B changes by $-6.9$ to $+1.9$ points over its base model, EuroLLM-1.7B by $-1.3$ to $+9.4$ points, no answer-level student exceeds 28.9\%, and the order of the eight configurations with the answer-only prompt carries over less well, with Spearman correlations of 0.79 and 0.02. Appendix~\ref{sec:appendix-native} gives every configuration with its interval and the paired tests.

\section{Analysis: What Makes the Adaptation Work?}
\label{sec:analysis}

Section~\ref{sec:adaptation-results} showed that reasoning traces with Swedish close the gap of LFM2.5-1.2B and bring EuroLLM-1.7B, which starts with Swedish above its English, to parity, while both students gain in both languages. We now ask where the Swedish gain goes, in which language a student should reason, and when English data is enough.

\begin{table}[t]
    \centering
    \footnotesize
    \setlength{\tabcolsep}{3pt}
    \begin{minipage}[t]{0.53\linewidth}
        \centering
        \vspace{0pt}
\begin{tabular}{@{}lrrrrr@{}}
\toprule
& & \multicolumn{2}{c}{\textbf{Base}} & \multicolumn{2}{c}{\textbf{EN+SV traces}} \\
\cmidrule(lr){3-4}\cmidrule(l){5-6}
\textbf{Test set} & \textbf{Train \%} & EN & SV & EN & SV \\
\midrule
MedMCQA & 85.2 & 40.7 & 32.9 & 59.9 & 62.2 \\
MedBullets & 7.2 & 35.7 & 21.6 & 55.1 & 60.8 \\
MedQA & 7.2 & 41.2 & 28.4 & 71.2 & 71.8 \\
PubMedQA & 0.296 & 66.3 & 54.1 & 70.8 & 70.8 \\
AfriMedQA & 0.083 & 41.6 & 19.9 & 62.0 & 62.0 \\
MedExQA & 0.015 & 58.4 & 21.1 & 63.1 & 64.7 \\
MMLU-Pro & 0.003 & 28.0 & 11.5 & 36.5 & 34.7 \\
MedXpertQA-R & 0.002 & 11.2 & 9.4 & 15.0 & 15.9 \\
MMLU & 0.002 & 53.6 & 22.4 & 65.0 & 68.2 \\
MedXpertQA-U & 0.001 & 11.6 & 9.2 & 15.6 & 16.7 \\
\midrule
Hard items (840) & & 17.4 & 21.3 & 18.7 & 21.9 \\
Other items (8,893) & & 39.1 & 23.7 & 53.5 & 54.8 \\
All core items (9,733) & 100.0 & 37.2 & 23.5 & 50.5 & 52.0 \\
\bottomrule
\end{tabular}

        \caption{LFM2.5-1.2B before and after bilingual reasoning-trace training: accuracy (\%) per test set, sorted by its share of the training data, on the 840 hard items, and overall.}
        \label{tab:transfer_main}
    \end{minipage}\hfill
    \begin{minipage}[t]{0.45\linewidth}
        \centering
        \vspace{0pt}
\begin{tabular}{lrrrr}
\toprule
 & \multicolumn{2}{c}{\textbf{Opening (\%)}} & \multicolumn{2}{c}{\textbf{Gap EN$-$SV}} \\
\cmidrule(lr){2-3} \cmidrule(lr){4-5}
\textbf{Model} & EN & SV & Direct & Reas. \\
\midrule
GPT-OSS-120B & 100.0 & 0.0 & 7.4 & 1.4 \\
GPT-OSS-20B & 99.9 & 0.0 & 8.4 & 2.1 \\
Gemma-4-31B & 99.9 & 0.0 & 3.0 & 0.8 \\
Qwen3.6-35B-A3B & 99.9 & 0.0 & 3.7 & 1.3 \\
Qwen3.8-27B & 2.6 & 95.8 & 4.1 & 2.7 \\
Qwen3-4B-Thinking & 100.0 & 0.0 & 10.5 & 1.8 \\
\midrule
LFM2.5 SV & 97.8 & 0.0 & 0.8 & 0.3 \\
LFM2.5 SV-RW & 0.0 & 99.6 & -0.1 & -0.3 \\
EuroLLM SV & 97.4 & 0.0 & 0.6 & -0.2 \\
EuroLLM SV-RW & 0.0 & 99.1 & 0.6 & 0.5 \\
\bottomrule
\end{tabular}

        \caption{Opening language of the Swedish reasoning traces (\%) and gap (points) of the open models and of students trained on Swedish traces. SV-RW: reasoning rewritten into Swedish.}
        \label{tab:reasoning_main}
    \end{minipage}
\end{table}

\subsection{Where Does the Swedish Gain Go?}
\label{sec:analysis-gains}

Of the 28.5-point Swedish gain of the bilingual LFM2.5-1.2B model, 13.3 points also appear in English, and the other 15.2 points close its gap, from 13.8 to $-1.5$ points. Table~\ref{tab:transfer_main} shows that the gain is not limited to the datasets that dominate the training data. MedMCQA gives 85.2\% of the training examples and rises from 32.9\% to 62.2\% in Swedish, but MMLU, with 0.002\%, rises from 22.4\% to 68.2\%, MedExQA from 21.1\% to 64.7\%, and AfriMedQA from 19.9\% to 62.0\%. The gain is smallest on the hardest questions. On MedXpertQA the student stays below 17\% in both languages, and on the 840 hard items it answers 21.9\% in Swedish, against 21.3\% for its base model, while on the other 8,893 items it rises from 23.7\% to 54.8\%. The bilingual EuroLLM-1.7B model even falls on the hard items, from 22.0\% to 16.8\% in Swedish, as Table~\ref{tab:hard_subset} shows. The hard items differ in their dataset mix, sit near chance, and were selected because models fail on them, so these results show where the gain is small, not why.

\subsection{Which Language Should the Student Reason In?}
\label{sec:analysis-language}

\begin{figure}[t]
    \centering
    \includegraphics[width=\linewidth]{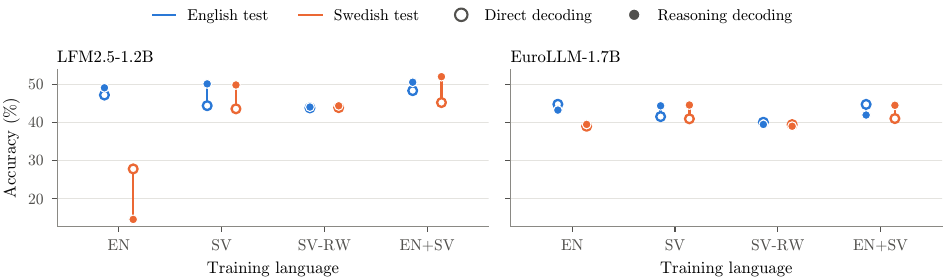}
    \caption{Direct (hollow) and reasoning (filled) decoding of each reasoning-trace student, on English (blue) and Swedish (orange) test items. SV-RW: Swedish data with the reasoning rewritten into Swedish.}
    \label{fig:reasoning_decoding}
\end{figure}

Table~\ref{tab:reasoning_main} shows that five of the six open models open at least 99.9\% of their Swedish reasoning traces in English, and a rule counts 98.8 to 99.6\% of them as restatements of the question, as Appendix~\ref{sec:analysis-reasoning} describes. The teacher GPT-OSS-120B does the same, so the students trained on the original Swedish data learn to reason in English: LFM2.5-1.2B opens 97.8\% and EuroLLM-1.7B 97.4\% of its Swedish traces in English. Students trained on the same data with the reasoning rewritten into Swedish reason in Swedish, but they are less accurate in both languages: in Swedish, 44.3\% against 49.8\% for LFM2.5-1.2B and 39.0\% against 44.5\% for EuroLLM-1.7B; in English, 44.0\% against 50.1\% and 39.4\% against 44.3\%. Figure~\ref{fig:reasoning_decoding} shows what reasoning adds at inference: over direct decoding of the same checkpoint, reasoning adds 0.5 points in Swedish for LFM2.5-1.2B and $-0.4$ for EuroLLM-1.7B when the student reasons in Swedish, against 6.3 and 3.6 points when it reasons in English; for LFM2.5-1.2B, Table~\ref{tab:headline_tests} finds the 6.3 points significant and the 0.5 points not. For our 1--2B students, the original traces with English reasoning thus worked better than the rewritten Swedish ones. Among the open models, only Qwen3.8-27B reasons in Swedish, in 95.8\% of its traces; its gap with reasoning, 2.7 points, is the largest, but on the items it answers validly in both languages it is close to that of the other models, as Appendix~\ref{sec:analysis-reasoning} shows. The comparison is not controlled: the rewritten traces are a second-hand copy of the teacher's reasoning made by GPT-OSS-120B, the LFM2.5-1.2B run on them started from another release of the base checkpoint and had a 10,240-token limit, and the EuroLLM run on them stopped early, as Appendices~\ref{sec:appendix-eval}, \ref{sec:appendix-training}, and~\ref{sec:appendix-repro} report.

\subsection{When Is English Data Enough?}
\label{sec:analysis-english}

English data needs no translation, so we ask whether English traces alone suffice. They do not for LFM2.5-1.2B: after English-only training, its Swedish accuracy falls to 14.5\%, because it does not finish 53.9\% of its Swedish reasoning outputs within the token limit, with a median of 11,375 generated tokens. Its direct Swedish answers are 99.2\% valid and 27.8\% correct, and Table~\ref{tab:validity} gives the valid share of every student. EuroLLM-1.7B, pretrained with Swedish text, leaves only 12.0\% of its Swedish outputs invalid after English-only training and gains 16.0 points in Swedish; Swedish traces add another 5.1 points. Whether English data suffices thus depends on the base model, although the two students differ in more than their pretraining. Reasoning also costs accuracy when outputs do not finish: the bilingual EuroLLM-1.7B model leaves 16.5\% of its English reasoning outputs unfinished, which makes it less accurate in English with reasoning than without, 41.9\% against 44.7\%.

\paragraph{Same accuracy, different items.} Equal accuracy in both languages does not mean that the same items are solved. The bilingual LFM2.5-1.2B model answers 1,894 of the 9,733 core items (19.5\%) correctly in only one language, 876 only in English and 1,018 only in Swedish; its base model does so for 3,499 items, 2,419 of them only in English. GPT-OSS-20B in its fast mode, which has a similar Swedish accuracy, does so for 2,100 items (21.6\%), and the four open models of 27B or more with reasoning for 793 to 1,150 items (8.1--11.8\%). Table~\ref{tab:benchmark_paired} gives these counts for the open models, and Appendix~\ref{sec:appendix-examples} for every student, with examples.

\FloatBarrier

\section{Conclusion}
\label{sec:conclusion}

We built an item-aligned English--Swedish medical benchmark from ten datasets. A conservative refinement loop with blind judges from three model families checked every translation. On this benchmark, all ten benchmark models are less accurate in Swedish, the two proprietary ones on the unrefined translation: at most 2.7 points with reasoning, and 10.5 to 12.4 points for the 3--4B models that answer directly. Fine-tuning 1--2B students on teacher reasoning traces for Swedish or bilingual training questions closes the gap: our 1.2B student rises from 23.5\% to 52.0\% in Swedish, and over three training seeds it is 1.0 to 1.3 points above a 20B open model in its fast mode without reasoning, a lead within the statistical uncertainty, with a sixth of its weight memory, but below that model with reasoning and below every open model of 27B or more. For this student, the gains reach test sets with 5 to 20 training questions, such as MMLU, but hardly the 840 hard items, and in our uncontrolled comparison, students trained to reason in English are more accurate than those trained on reasoning rewritten into Swedish. The best students also gain 28 to 44 points on two public sets of questions written in Swedish. Building the students needs only English data, a translation step, and open teacher models. We release the benchmark, the history of every edit, and the code.

\section*{Limitations}
\label{sec:limitations}

The training translations were not refined. MedMCQA gives 85.2\% of the bilingual reasoning-trace examples, four datasets have 5 training questions each and MedExQA has 20, and the MedBullets training file duplicates MedQA. Weighting each dataset by its share of the training data instead of the test data lifts the bilingual LFM2.5-1.2B model to 62.8\% in Swedish, as Appendix~\ref{sec:appendix-results} shows, so the overall score is held down by datasets the students never trained on. A split with training data from every dataset would show the value of dataset-specific Swedish supervision.

All Swedish text in our benchmark and training data is machine-translated: GPT-5.4 translated the test set, which it then answers itself, and GPT-OSS-120B translated the training set, so the students are trained and tested on translated Swedish, which can differ from native text~\citep{artetxe2020translation}. The students also gain on two public sets written in Swedish, but these sets have no English counterparts and we tested only the students and their base models on them, so they measure neither the gap nor the comparison with larger models. Translated exam questions are not Swedish clinical text; units, guidelines, and clinical practice may differ from Swedish care. The public English test sets may also be part of the pretraining data of the benchmark models, which would favor English; between our training and test splits, 11 core items have a near duplicate, as Table~\ref{tab:overlap} shows, and removing them changes the overall accuracy of the best students and their base models by at most 0.02 points. Multiple-choice accuracy does not measure the safety of free-text answers. We repeated only the two best LFM2.5-1.2B conditions with two further training seeds; with them, the lead of the best student over the 20B model in its fast mode is 1.0 to 1.3 points and not significant after the Holm correction for any seed, and every other condition is a single run. The best student is also chosen on the test items, so the paired intervals measure the sampling of test items, not the selection. The conditions also differ in more than one factor: the bilingual data has twice the examples of the monolingual data, the trace data about four times as many examples as the answer data, so the two supervision types are compared as training recipes, and two EuroLLM trace runs stopped early. We resumed every bilingual trace run once after its first early stop; the EuroLLM-1.7B run then stopped again after 1.23 epochs, so it passed over its data fewer times than the two-epoch runs, as Appendices~\ref{sec:appendix-training} and~\ref{sec:appendix-seeds} report. Our cost comparison estimates compute from parameter counts and token counts and memory from the size of the released weights; we did not measure latency, throughput, or energy, which depend on the hardware and the serving setup.

Our compute budget did not allow us to test the training choices on larger students. We are now studying whether the gains hold or grow with larger models and whether adding other training techniques on top of distillation, such as reinforcement learning with verifiable rewards~\citep{guo2025deepseek}, further improves performance.

Every judgment of translation quality in this paper comes from LLMs: the judges of the loop and the Claude Opus 5.5 raters of the blinded review. A review by Swedish-speaking clinicians is in progress and will be added in a later version.

\section*{Ethical Considerations}
\label{sec:ethics}

\paragraph{Intended use.} The benchmark and the students are research artifacts for measuring and improving Swedish medical question answering. They are not medical devices and must not be used for clinical decisions: a correct answer to an exam question does not make a model safe in clinical practice.

\paragraph{Data and licenses.} All test and training items come from public medical question-answering datasets. Their clinical cases are exam vignettes, not patient records, and we collected no personal data. Each released item keeps the license of its source dataset, the term base keeps the terms of its sources, which Appendix~\ref{sec:appendix-refinement} lists, and each student follows the license of its base model. Section~\ref{sec:adaptation-results} and Appendix~\ref{sec:appendix-cost} report the estimated compute of every step.

\paragraph{Use of AI assistants.} We used AI assistants to organize the repository and the code for the release and to help edit the text of the paper.

\ifpreprint
\section*{Acknowledgments}
This work was supported by the Carl Bennet AB PhD Program. Computational resources were provided by the National Academic Infrastructure for Supercomputing in Sweden (NAISS), funded by the Swedish Research Council.
\fi

\bibliography{references}
\bibliographystyle{tmlr}

\clearpage

\appendix
\renewcommand{\topfraction}{0.95}
\renewcommand{\textfraction}{0.05}
\renewcommand{\floatpagefraction}{0.5}

\section*{Appendix Overview}
\label{sec:appendix-overview}
This appendix gives the details behind the two parts of the paper, and each appendix can be read on its own. The code, the data, and the review are at \codeurl, and the trained students at \modelurl.

\begin{description}[leftmargin=0pt, itemsep=2pt, font=\normalfont\bfseries, style=unboxed]
    \item[\ref{sec:appendix-refinement}\quad Refinement Loop Specification.] The steps, routing classes, term base, gate rules, and acceptance rule of the refinement loop of Section~\ref{sec:refinement}, and the release layers.
    \item[\ref{sec:appendix-translation-quality}\quad Translation Quality: Measures and Review Protocol.] The accepted edits, the reliability checks of the judges, and the protocol and results of the blinded review of Section~\ref{sec:translation-results}.
    \item[\ref{sec:appendix-categories}\quad Results by Refinement Category.] Accuracy and the English--Swedish gap on clean, mechanically fixed, refined, and disputed items, per model group, the gap of every open configuration and of the best students, and checks of whether translation errors explain the gap, for Section~\ref{sec:benchmark-results}.
    \item[\ref{sec:appendix-benchmark}\quad Benchmark Details.] Datasets, models, evaluation settings, full and per-dataset results, paired statistics, generated tokens, the hard subset, train--test overlap, and the reasoning-language analysis, for Sections~\ref{sec:benchmark-results} and~\ref{sec:analysis-language}.
    \item[\ref{sec:appendix-hard}\quad What Makes a Question Hard.] The hard subset compared with the rest of the test set at a matched dataset mix, question types and specialties, accuracy and the English--Swedish gap by question type, and the students.
    \item[\ref{sec:appendix-distillation}\quad Distillation Data.] How the training data of Section~\ref{sec:adaptation} is built, with all counts.
    \item[\ref{sec:appendix-training}\quad Training and Model Selection.] Training settings, the final runs, and the choice of the training setup.
    \item[\ref{sec:appendix-results}\quad Additional Adaptation Results.] All student conditions, decoding modes, reasoning languages, paired statistics, variation between training seeds, the hard subset, and per-dataset results for Sections~\ref{sec:adaptation-results} and~\ref{sec:analysis}.
    \item[\ref{sec:appendix-native}\quad Native Swedish Exam Sets.] How the two native Swedish sets of Section~\ref{sec:native} are prepared, every student on them, and paired tests.
    \item[\ref{sec:appendix-examples}\quad Reasoning Traces: Examples.] Items that a student solves in only one language, with its reasoning in both.
    \item[\ref{sec:appendix-cost}\quad Compute Cost.] Inference compute, memory, and API prices of all models, and the decoding length of the students.
    \item[\ref{sec:appendix-repro}\quad Reproducibility.] What we release and all prompts.
\end{description}

\section{Refinement Loop Specification}
\label{sec:appendix-refinement}

This appendix specifies the refinement loop of Section~\ref{sec:refinement}. We release the code, the configuration, and the prompts with the paper.

\begin{figure}[htb]
    \centering
    \includegraphics[width=\textwidth]{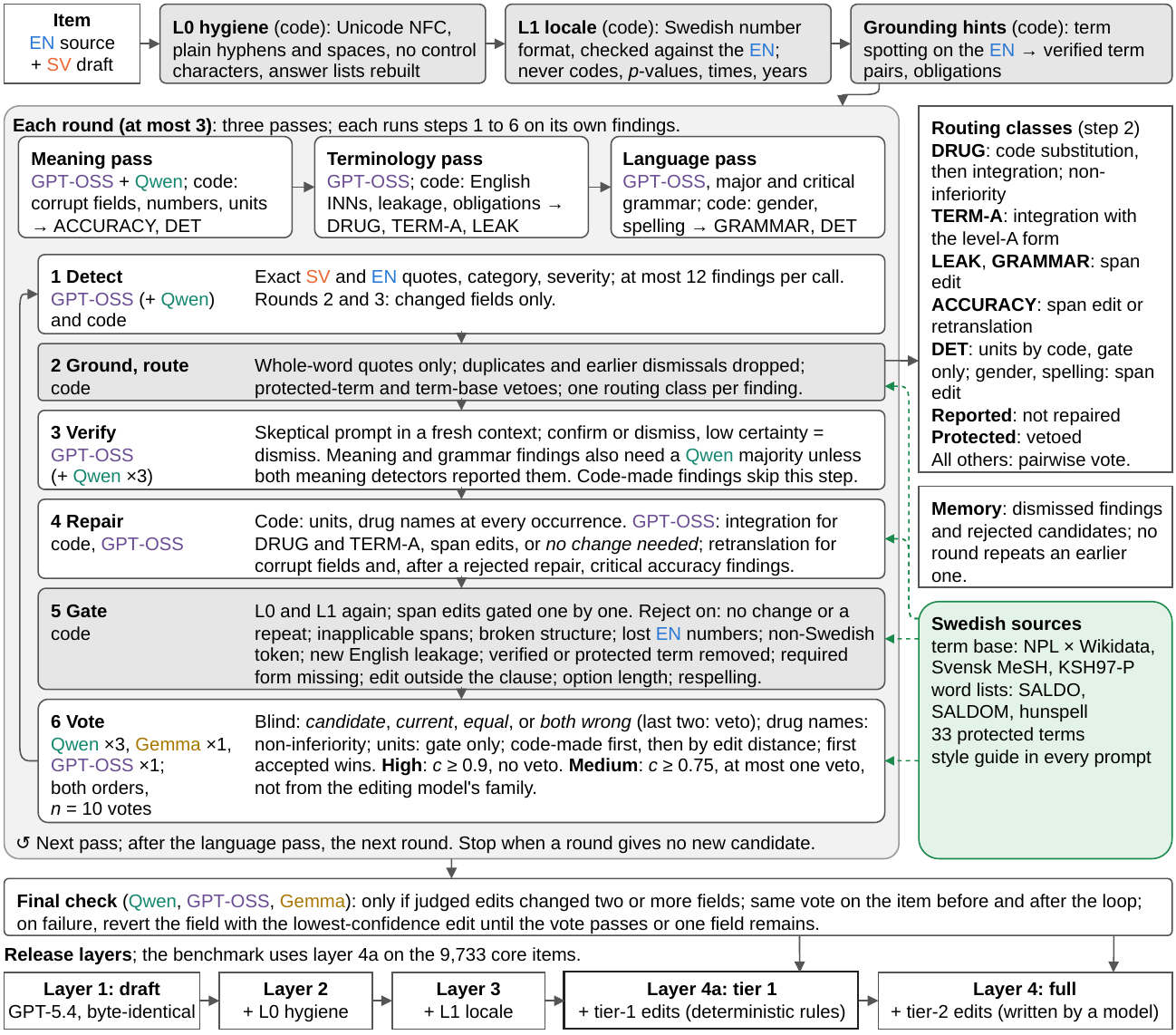}
    \caption{The refinement loop in full. Code layers L0 and L1 normalize the draft, and term spotting gives verified term pairs. Each round runs a meaning, a terminology, and a language pass, and each pass runs steps 1 to 6 on its own findings. Step 2 routes every finding into one class, which fixes its repair and its vote. The loop stops after three rounds or when a round gives no new candidate. A final check reverts edits if the whole item loses the vote. Accepted edits form tier~1, made by deterministic rules, and tier~2, written by a model. The benchmark uses release layer 4a. Gray boxes are code, and dashed arrows show where the Swedish sources are used.}
    \label{fig:refinement_loop_full}
\end{figure}

\paragraph{Steps and routing.} Figure~\ref{fig:refinement_loop_full} shows the loop for one item, and Table~\ref{tab:refine_routes} shows how step~2 routes each grounded finding. Each pass has its own accuracy-, terminology-, or grammar-only prompt. In the meaning pass, GPT-OSS and Qwen share the prompt, check long items sentence by sentence, and merge their findings with a support flag. Code checks the secondary units $^\circ$F, lb, and ft, adds obligations in round~1 only, and checks noun gender with SALDOM. Each detect prompt receives the verified term pairs, the protected terms, and the dismissed findings, and asks for MQM-style categories~\citep{lommel2014multidimensional}. Rounds~2 and~3 re-detect the changed fields, so an error introduced by a fix can be caught. Term spotting takes the longest match first, the term-base veto is concept-aware, and verification sees planted decoys and re-asks once for missing decisions. Code substitutes drug names with their inflection. The worker writes one candidate per field, or one per finding after a multi-finding candidate fails, and after a rejected repair it retranslates only for critical accuracy findings in options or in fields of at most 60 words.

\begin{table}[htb]
    \centering
    \small
    \begin{tabular}{>{\raggedright\arraybackslash}p{0.12\textwidth}>{\raggedright\arraybackslash}p{0.33\textwidth}>{\raggedright\arraybackslash}p{0.25\textwidth}>{\raggedright\arraybackslash}p{0.18\textwidth}}
        \toprule
        \textbf{Class} & \textbf{When} & \textbf{Repair} & \textbf{Acceptance} \\
        \midrule
        DRUG & English INN in the Swedish whose NPL or Svensk MeSH name differs only by transliteration (c/k, ph/f, th/t, y/i, final -e) & code substitution, then integration & non-inferiority \\
        TERM-A & single-sense level-A form exists and the Swedish span is not valid Swedish (or an obligation's English term is left in the Swedish) & integration repair with the form as constraint & pairwise \\
        LEAK & English left in the Swedish, no term-base form & span edit & pairwise \\
        ACCURACY & mistranslation, omission, addition & span edit (or retranslation) & pairwise \\
        GRAMMAR & major or critical model grammar finding & span edit & pairwise \\
        DET & missing secondary unit; noun gender; spelling & units: code; others: span edit & units: gate only; others: pairwise \\
        Reported & valid Swedish without term-base evidence; minor fluency; source issues; number format & none & -- \\
        Protected & span is a protected term & vetoed & -- \\
        \bottomrule
    \end{tabular}
    \caption{Routing classes of grounded findings. Pairwise: the vote of Equation~\ref{eq:confidence} on the question which version is better; non-inferiority: the same vote on the question whether the candidate introduces an error.}
    \label{tab:refine_routes}
\end{table}

\paragraph{Mechanical layers.} Two deterministic, logged, and reversible layers run before the loop. \emph{L0} applies Unicode NFC and removes control characters. It replaces the special hyphens U+2010 (hyphen) and U+2011 (non-breaking hyphen) with the plain hyphen, and the special spaces U+00A0 (no-break space), U+2009 (thin space), and U+202F (narrow no-break space) with a plain space, so every later check sees one form. A field with a NUL character is marked corrupt and must be retranslated. L0 also checks the English MedXpertQA answer list against the options, removes any Swedish list, and rebuilds it (\textit{Svarsalternativ}) from the final Swedish options in the English format, so both languages show the options twice. \emph{L1} applies the Swedish number style. Swedish groups thousands with a space and uses a comma as the decimal mark, so English \textit{1,234} becomes Swedish \textit{1~234}, and English \textit{0.5} becomes Swedish \textit{0,5}. L1 works conservatively, with the English text as reference. A number token that appears identically in the English field gets the Swedish thousands space or decimal comma. Other numbers get a decimal comma only next to a unit, and a thousands space only with five or more digits in an unambiguous context. L1 never touches codes (\textit{E11.9}), chemical names (\textit{2,3-BPG}), versions, lists, $p$-values, times, or years. Both layers run again on every candidate, because model edits re-introduce English typography.

\paragraph{Models and decoding.} The three models run with vLLM~0.28.0~\citep{kwon2023efficient}. The context is 32,768 tokens, and the output limit is 16,384 tokens. GPT-OSS-120B (\texttt{openai/gpt-oss-120b}) decodes greedily with reasoning effort \emph{medium}. Qwen3.8-27B (\texttt{Qwen/Qwen3.8-27B}) reasons at its default effort, \emph{xhigh}, with temperature 1.0, top-$p$ 0.95, and top-$k$ 20. Gemma-4-31B-it (\texttt{google/gemma-4-31B-it}) reasons in its thinking mode with temperature 1.0, top-$p$ 0.95, and top-$k$ 64. Each model reasons freely and then answers in a fixed JSON schema. An invalid, truncated, or empty GPT-OSS output is retried once at temperature 0.6 and then handed to Qwen; if Qwen also fails, the step is recorded as failed. A failed judge call casts no vote, and no failure stops an item.

\paragraph{Term base and lexicons.} Table~\ref{tab:termbase} lists the sources. The term base joins the national medicinal products register NPL\footnote{L\"akemedelsverket, \url{https://catalog.lakemedelsverket.se/store/1/resource/9}.} with Wikidata~\citep{vrandevcic2014wikidata} on the ATC code: Wikidata gives the English INN, and NPL the Swedish substance name (\textit{vancomycin}~$\rightarrow$~\textit{vankomycin}). Diseases, anatomy, symptoms, procedures, organisms, and chemicals come from Svensk MeSH (April 2026 release)\footnote{Karolinska Institutet University Library, \url{https://mesh.kib.ki.se}.}, with the concept structure of NLM MeSH 2026\footnote{\url{https://www.nlm.nih.gov/mesh/}.}, and diagnoses also from KSH97-P, the Swedish primary-care version of ICD-10 (Socialstyrelsen). Matching is inflection-aware and handles compounds. Word validity rests on the SALDO and SALDOM lexicons~\citep{borin2013saldo}, extended by the other lexicons in Table~\ref{tab:termbase}. On the test drafts, 98.1\% of items have at least one term-base hit, and 107,066 English--Swedish term pairs are verified in the draft. Level~A holds only preferred terms of trusted sources: the Swedish substance name of single-substance ATC codes (1,373 ATC codes from the NPL--Wikidata join), Swedish MeSH preferred terms and their synonyms within the preferred concept, KSH97-P labels, and NPL lexicon pairs. Terms of other MeSH concepts, trade names, and Wikidata's own Swedish labels are level~B. An entry is single-sense unless its English term is an acronym, a general English word that is not a cognate, an ambiguous list, or has several senses in one source. Only single-sense level-A entries can force a substitution or veto a finding. Inverted MeSH headings (\textit{x, y}) are never replacements, and for drugs NPL takes priority over MeSH, so a DRUG finding uses the Svensk MeSH name only when NPL has none; its English token must also occur in the English field. Obligations, that is, missing level-A terms, exist only for drug, supplementary, and cognate entries with a Swedish form of at least five characters: 5,419 on the test drafts. An obligation is repaired only if its English term is left verbatim in the Swedish field. Of the level-A single-sense hits in the test drafts, 95.2\% are verified in the Swedish draft.

\begin{table}[htb]
    \centering
    \small
    \begin{tabular}{>{\raggedright\arraybackslash}p{0.23\textwidth}>{\raggedright\arraybackslash}p{0.31\textwidth}rr>{\raggedright\arraybackslash}p{0.16\textwidth}}
        \toprule
        \textbf{Source} & \textbf{Provider and license} & \textbf{A} & \textbf{B} & \textbf{Content} \\
        \midrule
        NPL $\times$ Wikidata ATC & L\"akemedelsverket, open government data; Wikidata, CC0 & 2,031 & 444 & drug names \\
        Wikidata Swedish labels & Wikidata, CC0 & -- & 764 & drug names \\
        Svensk MeSH, April 2026 & KI University Library, CC0 with attribution; NLM MeSH terms and conditions & 61,460 & 61,102 & medical concepts \\
        KSH97-P & Socialstyrelsen, open & 634 & -- & diagnoses \\
        NPL lexicon & L\"akemedelsverket, open & 1,062 & -- & dose forms, routes, units \\
        Supplementary clinical forms & -- & 8 & -- & diagnoses \\
        \midrule
        \textbf{Term base} & & \textbf{65,195} & \textbf{62,310} & 127,505 rows \\
        \midrule
        \multicolumn{5}{l}{\textit{Lexicons (level V and English word list)}} \\
        SALDO + SALDOM & Spr\aa kbanken Text, CC BY 4.0 & \multicolumn{2}{r}{867,214 forms} & Swedish words \\
        hunspell sv\_SE & LGPL/GPL & \multicolumn{2}{r}{+76,827 forms} & Swedish words \\
        Socialstyrelsen term bank; KV\r{A} & Socialstyrelsen, open & \multicolumn{2}{r}{+283; +2,942 forms} & Swedish words \\
        Swedish lexicon (total) & & \multicolumn{2}{r}{974,793 forms} & incl.\ term-base forms \\
        FreeDict/WikDict EN--SV & CC BY-SA & \multicolumn{2}{r}{22,633 words} & English headwords \\
        Accepted abbreviations and Latin & -- & \multicolumn{2}{r}{814 tokens} & abbreviations, Latin \\
        \bottomrule
    \end{tabular}
    \caption{Sources of the term base and the lexicons, built on 2026-10-06. A: may force a substitution or veto a finding; B: hint only; V: word validity only. Supplementary clinical forms: the established Swedish forms of type~1 and type~2 diabetes (mellitus) and of metabolic and respiratory acidosis and alkalosis, such as \textit{typ 2-diabetes} and \textit{metabol acidos}, which the other sources give in only one form. Accepted abbreviations and Latin: clinical abbreviations, Latin terms, and international terms such as \textit{stroke}, accepted as valid Swedish tokens. Citations and links for the other sources are given in the text.}
    \label{tab:termbase}
\end{table}

\paragraph{Protected terms and style guide.} The 33 \emph{protected terms} are established Swedish clinical expressions that a model is likely to replace with a word-for-word rendering of the English, such as \textit{hereditet}, \textit{persisterande ductus arteriosus}, \textit{p\aa verkad}, \textit{resistens}, and \textit{lungr\"ontgen}. The loop vetoes findings against them, rejects candidates that remove them, and shows the list to the judges. Every prompt also contains a one-page style guide, which Appendix~\ref{sec:appendix-repro} prints.

\paragraph{Gate rules.} The gate applies L0 and L1 to the candidate, checks each span edit on its own, drops the failing edits, and checks the rest again. It rejects a span edit or candidate if any rule fires: (1)~it changes nothing or repeats a candidate rejected earlier for the field; (2)~its span edits cannot be applied: an old span is not found as whole words, a multi-word span occurs more than once, or spans overlap (a single-word span is applied at every occurrence); (3)~it breaks the structure with an empty field, a line break, or an option label; (4)~more numbers of the English field are missing than in the current text; (5)~it adds a token that is not in the Swedish lexicon, not numeric, not in the English field, not an accepted abbreviation or Latin term, and not an all-caps abbreviation; (6)~it adds a new English leakage candidate; (7)~it removes a verified Swedish term (by lemma) or a protected term (with its inflected and participle forms); (8)~a drug or TERM-A edit lacks the required authoritative form; (9)~a model edit changes text beyond the clause around the finding plus three tokens; (10)~an option's length relative to the median of the other options leaves $[0.33, 3]$ when it was inside before; (11)~a model edit replaces a valid Swedish word by a form at most two characters away, the difference is not confined to the last three characters (inflection), the old word is not in the English source, and the new form is not better attested in the lexicon, where a listed word form beats a word that is only analyzable as a compound.

\paragraph{Acceptance policy.} A full vote has ten votes, three Qwen samples and one each from GPT-OSS and Gemma in both presentation orders, and $c=n_+/10$; it has high confidence if $c\geq0.9$ with no veto and medium confidence if $c\geq0.75$ with at most one veto, which must not come from the family that wrote the edit; these thresholds were fixed before the run. A pairwise vote is \emph{candidate}, \emph{current}, \emph{equal}, or \emph{both wrong}; low certainty counts as \emph{equal}, and \emph{current} and \emph{both wrong} are vetoes. For drug names, the vote asks ``does version B introduce a change of meaning or a Swedish grammar error that version A does not have?'', with neutral labels and both orders, and a yes is a veto. The judges are told that a drug name comes from NPL or Svensk MeSH and is correct by definition, and that a TERM-A form is a Swedish term in the term base, as a reference and not a requirement. Voting stops once no outcome above \emph{rejected} is possible, except on a fixed 10\% sample of candidates that collects all votes. Per field, code-made candidates come first and the rest follow by edit distance; the first accepted candidate wins. Figure~\ref{fig:refinement_loop_full} shows the final check.

\paragraph{Release layers.}
\label{sec:appendix-layers}
The release contains five layers of the Swedish test set in the draft's format, so the effect of each step can be measured: layer~1 is the GPT-5.4 draft, byte-identical; layer~2 adds L0; layer~3 adds L1; layer~4a adds the code-made tier-1 edits (inserted units and drug-name substitutions); and layer~4 adds all accepted edits. An edit history records, for every item, each mechanical change, finding, routing class, candidate, gate result, vote, confidence level, tier, and status, so every released character can be traced. The benchmark uses layer~4a on the core items, which Appendix~\ref{sec:appendix-categories} defines.

\section{Translation Quality: Measures and Review Protocol}
\label{sec:appendix-translation-quality}

\paragraph{Accepted edits.} Table~\ref{tab:refine_edits_full} gives the accepted edits by class, tier, and confidence level. Besides the item-level review below, Claude Opus 5.5 with \textit{xhigh} reasoning effort judged 129 accepted edits one by one, sampled by edit class. It saw each edit blind next to the text just before it and chose the better version or \emph{equal}. This check gives the share of better and worse edits per class in Table~\ref{tab:refine_edits} and Table~\ref{tab:refine_edits_full}. The release decision rests on the item-level review.

\begin{table}[htb]
    \centering
    \small
\begin{tabular}{clrrrrrrr}
\toprule
& & & \multicolumn{2}{c}{\textbf{Confidence}} & \multicolumn{4}{c}{\textbf{Review}} \\
\cmidrule(lr){4-5}\cmidrule(lr){6-9}
\textbf{Tier} & \textbf{Edit class} & \textbf{Edits} & \textbf{High} & \textbf{Medium} & $\boldsymbol{n}$ & \textbf{Better} & \textbf{Equal} & \textbf{Worse} \\
\midrule
1 & Drug name $\rightarrow$ Swedish INN (code) & 156 & 156 & 0 & 14 & 100 & 0 & 0 \\
1 & Missing secondary unit (code) & 138 & \multicolumn{2}{c}{deterministic} & 6 & 50 & 50 & 0 \\
& \textit{Tier 1 total (weighted)} & \textit{294} & 156 & 0 & \textit{20} & \textit{77} & \textit{23} & \textit{0} \\
\midrule
2 & Grammar & 377 & 271 & 106 & 30 & 47 & 43 & 10 \\
2 & English left in the Swedish & 209 & 136 & 73 & 22 & 50 & 41 & 9 \\
2 & Meaning (accuracy) & 190 & 141 & 49 & 22 & 45 & 36 & 18 \\
2 & Term-base terminology & 159 & 140 & 19 & 22 & 64 & 5 & 32 \\
2 & Drug name with adjusted wording & 16 & 15 & 1 & 6 & 100 & 0 & 0 \\
2 & Spelling or noun gender & 7 & 4 & 3 & 7 & 100 & 0 & 0 \\
& \textit{Tier 2 total (weighted)} & \textit{958} & 707 & 251 & \textit{109} & \textit{51} & \textit{34} & \textit{15} \\
\bottomrule
\end{tabular}

    \caption{Accepted edits of the refinement loop by class, tier, and confidence level, with the result of the blinded per-edit check by Claude Opus 5.5 ($n$ reviewed edits; \% better, equal, or worse than the text before the edit). Totals are weighted by the number of edits per class.}
    \label{tab:refine_edits_full}
\end{table}

\paragraph{Outcome per dataset.} Table~\ref{tab:refine_status_dataset} gives the final status of every item per dataset. L0 changed the text of 1,979 items, 1,878 only through the rebuilt MedXpertQA answer list and 101 through typography or control characters, and L1 changed 1,758 items (6,849 Swedish number conventions). No item failed. Of the 1,063 refined items, 212 have only tier-1 edits and 851 have at least one tier-2 edit. Of the 624 disputed items, 10 keep a finding of critical severity.

\begin{table}[htb]
    \centering
    \small
\begin{tabular}{lrrrrrr}
\toprule
\textbf{Dataset} & \textbf{Items} & \textbf{Clean} & \textbf{Mechanical only} & \textbf{Refined} & \textbf{Refined (\%)} & \textbf{Disputed} \\
\midrule
AfriMedQA & 174 & 119 & 41 & 6 & 3.4 & 8 \\
MedBullets & 308 & 63 & 178 & 42 & 13.6 & 25 \\
MedExQA & 935 & 865 & 10 & 47 & 5.0 & 13 \\
MedMCQA & 2,816 & 2,447 & 63 & 156 & 5.5 & 150 \\
MedQA & 1,273 & 668 & 285 & 229 & 18.0 & 91 \\
MedXpertQA-R & 1,861 & 700 & 695 & 271 & 14.6 & 195 \\
MedXpertQA-U & 589 & 291 & 176 & 77 & 13.1 & 45 \\
MMLU & 1,089 & 928 & 51 & 81 & 7.4 & 29 \\
MMLU-Pro & 818 & 674 & 48 & 61 & 7.5 & 35 \\
PubMedQA & 500 & 213 & 161 & 93 & 18.6 & 33 \\
\midrule
\textbf{Total} & 10,363 & 6,968 & 1,708 & 1,063 & 10.3 & 624 \\
\bottomrule
\end{tabular}

    \caption{Final status of the test items per dataset. Mechanical only: only the deterministic layers changed the text. Disputed: a confirmed finding remains for which no edit was accepted.}
    \label{tab:refine_status_dataset}
\end{table}

\paragraph{Judge probes.} Table~\ref{tab:refine_probes} shows how often each judge picks the correct version of a fixed probe pair in both orders. All three judges answer the 250 probes with one injected error and the 50 identical pairs almost perfectly. The 40 labeled grammar edits are subtle: one version is only slightly more idiomatic. Here the judges mostly answer \emph{equal}, which keeps the current text and is the safe choice under our rule.

\begin{table}[htb]
    \centering
    \small
\begin{tabular}{lrrrr}
\toprule
\textbf{Probe} & $\boldsymbol{n}$ & \textbf{GPT-OSS} & \textbf{Qwen} & \textbf{Gemma} \\
\midrule
Negation removed & 50 & 96 & 100 & 100 \\
Number changed & 50 & 100 & 100 & 100 \\
English drug name & 50 & 98 & 99 & 100 \\
Laterality swapped & 50 & 98 & 95 & 100 \\
Agreement broken & 50 & 98 & 94 & 98 \\
Identical texts (``equal'') & 50 & 96 & 99 & 100 \\
Labeled grammar edits & 40 & 20 & 27 & 38 \\
\bottomrule
\end{tabular}

    \caption{Judge probes: share (\%) of probe pairs judged correctly in both presentation orders (Qwen: all three samples). Injected errors and identical pairs have a known answer; the labeled grammar edits are subtle stylistic cases.}
    \label{tab:refine_probes}
\end{table}

\paragraph{Other reliability measures.} (i)~\emph{Decoys} are plausible false findings that the loop plants among the real findings before verification. For a verified term pair, a decoy claims that the Swedish term is a word-for-word rendering of the English, names a different concept, or is outdated; for a number that matches the English source, it claims that the number is wrong. A decoy that verification confirms counts as a verifier error. Verification confirmed 0.74\% of the 3,803 decoys and none of the 2,824 false meaning claims, which say that a correct number is wrong. (ii)~\emph{Position consistency}: a judge gives the same verdict in both orders for 79.1\% (Qwen), 85.1\% (GPT-OSS), and 83.9\% (Gemma) of the candidates. (iii)~\emph{Agreement}: on the 1,489 candidates with all votes collected, Cohen's $\kappa$ between Qwen and GPT-OSS is 0.19 with the current text first and 0.22 with the candidate first. Single judges agree only weakly on these borderline cases, so an edit needs ten votes from three families rather than one judge's verdict. (iv)~\emph{Answer key}: 0.80\% of keyed options and 0.74\% of incorrect options were edited ($\chi^2$ $p=0.52$). The judges rejected 1,646 candidates in total. The run used 100,447 model calls and about $19\times10^{18}$ FLOPs.

\subsection{Review Protocol}
\label{sec:appendix-expert}

The protocol below was carried out by Claude Opus 5.5, as the paragraph on raters describes. A script generates the review sheets from the refinement output as XLSX and CSV files. A separate file, \emph{the key}, is kept apart from the sheets and never shown to the raters. For each pairwise item, it records which version is the earlier text, the item's stratum, and its inclusion probability.

\begin{enumerate}[leftmargin=*]
    \item \textbf{Pairwise sheet.} 200 \emph{refined} items in three strata, at least 40 each: tier-1 items (only code-made edits), and tier-2 items at high and at medium confidence, where an item's level is the lowest level of its accepted edits. The rater sees the English item and two Swedish versions in random order. They answer \emph{A better}, \emph{B better}, \emph{equal}, or \emph{both wrong}, and say whether the meaning changed. For tier-1 items, the versions are the draft and the tier-1 text; for tier-2 items, the tier-1 text and the full refined text. Both versions always have the same layers L0 and L1.
    \item \textbf{Findings sheet.} The accepted findings of the sampled items, and 100 dismissed, vetoed, or judge-rejected findings. The rater labels whether each flagged span is a real error.
    \item \textbf{Control sheet.} 100 \emph{clean} and 50 \emph{disputed} items. The rater says whether the Swedish contains an error, quotes it, and gives its severity (minor, major, critical).
\end{enumerate}

\paragraph{Endpoints.} The primary endpoint, per tier, is the harm rate: the share of pairwise items where the rater prefers the reference version, which is the draft for tier~1 and the tier-1 text for tier~2. We report it with a 95\% Wilson interval and an inverse-probability-weighted estimate over the strata. A tier succeeds if the upper Wilson bound is below 10\%. The release uses the full refined layer if tier~2 succeeds and the tier-1 layer otherwise. Secondary endpoints are the win rate (one-sided sign test on non-tie pairs), the harm rate per confidence level, the meaning-changed rate, the precision of accepted findings, the share of real errors among dismissed, vetoed, or judge-rejected findings, and the error rate in \emph{clean} and \emph{disputed} items. We report all release layers and state which layer the Swedish results use.

\paragraph{Raters.} The raters were not medical experts but Claude Opus 5.5 with \textit{xhigh} reasoning effort, a model family that the refinement loop does not use. The raters saw only the sheets, the rating guidelines, and the Swedish style guide, never the key or the loop's records. Two separate runs of the model rated each pairwise item. They agreed on 170 of 200 items (85\%, Cohen's $\kappa=0.78$), and a third blind pass settled the disagreements. The findings and control sheets were rated once.

\paragraph{Results.} Table~\ref{tab:review} gives the primary endpoint. Of the harmful tier-2 edits, 15 were minor and 3 major; the improvements include 7 major and 1 critical fix. Of the 18 worse tier-2 items, 10 had high confidence and 8 medium confidence. Medium-confidence edits are rarer but harmful twice as often (17.4\% against 8.8\%). High-confidence edits do not pass on their own either, with an upper bound of 15.4\%. Accepted findings were real errors in 55\% of cases and dismissed findings in 7\%, so the loop rarely discards a real error. In the control sheet, 9\% of the clean items and 20\% of the disputed items had an error, all of them minor.

\begin{table}[htb]
    \centering
    \small
\begin{tabular}{lrrrrrr}
\toprule
\textbf{Edits} & $\boldsymbol{n}$ & \textbf{Better} & \textbf{Equal} & \textbf{Worse} & \textbf{Both wrong} & \textbf{Harm (\%) [95\% CI]} \\
\midrule
\textbf{Tier 1, code-made} & 40 & 40 & 0 & 0 & 0 & \textbf{0.0} [0.0, 8.8] \\
Tier 2, by a model, high conf. & 114 & 61 & 40 & 10 & 3 & 8.8 [4.8, 15.4] \\
Tier 2, by a model, medium conf. & 46 & 16 & 22 & 8 & 0 & 17.4 [9.1, 30.7] \\
\midrule
\textbf{Tier 2, all} & 160 & 77 & 62 & 18 & 3 & \textbf{11.3} [7.2, 17.1] \\
\bottomrule
\end{tabular}

    \caption{Blinded review by LLM raters: pairwise comparison of each edited item with the text before the edit, by tier and confidence level. Harm: share of items where the earlier text was preferred, with a 95\% Wilson interval. A tier passes if the upper bound is below 10\%.}
    \label{tab:review}
\end{table}

\section{Results by Refinement Category}
\label{sec:appendix-categories}

\paragraph{Five categories.} The refinement gives every test item one final status, which Appendix~\ref{sec:appendix-translation-quality} describes. We group the items into five categories:
\begin{itemize}[leftmargin=*,itemsep=1pt]
    \item \textbf{Clean} (6,968 items): the loop found no error to fix, and the text is unchanged.
    \item \textbf{Mechanical fixes only} (1,708): only typography and Swedish number style changed.
    \item \textbf{Refined by code, tier~1} (212): at least one code-made edit (a Swedish drug name or a missing unit), and no model-written edit.
    \item \textbf{Refined by a model, tier~2} (845): at least one model-written edit was accepted. The evaluated layer has only code-made edits, so these items keep their tier-1 text there.
    \item \textbf{Disputed} (624): a confirmed finding remains for which no edit was accepted.
\end{itemize}

\paragraph{The main text uses the core items.} The Swedish test set is layer~4a of Appendix~\ref{sec:appendix-layers}, the refinement's safe release layer: it has the mechanical fixes and the code-made edits, but no model-written edit. As the last step of the refinement, we set aside the items that remain disputed and six items with a corrupt draft field or an unresolved critical finding, and we restrict English to the same items. These six have an accepted model-written edit, so 851 items have one in all, but they are counted in none of the five categories. The first four categories form the \emph{core}: 9,733 of the 10,363 items (93.9\%). All benchmark results in the main text and in the other appendices use the core items, so every such number rests on items without an unresolved finding; the 845 tier-2 items enter in their tier-1 text, without their model-written edits. We still run every model on all items, so we can report the disputed items here.

\paragraph{The gap by category.} Table~\ref{tab:categories} gives English and Swedish accuracy and the gap per category, averaged over three groups of configurations. The large open models are Gemma-4-31B, Qwen3.6-35B-A3B, Qwen3.8-27B, GPT-OSS-120B, and GPT-OSS-20B, each with reasoning on and off, and MedGemma-27B (11 configurations). The small open models are Qwen3-4B-Thinking with reasoning on and off and Qwen2.5-3B (3 configurations). The students are all 14 trained students: LFM2.5-1.2B and EuroLLM-1.7B, each trained at answer level on English, Swedish, and bilingual data, and on reasoning traces in English, Swedish, rewritten Swedish, and both languages. The reasoning-trace students decode with reasoning. Accuracy differs between categories because their dataset mix differs. The mechanical-only and disputed items come mostly from the long and hard MedXpertQA and MedQA cases, and the clean items mostly from short knowledge questions. The gap, however, is similar across categories: the large open models lose 3.4 points on clean items, 3.6 on items with a model-written edit, and 3.4 on the disputed items, and the students 3.3 to 4.5 points in every category. The small open models vary more, but their largest gap, 9.6 points, is on the clean items.

\begin{table}[htb]
    \centering
    \small
    \resizebox{\textwidth}{!}{%
\begin{tabular}{lrrrrrrrrrr}
\toprule
& & \multicolumn{3}{c}{\textbf{Large open models}} & \multicolumn{3}{c}{\textbf{Small open models}} & \multicolumn{3}{c}{\textbf{Students}} \\
\cmidrule(lr){3-5}\cmidrule(lr){6-8}\cmidrule(lr){9-11}
\textbf{Category} & $\boldsymbol{n}$ & \textbf{EN} & \textbf{SV} & $\boldsymbol{\Delta}$ \textbf{[95\% CI]} & \textbf{EN} & \textbf{SV} & $\boldsymbol{\Delta}$ \textbf{[95\% CI]} & \textbf{EN} & \textbf{SV} & $\boldsymbol{\Delta}$ \textbf{[95\% CI]} \\
\midrule
Clean & 6,968 & 73.8 & 70.4 & 3.4 {\scriptsize[3.0, 3.7]} & 57.5 & 48.0 & 9.6 {\scriptsize[8.8, 10.3]} & 42.3 & 38.4 & 3.9 {\scriptsize[3.5, 4.2]} \\
Mechanical only & 1,708 & 58.1 & 54.0 & 4.1 {\scriptsize[3.4, 4.7]} & 36.3 & 32.0 & 4.3 {\scriptsize[3.0, 5.6]} & 29.4 & 26.1 & 3.3 {\scriptsize[2.7, 3.9]} \\
Tier 1 (code) & 212 & 77.3 & 72.9 & 4.4 {\scriptsize[2.7, 6.1]} & 50.6 & 43.1 & 7.5 {\scriptsize[3.8, 11.3]} & 43.7 & 39.2 & 4.5 {\scriptsize[2.5, 6.6]} \\
Tier 2 (model)$^\ast$ & 845 & 64.3 & 60.7 & 3.6 {\scriptsize[2.6, 4.5]} & 46.1 & 40.7 & 5.4 {\scriptsize[3.4, 7.3]} & 37.0 & 33.2 & 3.8 {\scriptsize[2.9, 4.8]} \\
\addlinespace[1pt]
\textit{Core (reported)} & 9,733 & 70.3 & 66.8 & 3.5 {\scriptsize[3.3, 3.8]} & 52.7 & 44.4 & 8.3 {\scriptsize[7.7, 8.8]} & 39.6 & 35.8 & 3.8 {\scriptsize[3.5, 4.1]} \\
\midrule
Disputed & 624 & 62.7 & 59.4 & 3.4 {\scriptsize[2.3, 4.4]} & 42.5 & 35.3 & 7.2 {\scriptsize[4.7, 9.5]} & 34.6 & 30.8 & 3.8 {\scriptsize[2.7, 4.9]} \\
\bottomrule
\end{tabular}
}
    \caption{Accuracy (\%) and gap $\Delta=\mathrm{EN}-\mathrm{SV}$ by refinement category, with 95\% paired-bootstrap intervals for the gap. Each value is the mean over the configurations of a group: large open models with at least 20B parameters (11 configurations), small open models (3), and trained students (14). Core: the first four categories, on which all other results are reported. $^\ast$In the evaluated layer these items keep their tier-1 text.}
    \label{tab:categories}
\end{table}

Table~\ref{tab:categories_by_model} gives the gap per category for single configurations. On the disputed items, no open configuration has a gap more than 1.2 points above its gap on the core, or 1.8 points when the categories are reweighted as below. Relative to each configuration's gap on the core, the gap moves most on the 212 tier-1 items, by up to 5.5 points, as expected for so few items, and for Qwen2.5-3B, which loses 14.7 points on the clean items but only 4.9 on the mechanical-only items.

\begin{table}[htb]
    \centering
    \small
\begin{tabular}{lrrrrrr}
\toprule
\textbf{Model} & \textbf{Clean} & \textbf{Mech.} & \textbf{Tier 1} & \textbf{Tier 2$^\ast$} & \textbf{\textit{Core}} & \textbf{Disputed} \\
$n$ & {\scriptsize 6,968} & {\scriptsize 1,708} & {\scriptsize 212} & {\scriptsize 845} & {\scriptsize 9,733} & {\scriptsize 624} \\
\midrule
\multicolumn{7}{l}{\textit{Large open models}} \\
Gemma-4-31B & 0.3 & 2.6 & 0.0 & 0.9 & 0.8 & 0.5 \\
Gemma-4-31B (off) & 2.3 & 4.7 & 8.5 & 4.1 & 3.0 & 4.2 \\
Qwen3.6-35B-A3B & 1.3 & 1.5 & $-$0.5 & 1.9 & 1.3 & 1.0 \\
Qwen3.6-35B-A3B (off) & 3.5 & 5.4 & 5.2 & 1.5 & 3.7 & 2.9 \\
Qwen3.8-27B & 2.3 & 3.2 & 3.3 & 4.0 & 2.7 & 1.4 \\
Qwen3.8-27B (off) & 3.8 & 5.7 & 3.8 & 4.3 & 4.1 & 5.3 \\
GPT-OSS-120B & 1.3 & 1.5 & 5.2 & 0.9 & 1.4 & 2.2 \\
GPT-OSS-120B (off) & 7.4 & 6.4 & 4.7 & 9.3 & 7.4 & 8.0 \\
GPT-OSS-20B & 2.3 & 1.9 & 0.0 & 1.5 & 2.1 & 1.3 \\
GPT-OSS-20B (off) & 8.8 & 6.7 & 13.2 & 7.2 & 8.4 & 7.9 \\
MedGemma-27B & 3.7 & 5.1 & 4.7 & 3.3 & 4.0 & 2.4 \\
\midrule
\multicolumn{7}{l}{\textit{Small open models}} \\
Qwen3-4B-Thinking & 2.1 & 1.4 & 1.4 & 0.6 & 1.8 & 3.0 \\
Qwen3-4B-Thinking (off) & 12.0 & 6.7 & 8.5 & 6.7 & 10.5 & 9.8 \\
Qwen2.5-3B & 14.7 & 4.9 & 12.7 & 8.8 & 12.4 & 8.7 \\
\midrule
\multicolumn{7}{l}{\textit{Students, best Swedish model}} \\
LFM2.5-1.2B, EN+SV traces & $-$1.9 & $-$1.3 & 1.4 & 0.8 & $-$1.5 & $-$0.2 \\
EuroLLM-1.7B, SV traces & 0.1 & $-$0.2 & $-$2.4 & $-$1.9 & $-$0.2 & $-$1.1 \\
\bottomrule
\end{tabular}

    \caption{Gap $\Delta=\mathrm{EN}-\mathrm{SV}$ (points) by refinement category for single configurations: every open benchmark configuration and the best Swedish model of each student, trained on reasoning traces and decoded with reasoning. (off): reasoning off. Mech.: mechanical fixes only. Core: the first four categories. $^\ast$In the evaluated layer these items keep their tier-1 text.}
    \label{tab:categories_by_model}
\end{table}

\paragraph{Reweighted to the core dataset mix.} The categories differ in their dataset mix: model-written edits, for example, concentrate on clinical vignettes and research abstracts. Table~\ref{tab:categories_reweighted} therefore weights the gap of each dataset within a category by that dataset's share of the core items. For the large open models, the reweighted gaps lie between 3.2 and 4.1 points, against 3.5 on the core, and every 95\% interval contains the core gap. For the small open models, the difference between clean and mechanically fixed items shrinks from 5.3 to 3.3 points, so most of it comes from the dataset mix. Because the loop's recall is unknown, these checks show that the gap is not concentrated in the items where the loop found errors; they cannot exclude errors that the loop missed.

\begin{table}[htb]
    \centering
    \small
    \resizebox{\textwidth}{!}{%
\begin{tabular}{lrrrrrrrrr}
\toprule
& & \multicolumn{2}{c}{\textbf{Large open}} & \multicolumn{2}{c}{\textbf{All open}} & \multicolumn{2}{c}{\textbf{Small open}} & \multicolumn{2}{c}{\textbf{Students}} \\
\cmidrule(lr){3-4}\cmidrule(lr){5-6}\cmidrule(lr){7-8}\cmidrule(lr){9-10}
\textbf{Category} & $\boldsymbol{n}$ & $\boldsymbol{\Delta}$ & $\boldsymbol{\Delta}_{\mathrm{rw}}$ \textbf{[95\% CI]} & $\boldsymbol{\Delta}$ & $\boldsymbol{\Delta}_{\mathrm{rw}}$ \textbf{[95\% CI]} & $\boldsymbol{\Delta}$ & $\boldsymbol{\Delta}_{\mathrm{rw}}$ \textbf{[95\% CI]} & $\boldsymbol{\Delta}$ & $\boldsymbol{\Delta}_{\mathrm{rw}}$ \textbf{[95\% CI]} \\
\midrule
Clean & 6,968 & 3.4 & 3.4 {\scriptsize[3.0, 3.7]} & 4.7 & 4.5 {\scriptsize[4.2, 4.8]} & 9.6 & 8.7 {\scriptsize[7.9, 9.4]} & 3.9 & 3.5 {\scriptsize[3.2, 3.9]} \\
Mechanical only & 1,708 & 4.1 & 3.3 {\scriptsize[2.3, 4.2]} & 4.1 & 3.7 {\scriptsize[2.7, 4.8]} & 4.3 & 5.4 {\scriptsize[2.6, 8.2]} & 3.3 & 3.2 {\scriptsize[1.9, 4.7]} \\
Tier 1 (code) & 212 & 4.4 & 4.1 {\scriptsize[2.2, 6.0]} & 5.1 & 4.9 {\scriptsize[3.2, 6.6]} & 7.5 & 7.7 {\scriptsize[3.4, 11.8]} & 4.5 & 3.8 {\scriptsize[1.8, 5.9]} \\
Tier 2 (model)$^\ast$ & 845 & 3.6 & 3.7 {\scriptsize[2.5, 4.8]} & 3.9 & 4.3 {\scriptsize[3.2, 5.4]} & 5.4 & 6.7 {\scriptsize[4.4, 9.0]} & 3.8 & 3.7 {\scriptsize[2.5, 4.9]} \\
\addlinespace[1pt]
\textit{Core (reported)} & 9,733 & 3.5 & 3.5 & 4.5 & 4.5 & 8.3 & 8.3 & 3.8 & 3.8 \\
\midrule
Disputed & 624 & 3.4 & 3.2 {\scriptsize[2.1, 4.4]} & 4.2 & 4.3 {\scriptsize[3.2, 5.5]} & 7.2 & 8.3 {\scriptsize[5.7, 11.0]} & 3.8 & 4.2 {\scriptsize[2.7, 5.8]} \\
\bottomrule
\end{tabular}
}
    \caption{English--Swedish gap (points) by refinement category, as measured ($\Delta$) and reweighted to the dataset mix of the core items ($\Delta_{\mathrm{rw}}$), with 95\% bootstrap intervals that resample items within each category and dataset. Groups as in Table~\ref{tab:categories}; all open: the large and small open models together.}
    \label{tab:categories_reweighted}
\end{table}

\paragraph{Effect of the core rule and of the model-written edits.} On all 10,363 items instead of the core, the gap of a configuration changes by $-0.01$ points on average and by at most 0.32 points (28 configurations). We also evaluate the items whose text differs in layer~4, which adds the model-written edits. With layer~4 instead of layer~4a, the gap on the core items changes by 0.01 points on average and by at most 0.24 points. The model-written edits change the quality of the text in places, as Section~\ref{sec:translation-results} shows, but have a negligible effect on the measured gap.

\section{Benchmark Details}
\label{sec:appendix-benchmark}
\label{sec:appendix-eval}

\paragraph{Datasets.} Table~\ref{tab:dataset_statistics} in Section~\ref{sec:datasets} gives the size of every dataset. The training splits are very unbalanced: MedMCQA alone has 120,765 of the 141,735 questions, and five datasets have 20 or fewer. The authors of the collection~\citep{shao2026medicalagentsbenchcomplexmedicalreasoning} selected as hard the questions that fewer than half of the models they tested answered correctly, sampled this subset per dataset, and screened it for contamination; we use it as released.

\begin{table}[htb]
    \centering
    \small
    \begin{tabular}{llc>{\raggedright\arraybackslash}p{0.31\textwidth}}
        \toprule
        \textbf{Model} & \textbf{Size} & \textbf{Open} & \textbf{Role} \\
        \midrule
        GPT-5.4 & -- & no & benchmark; test-set translation \\
        Claude Sonnet 4.6 & -- & no & benchmark \\
        Qwen3.6-35B-A3B~\citep{qwen36_35b_a3b} & 35B (3B) & yes & benchmark; teacher \\
        Qwen3.5-35B-A3B~\citep{qwen3.5} & 35B (3B) & yes & benchmark, this appendix only \\
        GPT-OSS-120B~\citep{openai2025gptoss120bgptoss20bmodel} & 117B (5.1B) & yes & benchmark; teacher; training-set translation; refinement worker and judge \\
        GPT-OSS-20B~\citep{openai2025gptoss120bgptoss20bmodel} & 21B (3.6B) & yes & benchmark \\
        MedGemma-27B~\citep{sellergren2025medgemma} & 27B & yes & benchmark \\
        Qwen3.8-27B~\citep{qwen38} & 27.8B & yes & benchmark; refinement detector, checker, and judge \\
        Gemma-4-31B-it~\citep{gemmateam2026gemma4} & 31.3B & yes & benchmark; refinement judge \\
        Qwen3-4B-Thinking-2507~\citep{qwen2025qwen3technicalreport} & 4B & yes & benchmark \\
        Qwen2.5-3B-Instruct~\citep{qwen2025qwen25technicalreport} & 3.1B & yes & benchmark \\
        LFM2.5-1.2B-Instruct~\citep{amini2025lfm2} & 1.2B & yes & student \\
        EuroLLM-1.7B-Instruct~\citep{martins2025eurollm} & 1.7B & yes & student \\
        \bottomrule
    \end{tabular}
    \caption{Models and their roles. Size: total parameters (active parameters per token for mixture-of-experts models). Benchmark: evaluated in Part~1.}
    \label{tab:models}
\end{table}

\paragraph{Benchmark models.} Table~\ref{tab:models} lists all models and their roles. The open-weight models decode greedily (temperature 0), answer in the JSON schema \{explanation, answer\_index\}, and generate up to 30,720 tokens. Reasoning is turned off with each model's documented switch. With reasoning on, Qwen3.8-27B thinks at its default effort, and Gemma-4-31B always reasons before it answers. Qwen2.5-3B and MedGemma-27B have one mode. With reasoning on, Qwen3.8-27B sometimes ends its output inside the reasoning, without an answer; this happens on 207 Swedish core items, which count as wrong. GPT-5.4 and Claude Sonnet 4.6 ran through their batch interfaces with medium reasoning effort, JSON-schema output, and at most 4,096 output tokens, reasoning included. Table~\ref{tab:api_prices} gives their prices and costs, and Appendix~\ref{sec:appendix-repro} gives their API versions.

\paragraph{Students.} The students and their base models decode greedily into the same JSON schema. The context is 12,288 tokens; LFM2.5-1.2B may generate up to 12,288 tokens (10,240 for the SV-rewritten model) and the trained EuroLLM-1.7B students up to 10,240; the base models may generate up to 12,288. The base students use their default chat behavior, and the trained students have two decoding modes. In \emph{reasoning} decoding, the student writes its reasoning and then the JSON answer. In \emph{direct} decoding, it answers at once, and the answer still contains a short explanation.

\paragraph{Scoring.} An output that is unfinished at the length limit, does not match the schema, or names an option that does not exist counts as wrong; we report this \emph{invalid} share separately. For the open models and the students, we compute 95\% bootstrap intervals (10,000 resamples of items) for each accuracy and for the paired English--Swedish gap, and an exact McNemar test on the paired items.

\paragraph{Averages.} The main text weights each dataset by its number of test items. Table~\ref{tab:benchmark_full} also gives the macro average, in which every dataset counts equally, as in the original reports of several of these datasets; both averages give the same picture.

\begin{table}[htb]
    \centering
    \small
\begin{tabular}{lrrrrrr}
\toprule
& \multicolumn{3}{c}{\textbf{Macro average}} & \multicolumn{3}{c}{\textbf{Item-weighted}} \\
\cmidrule(lr){2-4}\cmidrule(lr){5-7}
\textbf{Model} & \textbf{EN} & \textbf{SV} & $\boldsymbol{\Delta}$ & \textbf{EN} & \textbf{SV} & $\boldsymbol{\Delta}$ \\
\midrule
\multicolumn{7}{l}{\textit{Proprietary (medium reasoning)}} \\
GPT-5.4 & 80.82 & 80.51 & \pos{0.31} & 81.30 & 80.98 & \pos{0.32} \\
Claude Sonnet 4.6 & 77.84 & 76.19 & \pos{1.65} & 77.70 & 76.27 & \pos{1.43} \\
\midrule
\multicolumn{7}{l}{\textit{Open-weight, reasoning on or single mode}} \\
Gemma-4-31B & 78.53 & 78.21 & \pos{0.33} & 78.74 & 77.97 & \pos{0.77} \\
Qwen3.6-35B-A3B & 75.48 & 74.10 & \pos{1.38} & 74.68 & 73.34 & \pos{1.35} \\
Qwen3.8-27B & 75.26 & 72.66 & \pos{2.60} & 74.62 & 71.97 & \pos{2.65} \\
GPT-OSS-120B & 72.84 & 71.34 & \pos{1.49} & 71.82 & 70.42 & \pos{1.40} \\
GPT-OSS-20B & 67.07 & 65.81 & \pos{1.26} & 65.76 & 63.67 & \pos{2.09} \\
MedGemma-27B$^\dagger$ & 65.50 & 61.87 & \pos{3.63} & 64.50 & 60.55 & \pos{3.96} \\
Qwen3-4B-Thinking & 60.12 & 59.05 & \pos{1.07} & 59.28 & 57.43 & \pos{1.85} \\
Qwen2.5-3B$^\dagger$ & 45.35 & 34.04 & \pos{11.32} & 44.88 & 32.48 & \pos{12.40} \\
\midrule
\multicolumn{7}{l}{\textit{Open-weight, reasoning off}} \\
Gemma-4-31B & 77.32 & 74.57 & \pos{2.76} & 77.07 & 74.03 & \pos{3.04} \\
Qwen3.6-35B-A3B & 72.87 & 69.66 & \pos{3.21} & 72.16 & 68.48 & \pos{3.68} \\
Qwen3.8-27B & 71.37 & 67.73 & \pos{3.64} & 70.61 & 66.46 & \pos{4.14} \\
GPT-OSS-120B & 65.95 & 58.71 & \pos{7.24} & 64.29 & 56.91 & \pos{7.38} \\
GPT-OSS-20B & 60.63 & 52.39 & \pos{8.24} & 59.01 & 50.64 & \pos{8.36} \\
Qwen3-4B-Thinking & 54.73 & 44.89 & \pos{9.83} & 53.87 & 43.35 & \pos{10.52} \\
\midrule
Mean of open-weight configurations & 67.36 & 63.22 & \pos{4.14} & 66.52 & 61.98 & \pos{4.54} \\
\bottomrule
\end{tabular}

    \caption{Benchmark accuracy (\%) with two averages. $\Delta=\mathrm{EN}-\mathrm{SV}$. Macro: mean over the ten datasets. Item-weighted: mean weighted by the number of core items. $^\dagger$Models without a reasoning mode.}
    \label{tab:benchmark_full}
\end{table}

\paragraph{Qwen3.5-35B-A3B.} We also evaluated Qwen3.5-35B-A3B, the previous version of Qwen3.6-35B-A3B, with the same size and settings, but in Swedish only on the unrefined draft translation. Table~\ref{tab:benchmark_appendix_only} shows its results next to those of Qwen3.6. Because its Swedish text differs from the benchmark's, we leave it out of all other tables, figures, and averages.

\begin{table}[htb]
    \centering
    \footnotesize
    \setlength{\tabcolsep}{4pt}
\begin{tabular}{lrrrrrr}
\toprule
& \multicolumn{3}{c}{\textbf{Reasoning on}} & \multicolumn{3}{c}{\textbf{Reasoning off}} \\
\cmidrule(lr){2-4}\cmidrule(lr){5-7}
\textbf{Model} & \textbf{EN} & \textbf{SV} & $\boldsymbol{\Delta}$ & \textbf{EN} & \textbf{SV} & $\boldsymbol{\Delta}$ \\
\midrule
Qwen3.6-35B-A3B & 74.7 & 73.3 & \pos{1.3} & 72.2 & 68.5 & \pos{3.7} \\
Qwen3.5-35B-A3B & 73.3 & 70.7 & \pos{2.6} & 70.6 & 67.4 & \pos{3.3} \\
\bottomrule
\end{tabular}

    \caption{Qwen3.5-35B-A3B next to Qwen3.6-35B-A3B: accuracy (\%, weighted by items) on the core test items; the Swedish items of Qwen3.5 use the unrefined draft translation. $\Delta=\mathrm{EN}-\mathrm{SV}$.}
    \label{tab:benchmark_appendix_only}
\end{table}

\begin{figure}[htb]
    \centering
    \includegraphics[width=\textwidth]{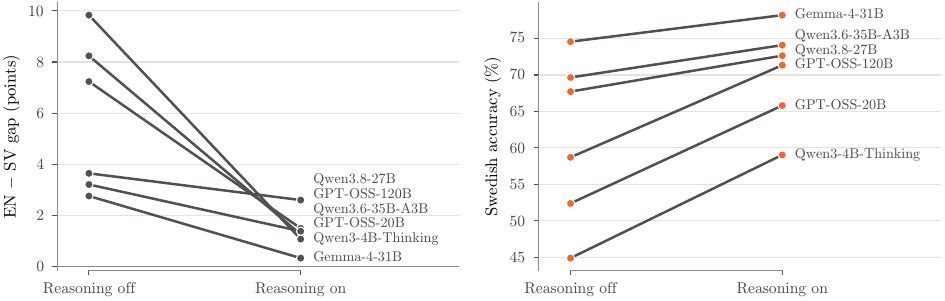}
    \caption{Effect of reasoning for the 6 open models that support both modes. Left: English--Swedish gap. Right: Swedish accuracy. Macro averages over the ten datasets.}
    \label{fig:reasoning_gap}
\end{figure}

\begin{figure}[htb]
    \centering
    \includegraphics[width=\textwidth]{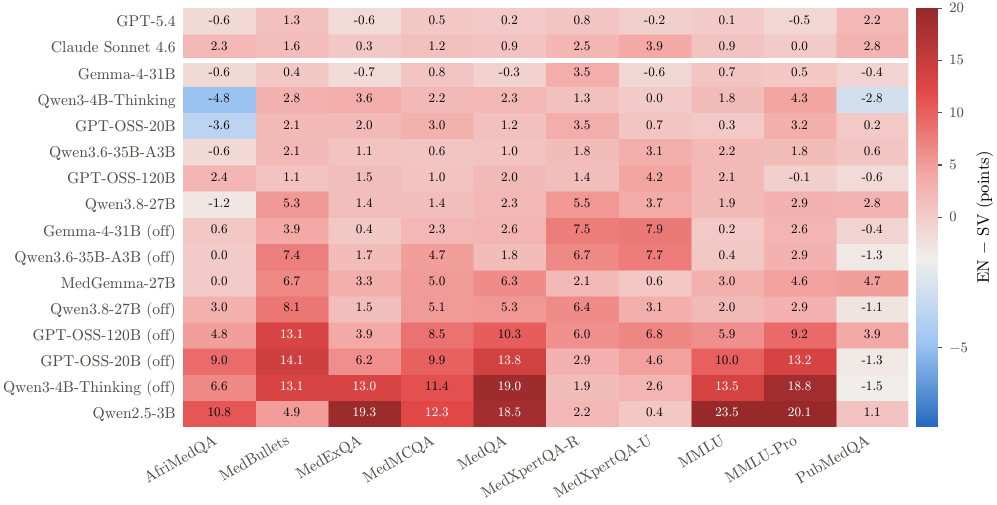}
    \caption{English--Swedish gap (points) per model configuration and dataset. Red: Swedish lower; blue: Swedish higher. Proprietary models are above the white line; ``(off)'': reasoning off.}
    \label{fig:benchmark_heatmap}
\end{figure}

\paragraph{Effect of reasoning.} Figure~\ref{fig:reasoning_gap} compares the 6 open models that can answer with and without reasoning: with reasoning, the gap shrinks for every model, and Swedish accuracy rises more than English accuracy. For Qwen3-4B-Thinking, Swedish accuracy rises by 14.1 points and English by 5.4. Table~\ref{tab:reasoning_did} tests this on paired items: for each of the six models, the 95\% interval of the drop in the gap, which equals the Swedish gain minus the English gain, excludes zero, from 1.5 points for Qwen3.8-27B to 8.7 points for Qwen3-4B-Thinking. Figure~\ref{fig:benchmark_heatmap} shows the gap per configuration and dataset.

\begin{table}[htb]
    \centering
    \small
\begin{tabular}{lrrrrr}
\toprule
& \multicolumn{2}{c}{\textbf{EN $-$ SV gap}} & \multicolumn{2}{c}{\textbf{Gain from reasoning, 95\% CI}} & \textbf{Gap off $-$ gap on} \\
\cmidrule(lr){2-3}\cmidrule(lr){4-5}
\textbf{Model} & \textbf{Off} & \textbf{On} & \textbf{SV} & \textbf{EN} & \textbf{95\% CI} \\
\midrule
Gemma-4-31B & 3.0 & 0.8 & 3.9 [3.3, 4.6] & 1.7 [1.1, 2.3] & 2.3 [1.5, 3.1] \\
Qwen3.6-35B-A3B & 3.7 & 1.3 & 4.9 [4.2, 5.6] & 2.5 [1.9, 3.1] & 2.3 [1.5, 3.2] \\
Qwen3.8-27B & 4.1 & 2.7 & 5.5 [4.7, 6.3] & 4.0 [3.3, 4.7] & 1.5 [0.5, 2.4] \\
GPT-OSS-120B & 7.4 & 1.4 & 13.5 [12.6, 14.4] & 7.5 [6.7, 8.3] & 6.0 [5.0, 7.0] \\
GPT-OSS-20B & 8.4 & 2.1 & 13.0 [12.1, 14.0] & 6.8 [5.9, 7.6] & 6.3 [5.2, 7.3] \\
Qwen3-4B-Thinking & 10.5 & 1.8 & 14.1 [13.1, 15.0] & 5.4 [4.6, 6.2] & 8.7 [7.6, 9.8] \\
\midrule
Mean (6 models) & 6.2 & 1.7 & 9.2 [8.8, 9.6] & 4.7 [4.3, 5.0] & 4.5 [4.1, 4.9] \\
\bottomrule
\end{tabular}

    \caption{Effect of reasoning for the six open models with both modes, on the core items (item-weighted). Gain: accuracy with reasoning minus without, per language. Gap off $-$ gap on: the drop of the English--Swedish gap. Intervals: paired bootstrap over items, 10,000 resamples.}
    \label{tab:reasoning_did}
\end{table}

\paragraph{Why reasoning narrows the gap.}\label{sec:analysis-reasoning} We test whether reasoning helps on Swedish questions because the models restate the question in English. A rule classifies the first 150 words of each Swedish reasoning trace by language and checks whether the trace renders the Swedish question in English. We tuned the rule on 102 traces and checked it on 48 more, all labeled by Claude Opus 5.5 with \textit{xhigh} reasoning effort from the trace and the English question. The rule agrees with these labels on 126 of the 150 traces, on 70 of 72 random ones, and on all 39 it labels Swedish; on the 48 check traces it agrees on 32, mostly because it calls some restatements English reasoning without a restatement, a class that is rare (1 of 72 random traces). Table~\ref{tab:reasoning_translation} gives the full results, including Swedish accuracy and the gap by the language of the opening. Qwen3.8-27B, which reasons in Swedish in 95.8\% of its traces, has the largest gap with reasoning, 2.7 points, but outputs that end without an answer add to it; on the items that each model answers validly in both languages, its gap is 1.6 points, within the 1.2--1.8 points of the other five models. Restating the question in English is thus common, as Section~\ref{sec:analysis-language} shows, but not necessary; this comparison is observational, not a controlled test. The boxes below show the three kinds of openings.

\begin{table}[htb]
    \centering
    \small
    \resizebox{\linewidth}{!}{%
\begin{tabular}{lrrrrrrrrr}
\toprule
 & \multicolumn{3}{c}{\textbf{Opening (\%)}} & \multicolumn{2}{c}{\textbf{Gap EN$-$SV}} & \multicolumn{2}{c}{\textbf{SV acc.\ by opening}} & \multicolumn{2}{c}{\textbf{Gap by opening}} \\
\cmidrule(lr){2-4} \cmidrule(lr){5-6} \cmidrule(lr){7-8} \cmidrule(lr){9-10}
\textbf{Model} & Tr. & EN & SV & Direct & Reas. & EN & SV & EN & SV \\
\midrule
GPT-OSS-120B & 99.4 & 0.6 & 0.0 & 7.4 & 1.4 & 70.4 & -- & 1.4 & -- \\
GPT-OSS-20B & 98.8 & 1.1 & 0.0 & 8.4 & 2.1 & 63.7 & -- & 2.1 & -- \\
Gemma-4-31B & 99.6 & 0.3 & 0.0 & 3.0 & 0.8 & 78.0 & -- & 0.8 & -- \\
Qwen3.6-35B-A3B & 99.0 & 1.0 & 0.0 & 3.7 & 1.3 & 73.4 & -- & 1.3 & -- \\
Qwen3.8-27B & 2.4 & 0.2 & 95.8 & 4.1 & 2.7 & 66.7 & 71.7 & 0.0 & 2.7 \\
Qwen3-4B-Thinking & 99.3 & 0.7 & 0.0 & 10.5 & 1.8 & 57.4 & -- & 1.8 & -- \\
\midrule
LFM2.5-1.2B SV & 96.9 & 0.9 & 0.0 & 0.8 & 0.3 & 50.9 & -- & -0.2 & -- \\
LFM2.5-1.2B EN+SV & 97.3 & 0.6 & 0.0 & 3.1 & -1.5 & 53.0 & -- & -1.8 & -- \\
LFM2.5-1.2B SV (rewritten) & 0.0 & 0.0 & 99.6 & -0.1 & -0.3 & -- & 44.5 & -- & -0.4 \\
EuroLLM-1.7B SV & 96.7 & 0.7 & 0.0 & 0.6 & -0.2 & 45.7 & -- & -0.8 & -- \\
EuroLLM-1.7B SV (rewritten) & 0.0 & 0.0 & 99.1 & 0.6 & 0.5 & -- & 39.3 & -- & 0.3 \\
\bottomrule
\end{tabular}
}
    \caption{Opening of the Swedish reasoning traces and the English--Swedish gap. Opening: share of traces that restate the question in English (Tr.), reason in English over the quoted Swedish text without a restatement (EN), or reason in Swedish (SV); the EN class is rare and not reliably detected. Gap: EN$-$SV accuracy in points on the 9,733 core items, with direct answers and with reasoning. By opening: Swedish accuracy and gap on the items with an English (Tr.\ or EN) or a Swedish opening; cells with fewer than 30 items show --. Students use reasoning decoding; direct is their direct decoding.}
    \label{tab:reasoning_translation}
\end{table}

\begin{tcolorbox}[colback=white, colframe=black!40, boxrule=0.4pt, left=3pt, right=3pt, top=2pt, bottom=2pt, fontupper=\small\raggedright]
\textbf{GPT-OSS-20B, restates in English.} \textit{We need to answer: ``Vilken är glykolysens slutliga metabola produkt?'' Swedish: ``What is the final metabolic product of glycolysis?'' Options: A) amino acid, B) neutral fat, C) glycerol, D) pyruvate.}

\smallskip
\textbf{Gemma-4-31B, English gloss.} \textit{Question: ``Hur snabbt börjar proparakain verka?'' (How quickly does proparacaine start working?) Options: A) 2 minutes, B) 5 minutes, C) 15 minutes, D) 20 minutes.}

\smallskip
\textbf{Qwen3.8-27B, reasons in Swedish.} \textit{Vi behöver svara på svenska i JSON. Frågan: Vilken artär palperas för att mäta pulsen på handledens framsida? Alternativ: ulnaris, radialis, brachialis, axillaris. Korrekt är radialisartären \ldots}
\end{tcolorbox}

\paragraph{Uncertainty and item-level agreement.} For the open models we have the output for every item, and Table~\ref{tab:benchmark_paired} gives their paired statistics. The gap is significant for all 14 open configurations under the exact McNemar test ($p<0.05$), also after the Holm correction over the 14 tests, even for the best models with reasoning. Still, many items are solved in only one of the two languages, in both directions.

\begin{table}[htb]
    \centering
    \small
    \resizebox{\textwidth}{!}{%
\begin{tabular}{llrrrrrrr}
\toprule
\textbf{Model} & \textbf{Reas.} & \textbf{EN} & \textbf{SV} & $\boldsymbol{\Delta}$ & \textbf{EN only} & \textbf{SV only} & \textbf{McNemar $p$} & \textbf{Holm $p$} \\
\midrule
Gemma-4-31B & on & 78.7 [77.9, 79.6] & 78.0 [77.1, 78.8] & 0.8 [0.2, 1.4] & 434 & 359 & 0.009 & 0.009 \\
Gemma-4-31B & off & 77.1 [76.2, 77.9] & 74.0 [73.1, 74.9] & 3.0 [2.4, 3.7] & 639 & 343 & $<$0.001 & $<$0.001 \\
Qwen3.6-35B-A3B & on & 74.7 [73.8, 75.6] & 73.3 [72.5, 74.2] & 1.3 [0.7, 1.9] & 537 & 406 & $<$0.001 & $<$0.001 \\
Qwen3.6-35B-A3B & off & 72.2 [71.3, 73.1] & 68.5 [67.6, 69.4] & 3.7 [3.0, 4.4] & 807 & 449 & $<$0.001 & $<$0.001 \\
Qwen3.8-27B & on & 74.6 [73.7, 75.5] & 72.0 [71.1, 72.9] & 2.7 [2.0, 3.3] & 704 & 446 & $<$0.001 & $<$0.001 \\
Qwen3.8-27B & off & 70.6 [69.7, 71.5] & 66.5 [65.5, 67.4] & 4.1 [3.4, 4.9] & 930 & 527 & $<$0.001 & $<$0.001 \\
GPT-OSS-120B & on & 71.8 [70.9, 72.7] & 70.4 [69.5, 71.3] & 1.4 [0.8, 2.0] & 554 & 418 & $<$0.001 & $<$0.001 \\
GPT-OSS-120B & off & 64.3 [63.3, 65.3] & 56.9 [55.9, 57.9] & 7.4 [6.5, 8.2] & 1,266 & 548 & $<$0.001 & $<$0.001 \\
GPT-OSS-20B & on & 65.8 [64.8, 66.7] & 63.7 [62.7, 64.6] & 2.1 [1.4, 2.8] & 756 & 553 & $<$0.001 & $<$0.001 \\
GPT-OSS-20B & off & 59.0 [58.0, 60.0] & 50.6 [49.7, 51.6] & 8.4 [7.4, 9.3] & 1,457 & 643 & $<$0.001 & $<$0.001 \\
MedGemma-27B & -- & 64.5 [63.6, 65.5] & 60.5 [59.6, 61.5] & 4.0 [3.2, 4.7] & 925 & 540 & $<$0.001 & $<$0.001 \\
Qwen3-4B-Thinking & on & 59.3 [58.3, 60.3] & 57.4 [56.4, 58.4] & 1.8 [1.1, 2.6] & 832 & 652 & $<$0.001 & $<$0.001 \\
Qwen3-4B-Thinking & off & 53.9 [52.9, 54.8] & 43.3 [42.4, 44.3] & 10.5 [9.6, 11.5] & 1,697 & 673 & $<$0.001 & $<$0.001 \\
Qwen2.5-3B & -- & 44.9 [43.9, 45.9] & 32.5 [31.6, 33.4] & 12.4 [11.3, 13.5] & 2,104 & 897 & $<$0.001 & $<$0.001 \\
\bottomrule
\end{tabular}
}
    \caption{Open models: accuracy (\%) with 95\% bootstrap intervals, the paired gap $\Delta=\mathrm{EN}-\mathrm{SV}$ with its interval, the number of items solved only in English or only in Swedish, and the exact McNemar test. Reas.: reasoning mode ($--$: single-mode model).}
    \label{tab:benchmark_paired}
\end{table}

\paragraph{Hard subset.} Table~\ref{tab:benchmark_hard} splits accuracy into the 840 hard items and the remainder. Averaged over all 14 open configurations, with and without reasoning, accuracy on hard items is 28.1\% in English and 26.8\% in Swedish, against 70.2\% and 65.3\% on the remainder, so the gap is 1.3 points on hard items and 4.8 points on the remainder. The two best students, which are not in these averages, answer 18.1 to 21.9\% of the hard items correctly in either language. The hard items are near chance, which is 20.4\% on them and 20.7\% on the core, and the collection's authors selected them because models fail on them, which makes them hard by construction; this selection also compresses their accuracies and gaps. Table~\ref{tab:benchmark_hard_chance} therefore rescales accuracy as $(a-c)/(1-c)$, with chance $c$; on this scale the two students score between $-2.8$ and 1.9 on the hard items, that is, at chance. Appendix~\ref{sec:appendix-hard} looks at what makes these items hard.

\begin{table}[htb]
    \centering
    \small
\begin{tabular}{llrrrrrr}
\toprule
& & \multicolumn{2}{c}{\textbf{Hard (840)}} & \multicolumn{2}{c}{\textbf{Remainder (8,893)}} & \multicolumn{2}{c}{$\boldsymbol{\Delta}$} \\
\cmidrule(lr){3-4}\cmidrule(lr){5-6}\cmidrule(lr){7-8}
\textbf{Model} & \textbf{Reas.} & \textbf{EN} & \textbf{SV} & \textbf{EN} & \textbf{SV} & \textbf{Hard} & \textbf{Rem.} \\
\midrule
Gemma-4-31B & on & 44.0 & 44.0 & 82.0 & 81.2 & 0.0 & 0.8 \\
Gemma-4-31B & off & 40.0 & 35.5 & 80.6 & 77.7 & 4.5 & 2.9 \\
Qwen3.6-35B-A3B & on & 37.1 & 35.2 & 78.2 & 76.9 & 1.9 & 1.3 \\
Qwen3.6-35B-A3B & off & 29.8 & 27.5 & 76.2 & 72.3 & 2.3 & 3.8 \\
Qwen3.8-27B & on & 42.3 & 33.9 & 77.7 & 75.6 & 8.3 & 2.1 \\
Qwen3.8-27B & off & 29.0 & 27.4 & 74.5 & 70.2 & 1.7 & 4.4 \\
GPT-OSS-120B & on & 33.1 & 33.0 & 75.5 & 74.0 & 0.1 & 1.5 \\
GPT-OSS-120B & off & 21.3 & 20.7 & 68.3 & 60.3 & 0.6 & 8.0 \\
GPT-OSS-20B & on & 25.1 & 26.5 & 69.6 & 67.2 & -1.4 & 2.4 \\
GPT-OSS-20B & off & 17.3 & 18.0 & 62.9 & 53.7 & -0.7 & 9.2 \\
MedGemma-27B & -- & 21.3 & 19.0 & 68.6 & 64.5 & 2.3 & 4.1 \\
Qwen3-4B-Thinking & on & 20.1 & 21.1 & 63.0 & 60.9 & -1.0 & 2.1 \\
Qwen3-4B-Thinking & off & 15.7 & 15.7 & 57.5 & 46.0 & 0.0 & 11.5 \\
Qwen2.5-3B & -- & 16.8 & 17.1 & 47.5 & 33.9 & -0.4 & 13.6 \\
\midrule
\multicolumn{8}{l}{\textit{Students (best Swedish model)}} \\
LFM2.5-1.2B, EN+SV traces & on & 18.7 & 21.9 & 53.5 & 54.8 & -3.2 & -1.3 \\
EuroLLM-1.7B, SV traces & on & 18.9 & 18.1 & 46.7 & 47.0 & 0.8 & -0.3 \\
\bottomrule
\end{tabular}

    \caption{Open models and the two best students: accuracy (\%) on the 840 hard items and on the 8,893 remaining core items, and the English--Swedish gap on each. Students: the best Swedish model of each base model, LFM2.5-1.2B trained on English and Swedish traces and EuroLLM-1.7B trained on Swedish traces, both with reasoning decoding; they do not enter the averages in the text.}
    \label{tab:benchmark_hard}
\end{table}

\begin{table}[htb]
    \centering
    \small
\begin{tabular}{llrrrrrr}
\toprule
& & \multicolumn{2}{c}{\textbf{Hard (840)}} & \multicolumn{2}{c}{\textbf{Remainder (8,893)}} & \multicolumn{2}{c}{$\boldsymbol{\Delta}$} \\
\cmidrule(lr){3-4}\cmidrule(lr){5-6}\cmidrule(lr){7-8}
\textbf{Model} & \textbf{Reas.} & \textbf{EN} & \textbf{SV} & \textbf{EN} & \textbf{SV} & \textbf{Hard} & \textbf{Rem.} \\
\midrule
Gemma-4-31B & on & 29.7 & 29.7 & 77.3 & 76.3 & 0.0 & 1.1 \\
Gemma-4-31B & off & 24.7 & 19.0 & 75.5 & 71.8 & 5.7 & 3.7 \\
Qwen3.6-35B-A3B & on & 21.1 & 18.7 & 72.5 & 70.9 & 2.4 & 1.6 \\
Qwen3.6-35B-A3B & off & 11.8 & 9.0 & 69.9 & 65.1 & 2.8 & 4.8 \\
Qwen3.8-27B & on & 27.5 & 17.0 & 71.8 & 69.2 & 10.5 & 2.7 \\
Qwen3.8-27B & off & 10.9 & 8.8 & 67.9 & 62.3 & 2.1 & 5.5 \\
GPT-OSS-120B & on & 16.0 & 15.8 & 69.1 & 67.1 & 0.1 & 1.9 \\
GPT-OSS-120B & off & 1.2 & 0.4 & 60.1 & 49.9 & 0.7 & 10.1 \\
GPT-OSS-20B & on & 6.0 & 7.8 & 61.6 & 58.6 & -1.8 & 3.1 \\
GPT-OSS-20B & off & -3.9 & -3.0 & 53.3 & 41.6 & -0.9 & 11.6 \\
MedGemma-27B & -- & 1.2 & -1.6 & 60.4 & 55.2 & 2.8 & 5.2 \\
Qwen3-4B-Thinking & on & -0.3 & 0.9 & 53.3 & 50.6 & -1.2 & 2.7 \\
Qwen3-4B-Thinking & off & -5.8 & -5.8 & 46.3 & 31.8 & 0.0 & 14.5 \\
Qwen2.5-3B & -- & -4.5 & -4.0 & 33.8 & 16.6 & -0.4 & 17.2 \\
\midrule
\multicolumn{8}{l}{\textit{Students (best Swedish model)}} \\
LFM2.5-1.2B, EN+SV traces & on & -2.1 & 1.9 & 41.3 & 43.0 & -4.0 & -1.6 \\
EuroLLM-1.7B, SV traces & on & -1.8 & -2.8 & 32.8 & 33.2 & 1.0 & -0.4 \\
\bottomrule
\end{tabular}

    \caption{Accuracy above chance, $(a-c)/(1-c)$ in \%, on the 840 hard items and the 8,893 remaining core items, where $c$ is the mean chance level $1/k$ of the items, with $k$ options. Negative values are below chance.}
    \label{tab:benchmark_hard_chance}
\end{table}

\paragraph{Overlap between training and test questions.} The splits of the collection are disjoint by item, but the same question can appear in another dataset's training split; the MedBullets training file, for example, is an exact copy of the MedQA file, which we keep as distributed. We compared every core test question with the 141,735 training questions, after lowercasing and removing punctuation, both exactly and by the Jaccard similarity of their sets of word 8-grams, with a threshold of 0.8 and an exact index, so that no pair above it is missed. Table~\ref{tab:overlap} shows that 11 core items have a near duplicate in training, 6 of them an exact one. Removing them changes the Swedish accuracy of the best LFM2.5-1.2B student by at most 0.23 points on any dataset and by less than 0.01 points overall. Within the test set, 536 pairs of MMLU and MMLU-Pro items are near duplicates, because MMLU-Pro reuses MMLU questions.

\begin{table}[htb]
    \centering
    \small
\begin{tabular}{lrrrrr}
\toprule
& & \multicolumn{3}{c}{\textbf{Test items with a training duplicate}} & \\
\cmidrule(lr){3-5}
\textbf{Dataset} & \textbf{Core} & \textbf{Exact Q} & \textbf{Exact Q+O} & \textbf{Near} & $\boldsymbol{\Delta}$\textbf{SV} \\
\midrule
AfriMedQA & 166 & 1 & 0 & 1 & $-$0.23 \\
MedBullets & 283 & 0 & 0 & 4 & +0.15 \\
MedExQA & 921 & 0 & 0 & 0 & 0.00 \\
MedMCQA & 2,666 & 2 & 1 & 3 & $-$0.01 \\
MedQA & 1,179 & 0 & 0 & 0 & 0.00 \\
MedXpertQA-R & 1,665 & 0 & 0 & 0 & 0.00 \\
MedXpertQA-U & 544 & 0 & 0 & 0 & 0.00 \\
MMLU & 1,060 & 1 & 1 & 1 & +0.06 \\
MMLU-Pro & 783 & 2 & 0 & 2 & $-$0.04 \\
PubMedQA & 466 & 0 & 0 & 0 & 0.00 \\
\midrule
Total & 9,733 & 6 & 2 & 11 & 0.00 \\
\bottomrule
\end{tabular}

    \caption{Core test items with a duplicate among the training questions of any dataset: exact question, exact question and options, and near duplicate (Jaccard similarity of word 8-grams $\geq 0.8$, including exact ones). $\Delta$SV: change in the Swedish accuracy (points) of LFM2.5-1.2B trained on English and Swedish traces when these items are removed.}
    \label{tab:overlap}
\end{table}

\begin{figure}[htb]
    \centering
    \includegraphics[width=\textwidth]{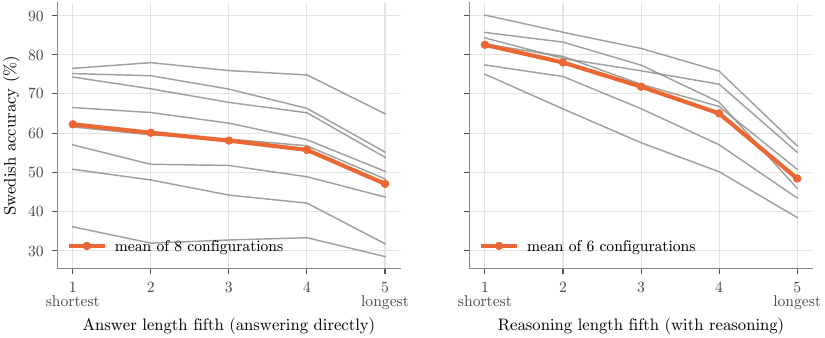}
    \caption{Swedish accuracy by output length, within each open configuration (gray lines) and on average (orange). Left: configurations that answer directly, by the length of the explanation. Right: configurations with reasoning, by the length of the reasoning. Fifths are formed within each dataset, so every fifth has the same mix of datasets.}
    \label{fig:benchmark_tokens}
\end{figure}

\paragraph{Generated tokens.} Table~\ref{tab:benchmark_tokens} splits the generated tokens of each open configuration into reasoning and answer; Swedish outputs are longer than English ones in 12 of the 14 configurations. Across configurations, accuracy does not follow length: the rank correlation between Swedish accuracy and the median number of generated tokens is 0.03 with reasoning and 0.19 without, and neither is significant. Within a configuration, however, long outputs go with wrong answers. Figure~\ref{fig:benchmark_tokens} shows that Swedish accuracy falls from 82.5\% in the shortest fifth of reasoning lengths to 48.4\% in the longest, and from 62.2\% to 47.0\% by explanation length, and wrong answers are longer than correct ones in every configuration: by explanation length in the 8 that answer directly and by reasoning length in the 6 that reason. Output length thus mostly reflects how hard an item is for the model, not how carefully it was answered.

\begin{table}[htb]
    \centering
    \small
    \resizebox{\textwidth}{!}{%
\begin{tabular}{llrrrrrrrr}
\toprule
 & & \multicolumn{2}{c}{\textbf{Reasoning}} & \multicolumn{2}{c}{\textbf{Answer}} & \multicolumn{2}{c}{\textbf{SV answer}} & \multicolumn{2}{c}{\textbf{SV acc.\ by length}} \\
\cmidrule(lr){3-4} \cmidrule(lr){5-6} \cmidrule(lr){7-8} \cmidrule(lr){9-10}
\textbf{Model} & \textbf{Reas.} & EN & SV & EN & SV & correct & wrong & shortest & longest \\
\midrule
Gemma-4-31B & on & 819 & 688 & 162 & 189 & 180 & 229 & 90.1 & 56.7 \\
Gemma-4-31B & off & -- & -- & 148 & 174 & 168 & 196 & 76.5 & 64.9 \\
Qwen3.6-35B-A3B & on & 1,602 & 1,936 & 155 & 186 & 181 & 201 & 84.3 & 55.1 \\
Qwen3.6-35B-A3B & off & -- & -- & 209 & 219 & 203 & 262 & 75.2 & 55.1 \\
Qwen3.8-27B & on & 505 & 529 & 117 & 143 & 141 & 152 & 85.7 & 45.9 \\
Qwen3.8-27B & off & -- & -- & 200 & 236 & 217 & 284 & 74.3 & 53.8 \\
GPT-OSS-120B & on & 272 & 298 & 101 & 135 & 132 & 143 & 82.7 & 50.7 \\
GPT-OSS-120B & off & -- & -- & 102 & 138 & 128 & 153 & 61.6 & 48.3 \\
GPT-OSS-20B & on & 337 & 380 & 115 & 129 & 125 & 137 & 77.4 & 43.4 \\
GPT-OSS-20B & off & -- & -- & 120 & 140 & 130 & 153 & 57.0 & 43.7 \\
MedGemma-27B & -- & -- & -- & 223 & 271 & 239 & 361 & 66.5 & 50.2 \\
Qwen3-4B-Thinking & on & 2,724 & 2,076 & 92 & 170 & 163 & 179 & 75.0 & 38.5 \\
Qwen3-4B-Thinking & off & -- & -- & 180 & 238 & 211 & 262 & 50.7 & 31.7 \\
Qwen2.5-3B & -- & -- & -- & 127 & 169 & 162 & 172 & 36.1 & 28.5 \\
\bottomrule
\end{tabular}
}
    \caption{Generated tokens of the open models (median per core item). Reasoning: tokens before the answer; answer: the JSON with the explanation and the answer index. SV answer: median answer length of correct and wrong Swedish answers. SV acc.\ by length: Swedish accuracy (\%) in the shortest and longest fifth of the outputs, by answer length when answering directly and by reasoning length otherwise; fifths are formed within each dataset.}
    \label{tab:benchmark_tokens}
\end{table}

\paragraph{Per-dataset results.} Tables~\ref{tab:benchmark_per_dataset_a} and~\ref{tab:benchmark_per_dataset_b} give the accuracy of every configuration per dataset.

\begin{table}[htb]
    \centering
    \resizebox{\textwidth}{!}{%
\begin{tabular}{lrrrrrrrrrrrrrrr}
\toprule
& \multicolumn{3}{c}{\textbf{AfriMedQA}} & \multicolumn{3}{c}{\textbf{MedBullets}} & \multicolumn{3}{c}{\textbf{MedExQA}} & \multicolumn{3}{c}{\textbf{MedMCQA}} & \multicolumn{3}{c}{\textbf{MedQA}} \\
\cmidrule(lr){2-4}\cmidrule(lr){5-7}\cmidrule(lr){8-10}\cmidrule(lr){11-13}\cmidrule(lr){14-16}
\textbf{Model} & EN & SV & $\Delta$ & EN & SV & $\Delta$ & EN & SV & $\Delta$ & EN & SV & $\Delta$ & EN & SV & $\Delta$ \\
\midrule
GPT-5.4 & 81.03 & 81.61 & \negative{-0.58} & 88.64 & 87.34 & \pos{1.30} & 87.49 & 88.13 & \negative{-0.64} & 86.58 & 86.08 & \pos{0.50} & 96.39 & 96.23 & \pos{0.16} \\
Claude Sonnet 4.6 & 84.48 & 82.18 & \pos{2.30} & 83.77 & 82.14 & \pos{1.63} & 87.17 & 86.84 & \pos{0.33} & 83.66 & 82.49 & \pos{1.17} & 94.74 & 93.79 & \pos{0.95} \\
\midrule
Gemma-4-31B & 78.31 & 78.92 & \negative{-0.60} & 83.39 & 83.04 & \pos{0.35} & 85.56 & 86.21 & \negative{-0.65} & 80.80 & 80.05 & \pos{0.75} & 92.79 & 93.04 & \negative{-0.25} \\
Gemma-4-31B$^*$ & 78.31 & 77.71 & \pos{0.60} & 80.57 & 76.68 & \pos{3.89} & 84.80 & 84.36 & \pos{0.43} & 78.96 & 76.67 & \pos{2.29} & 91.26 & 88.63 & \pos{2.63} \\
Qwen3.6-35B-A3B & 80.12 & 80.72 & \negative{-0.60} & 83.39 & 81.27 & \pos{2.12} & 86.21 & 85.12 & \pos{1.09} & 79.29 & 78.66 & \pos{0.64} & 92.96 & 91.94 & \pos{1.02} \\
Qwen3.6-35B-A3B$^*$ & 77.11 & 77.11 & 0.00 & 78.45 & 71.02 & \pos{7.42} & 84.15 & 82.41 & \pos{1.74} & 77.76 & 73.03 & \pos{4.73} & 90.67 & 88.89 & \pos{1.78} \\
Qwen3.8-27B & 76.51 & 77.71 & \negative{-1.20} & 87.63 & 82.33 & \pos{5.30} & 85.45 & 84.04 & \pos{1.41} & 77.38 & 75.96 & \pos{1.43} & 94.74 & 92.45 & \pos{2.29} \\
Qwen3.8-27B$^*$ & 75.90 & 72.89 & \pos{3.01} & 77.74 & 69.61 & \pos{8.13} & 83.17 & 81.65 & \pos{1.52} & 75.43 & 70.33 & \pos{5.10} & 90.08 & 84.73 & \pos{5.34} \\
GPT-OSS-120B & 79.52 & 77.11 & \pos{2.41} & 81.27 & 80.21 & \pos{1.06} & 85.12 & 83.60 & \pos{1.52} & 76.37 & 75.36 & \pos{1.01} & 92.62 & 90.67 & \pos{1.95} \\
GPT-OSS-120B$^*$ & 72.89 & 68.07 & \pos{4.82} & 67.49 & 54.42 & \pos{13.07} & 80.89 & 76.98 & \pos{3.91} & 69.35 & 60.88 & \pos{8.48} & 81.17 & 70.91 & \pos{10.26} \\
GPT-OSS-20B & 72.89 & 76.51 & \negative{-3.61} & 71.38 & 69.26 & \pos{2.12} & 80.78 & 78.83 & \pos{1.95} & 70.18 & 67.18 & \pos{3.00} & 85.67 & 84.48 & \pos{1.19} \\
GPT-OSS-20B$^*$ & 66.87 & 57.83 & \pos{9.04} & 57.95 & 43.82 & \pos{14.13} & 77.85 & 71.66 & \pos{6.19} & 63.80 & 53.94 & \pos{9.86} & 74.30 & 60.47 & \pos{13.83} \\
MedGemma-27B$^\dagger$ & 72.29 & 72.29 & 0.00 & 66.78 & 60.07 & \pos{6.71} & 80.35 & 77.09 & \pos{3.26} & 71.79 & 66.77 & \pos{5.03} & 83.80 & 77.52 & \pos{6.28} \\
Qwen3-4B-Thinking & 61.45 & 66.27 & \negative{-4.82} & 57.24 & 54.42 & \pos{2.83} & 79.26 & 75.68 & \pos{3.58} & 63.50 & 61.33 & \pos{2.18} & 75.66 & 73.37 & \pos{2.29} \\
Qwen3-4B-Thinking$^*$ & 60.84 & 54.22 & \pos{6.63} & 45.94 & 32.86 & \pos{13.07} & 74.92 & 61.89 & \pos{13.03} & 59.45 & 48.05 & \pos{11.40} & 65.06 & 46.06 & \pos{19.00} \\
Qwen2.5-3B$^\dagger$ & 48.19 & 37.35 & \pos{10.84} & 32.86 & 27.92 & \pos{4.95} & 69.38 & 50.05 & \pos{19.33} & 49.17 & 36.83 & \pos{12.34} & 51.57 & 33.08 & \pos{18.49} \\
\midrule
Mean (open-weight) & 71.51 & 69.62 & \pos{1.89} & 69.43 & 63.35 & \pos{6.08} & 81.28 & 77.11 & \pos{4.16} & 70.95 & 66.07 & \pos{4.87} & 83.02 & 76.88 & \pos{6.15} \\
\bottomrule
\end{tabular}
}
    \caption{Benchmark accuracy (\%) per dataset, part 1. $^\ast$Reasoning off. $^\dagger$Single-mode model.}
    \label{tab:benchmark_per_dataset_a}
\end{table}

\begin{table}[htb]
    \centering
    \resizebox{\textwidth}{!}{%
\begin{tabular}{lrrrrrrrrrrrrrrr}
\toprule
& \multicolumn{3}{c}{\textbf{MedXpertQA-R}} & \multicolumn{3}{c}{\textbf{MedXpertQA-U}} & \multicolumn{3}{c}{\textbf{MMLU}} & \multicolumn{3}{c}{\textbf{MMLU-Pro}} & \multicolumn{3}{c}{\textbf{PubMedQA}} \\
\cmidrule(lr){2-4}\cmidrule(lr){5-7}\cmidrule(lr){8-10}\cmidrule(lr){11-13}\cmidrule(lr){14-16}
\textbf{Model} & EN & SV & $\Delta$ & EN & SV & $\Delta$ & EN & SV & $\Delta$ & EN & SV & $\Delta$ & EN & SV & $\Delta$ \\
\midrule
GPT-5.4 & 57.76 & 57.01 & \pos{0.75} & 55.86 & 56.03 & \negative{-0.17} & 95.68 & 95.59 & \pos{0.09} & 81.17 & 81.66 & \negative{-0.49} & 77.60 & 75.40 & \pos{2.20} \\
Claude Sonnet 4.6 & 46.96 & 44.49 & \pos{2.47} & 46.86 & 42.95 & \pos{3.91} & 95.87 & 94.95 & \pos{0.92} & 81.91 & 81.91 & 0.00 & 73.00 & 70.20 & \pos{2.80} \\
\midrule
Gemma-4-31B & 58.98 & 55.50 & \pos{3.48} & 49.82 & 50.37 & \negative{-0.55} & 95.09 & 94.43 & \pos{0.66} & 79.69 & 79.18 & \pos{0.51} & 80.90 & 81.33 & \negative{-0.43} \\
Gemma-4-31B$^*$ & 54.77 & 47.27 & \pos{7.51} & 51.65 & 43.75 & \pos{7.90} & 94.15 & 93.96 & \pos{0.19} & 78.93 & 76.37 & \pos{2.55} & 79.83 & 80.26 & \negative{-0.43} \\
Qwen3.6-35B-A3B & 42.10 & 40.30 & \pos{1.80} & 37.87 & 34.74 & \pos{3.12} & 94.72 & 92.55 & \pos{2.17} & 78.29 & 76.50 & \pos{1.79} & 79.83 & 79.18 & \pos{0.64} \\
Qwen3.6-35B-A3B$^*$ & 37.30 & 30.63 & \pos{6.67} & 36.40 & 28.68 & \pos{7.72} & 91.89 & 91.51 & \pos{0.38} & 75.99 & 73.05 & \pos{2.94} & 78.97 & 80.26 & \negative{-1.29} \\
Qwen3.8-27B & 45.71 & 40.24 & \pos{5.47} & 35.85 & 32.17 & \pos{3.68} & 93.87 & 91.98 & \pos{1.89} & 77.52 & 74.58 & \pos{2.94} & 77.90 & 75.11 & \pos{2.79} \\
Qwen3.8-27B$^*$ & 36.22 & 29.85 & \pos{6.37} & 30.33 & 27.21 & \pos{3.12} & 91.04 & 89.06 & \pos{1.98} & 74.84 & 71.90 & \pos{2.94} & 78.97 & 80.04 & \negative{-1.07} \\
GPT-OSS-120B & 36.94 & 35.50 & \pos{1.44} & 35.11 & 30.88 & \pos{4.23} & 92.36 & 90.28 & \pos{2.08} & 74.58 & 74.71 & \negative{-0.13} & 74.46 & 75.11 & \negative{-0.64} \\
GPT-OSS-120B$^*$ & 24.02 & 18.02 & \pos{6.01} & 25.92 & 19.12 & \pos{6.80} & 88.58 & 82.64 & \pos{5.94} & 69.09 & 59.90 & \pos{9.20} & 80.04 & 76.18 & \pos{3.86} \\
GPT-OSS-20B & 27.21 & 23.72 & \pos{3.48} & 23.35 & 22.61 & \pos{0.74} & 88.68 & 88.40 & \pos{0.28} & 72.67 & 69.48 & \pos{3.19} & 77.90 & 77.68 & \pos{0.21} \\
GPT-OSS-20B$^*$ & 18.26 & 15.38 & \pos{2.88} & 20.96 & 16.36 & \pos{4.60} & 83.49 & 73.49 & \pos{10.00} & 64.75 & 51.60 & \pos{13.15} & 78.11 & 79.40 & \negative{-1.29} \\
MedGemma-27B$^\dagger$ & 22.28 & 20.18 & \pos{2.10} & 22.61 & 22.06 & \pos{0.55} & 86.13 & 83.11 & \pos{3.02} & 71.26 & 66.67 & \pos{4.60} & 77.68 & 72.96 & \pos{4.72} \\
Qwen3-4B-Thinking & 18.44 & 17.12 & \pos{1.32} & 18.01 & 18.01 & 0.00 & 85.85 & 84.06 & \pos{1.79} & 67.94 & 63.60 & \pos{4.34} & 73.82 & 76.61 & \negative{-2.79} \\
Qwen3-4B-Thinking$^*$ & 14.11 & 12.25 & \pos{1.86} & 14.89 & 12.32 & \pos{2.57} & 80.47 & 66.98 & \pos{13.49} & 59.26 & 40.49 & \pos{18.77} & 72.32 & 73.82 & \negative{-1.50} \\
Qwen2.5-3B$^\dagger$ & 12.55 & 10.33 & \pos{2.22} & 11.76 & 11.40 & \pos{0.37} & 68.30 & 44.81 & \pos{23.49} & 40.61 & 20.56 & \pos{20.05} & 69.10 & 68.03 & \pos{1.07} \\
\midrule
Mean (open-weight) & 32.06 & 28.31 & \pos{3.76} & 29.61 & 26.40 & \pos{3.20} & 88.19 & 83.38 & \pos{4.81} & 70.39 & 64.19 & \pos{6.20} & 77.13 & 76.85 & \pos{0.28} \\
\bottomrule
\end{tabular}
}
    \caption{Benchmark accuracy (\%) per dataset, part 2. $^\ast$Reasoning off. $^\dagger$Single-mode model.}
    \label{tab:benchmark_per_dataset_b}
\end{table}

\section{What Makes a Question Hard?}
\label{sec:appendix-hard}

\textit{MedicalAgentsBench} marks 894 test items as hard. Like Section~\ref{sec:datasets}, this appendix uses the 840 of them that are core items. We ask what sets these items apart and which kinds of questions are hard for the models in either language.

\paragraph{Labeling the questions.} We label every English test item with GPT-OSS-120B (greedy decoding, low reasoning effort, JSON output). The model sees the question and its options but not the answer. It assigns one of ten \emph{question types}, that is, what the question asks the reader to do, for example find the diagnosis or choose the next step; one of 17 \emph{specialties}; and three flags: whether the item is a clinical vignette, reports lab values, or needs a calculation. Appendix~\ref{sec:appendix-prompts} gives the prompt with all label definitions. We also compute simple features with code: question length, number of options, numbers in the question, and regular expressions for vignettes, lab values, and calculations. The code and the model flags agree on 93.6\% of items for vignettes and 85.9\% for lab values; most of the remaining difference for lab values is blood pressure, which the code counts and the prompt excludes.

\begin{figure}[htb]
    \centering
    \includegraphics[width=\textwidth]{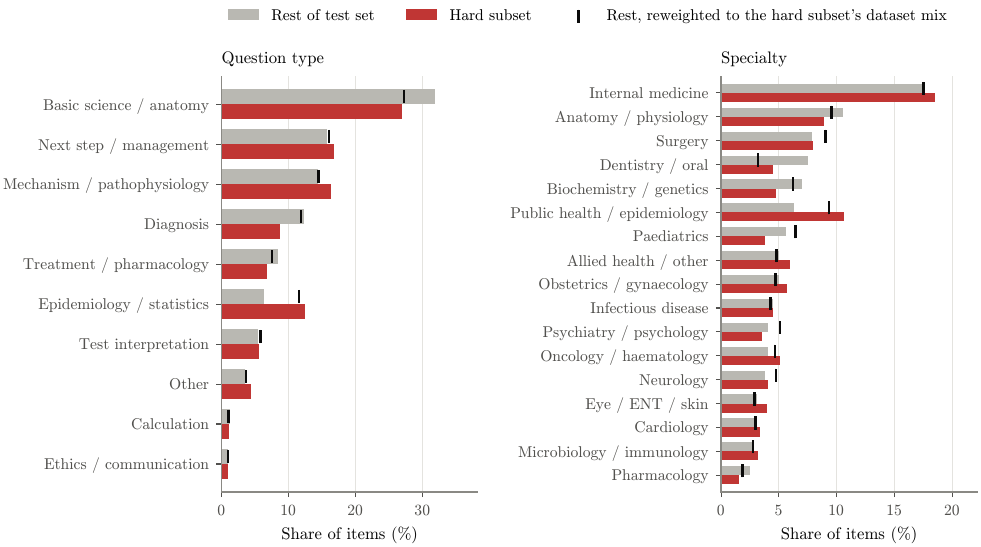}
    \caption{Share of question types (left) and specialties (right) in the hard subset (red) and in the rest of the test set (gray). Black ticks: the rest reweighted to the dataset mix of the hard subset. Labels assigned by GPT-OSS-120B.}
    \label{fig:hard_composition}
\end{figure}

\paragraph{The dataset mix explains most differences.} The hard subset has about 100 items per dataset, so its dataset mix differs strongly from that of the full test set. For a fair comparison, we also reweight the remaining items to the dataset mix of the hard subset; Table~\ref{tab:hard_features} shows them as ``rest, matched'', and Figure~\ref{fig:hard_composition} as ticks. Without this correction, hard questions look much longer, with a median of 61 against 27 words. With it, the difference disappears (65 words), and vignettes ($39.0\%$ against $42.8\%$) and lab values ($26.2\%$ against $25.3\%$) are just as common in both groups. The question types also differ little. Hard items contain somewhat fewer diagnosis questions (8.7\% against 11.8\%). The larger share of epidemiology and research questions (12.5\% against 6.3\%) comes from PubMedQA and nearly vanishes after matching (11.6\%). In short, a hard question looks like any other question from the same dataset; the features we measure do not explain why it is hard.

\begin{table}[htb]
    \centering
    \small
\begin{tabular}{lrrrr}
\toprule
\textbf{Feature} & \textbf{Hard} & \textbf{Rest} & \textbf{Rest, matched} & \textbf{All} \\
\midrule
Items & 840 & 8,893 & 8,893 & 9,733 \\
Median words in question & 61 & 27 & 65 & 28 \\
Mean number of options & 6.0 & 5.8 & 5.9 & 5.8 \\
Mean numbers in question & 5.4 & 3.8 & 5.2 & 3.9 \\
Clinical vignette, regex (\%) & 39.0 & 38.9 & 42.8 & 38.9 \\
Lab values, regex (\%) & 26.2 & 20.6 & 25.3 & 21.1 \\
Calculation, regex (\%) & 0.7 & 0.9 & 0.7 & 0.9 \\
Clinical vignette, LLM (\%) & 44.0 & 44.1 & 47.0 & 44.1 \\
Lab values, LLM (\%) & 9.5 & 9.3 & 9.8 & 9.3 \\
Calculation, LLM (\%) & 1.1 & 1.5 & 1.2 & 1.4 \\
Open-weight accuracy EN (\%) & 30.0 & 70.3 & 72.3 & 66.8 \\
Open-weight accuracy SV (\%) & 28.8 & 66.8 & 69.1 & 63.5 \\
EN $-$ SV gap (points) & \pos{1.2} & \pos{3.5} & \pos{3.2} & \pos{3.3} \\
Universally hard (\%) & 38.5 & 11.0 & 10.0 & 13.4 \\
\bottomrule
\end{tabular}

    \caption{Simple features of the hard subset and the rest of the test set. Rest, matched: the rest reweighted to the dataset mix of the hard subset. Accuracy: item-weighted mean over the 8 open configurations with reasoning on (or single mode). Universally hard: at most two of the 14 open configurations answer correctly, in both languages.}
    \label{tab:hard_features}
\end{table}

\paragraph{Hard in both languages.} The 8 open configurations with reasoning on, or in their single mode, answer 30.0\% of the hard items correctly in English and 28.8\% in Swedish, against 70.3\% and 66.8\% on the rest. The English--Swedish gap is 1.2 points on hard items and 3.5 points on the rest (3.2 after matching); over all 14 open configurations, with and without reasoning, it is 1.3 and 4.8 points, as Appendix~\ref{sec:appendix-benchmark} shows. Because the hard items are near chance and were chosen because models fail on them, their gaps are compressed by construction, so a small gap there is weak evidence. Language may matter little once a question is beyond the model; the gap appears on questions that a model can almost answer. On hard items, the gap is at most 2.4 points for every question type with more than 50 hard items; it is largest for treatment and pharmacology, where drug names and doses matter for translation.

\begin{figure}[htb]
    \centering
    \includegraphics[width=\textwidth]{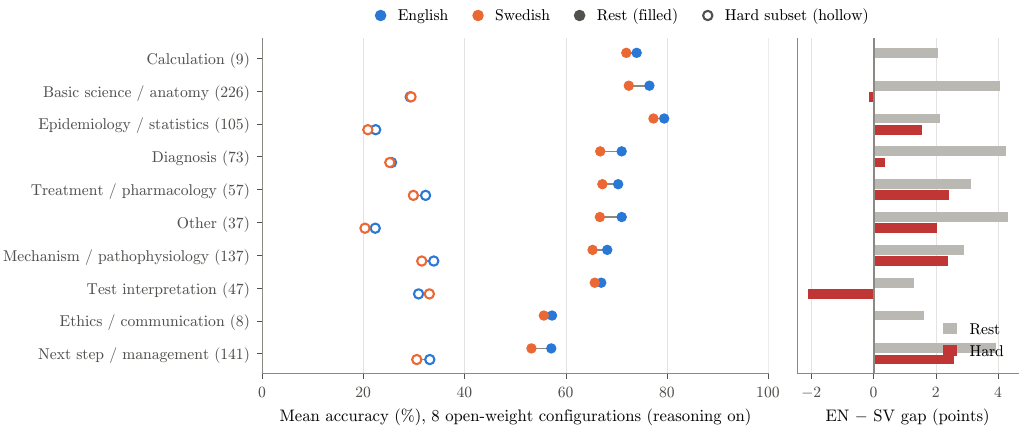}
    \caption{Left: mean accuracy of the open models with reasoning per question type, on the rest of the test set (filled) and on the hard subset (hollow), in English (blue) and Swedish (orange). The number after each type is its count in the hard subset; hard-subset points are hidden for types with fewer than 10 hard items. Right: English--Swedish gap.}
    \label{fig:hard_accuracy}
\end{figure}

\paragraph{Question types.} Table~\ref{tab:hard_summary} and Figure~\ref{fig:hard_accuracy} show accuracy by question type over the whole test set. Questions about the next step in management and about ethics and communication are the hardest on average; calculation, basic science, and epidemiology questions are the easiest. The language gap is largest for diagnosis, next-step, and basic-science questions, which may depend most on medical terms, and smallest for test interpretation, calculation, and epidemiology questions, which may depend more on reasoning and numbers.

\begin{table}[htb]
    \centering
    \small
\begin{tabular}{lrrrrrr}
\toprule
& & & \multicolumn{3}{c}{\textbf{Open-weight (\%)}} & \textbf{Student} \\
\cmidrule(lr){4-6}\cmidrule(lr){7-7}
\textbf{Question type} & $\boldsymbol{n}$ & \textbf{Hard (\%)} & \textbf{EN} & \textbf{SV} & $\boldsymbol{\Delta}$ & \textbf{SV (\%)} \\
\midrule
Basic science / anatomy & 3,056 & 7.4 & 72.9 & 69.2 & \pos{3.8} & 56.8 \\
Next step / management & 1,542 & 9.1 & 54.9 & 51.1 & \pos{3.8} & 38.5 \\
Mechanism / pathophysiology & 1,421 & 9.6 & 64.8 & 62.0 & \pos{2.8} & 51.6 \\
Diagnosis & 1,162 & 6.3 & 68.1 & 64.1 & \pos{4.0} & 57.8 \\
Treatment / pharmacology & 799 & 7.1 & 67.6 & 64.5 & \pos{3.1} & 57.4 \\
Epidemiology / statistics & 664 & 15.8 & 70.4 & 68.3 & \pos{2.0} & 54.1 \\
Test interpretation & 524 & 9.0 & 63.7 & 62.7 & \pos{1.0} & 51.5 \\
Other & 347 & 10.7 & 65.8 & 61.7 & \pos{4.1} & 42.9 \\
Calculation & 125 & 7.2 & 73.8 & 71.9 & \pos{1.9} & 44.0 \\
Ethics / communication & 93 & 8.6 & 55.6 & 53.2 & \pos{2.4} & 34.4 \\
\midrule
All items & 9,733 & 8.6 & 66.8 & 63.5 & \pos{3.3} & 52.0 \\
Hard subset & 840 & 100.0 & 30.0 & 28.8 & \pos{1.2} & 21.9 \\
\bottomrule
\end{tabular}

    \caption{Accuracy by question type. Hard (\%): share of the type's items in the hard subset. Open-weight: item-weighted mean over the 8 open configurations with reasoning on (or single mode). Student: the bilingual LFM2.5-1.2B reasoning-trace model on Swedish items.}
    \label{tab:hard_summary}
\end{table}

\paragraph{Universally hard items.} 1,300 test items are answered correctly by at most two of the 14 open configurations in both languages, and 530 by none. Most of them, 782, come from MedXpertQA. Only 323 of them are in the released hard subset, but they still make up 38.5\% of it, and only 103 hard items are solved by a majority of the configurations in both languages. The released subset is therefore a good, compact core of hard questions, but not the only one.

\paragraph{Students.} The best student, bilingual LFM2.5-1.2B with reasoning, answers 21.9\% of the hard items correctly in Swedish, close to the chance level of 20.4\% on these items, against 54.8\% on the rest. Its weakest question types are the same as for the large models: ethics and communication, and next-step management.

\section{Distillation Data}
\label{sec:appendix-distillation}

\begin{figure}[htb]
    \centering
    \includegraphics[width=\textwidth]{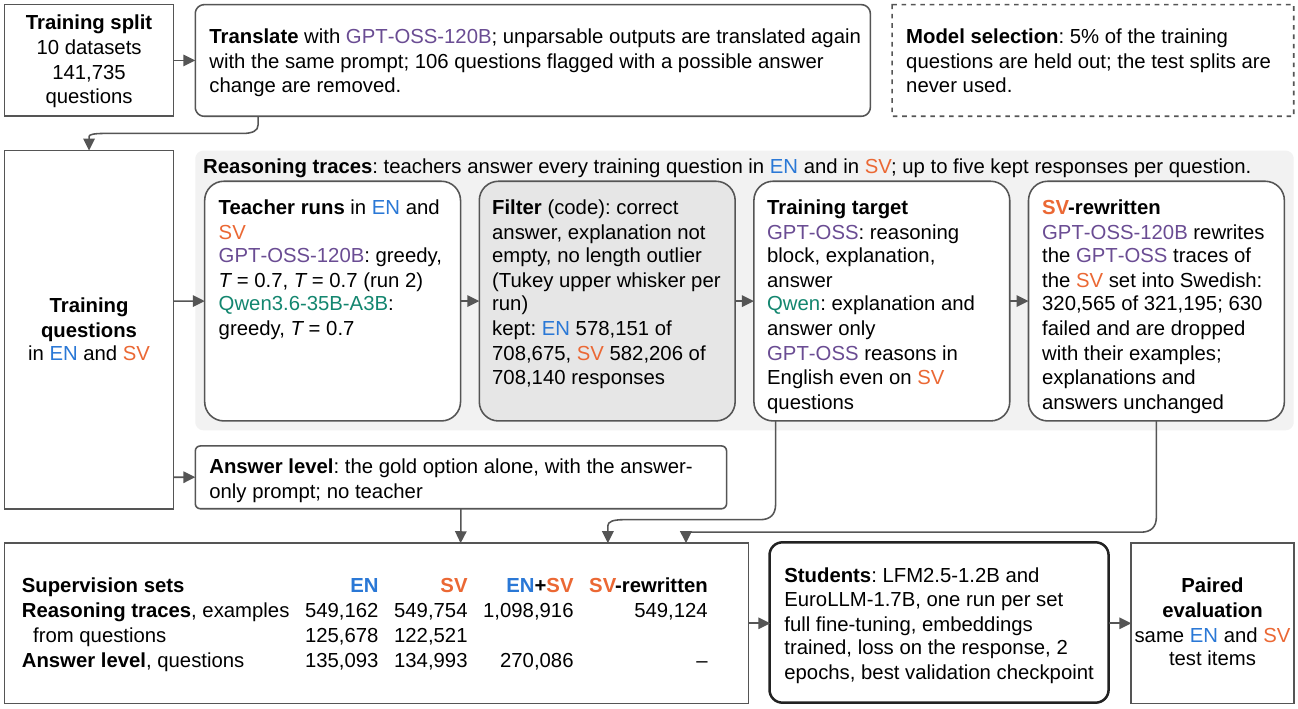}
    \caption{How we build the training data of the students. GPT-OSS-120B translates the training split. Two teachers answer every question in English and Swedish, and a code filter keeps correct answers of normal length. GPT-OSS traces keep their reasoning block; Qwen keeps only the explanation and the answer. A second Swedish set has the GPT-OSS reasoning rewritten into Swedish. Answer-level data needs no teacher. Each set trains both students, and all students are evaluated on the same paired test items.}
    \label{fig:part2_detail}
\end{figure}

Figure~\ref{fig:part2_detail} shows how we build the supervision for the students, with all counts. GPT-OSS-120B translates the 141,735 training questions, and unparsable translations are translated again with the same prompt. One question, medqa\_7440, fails in both passes, and we remove the 106 questions that the translator flagged with a \texttt{possible\_answer\_change\_risk} warning; dropping these 107 leaves 141,628. The teachers, chosen for their quality and cost, are GPT-OSS-120B, sampled greedily and twice at temperature 0.7, and Qwen3.6-35B-A3B, sampled greedily and once at temperature 0.7; only GPT-OSS keeps its reasoning block.

\begin{figure}[htb]
    \centering
    \includegraphics[width=\textwidth]{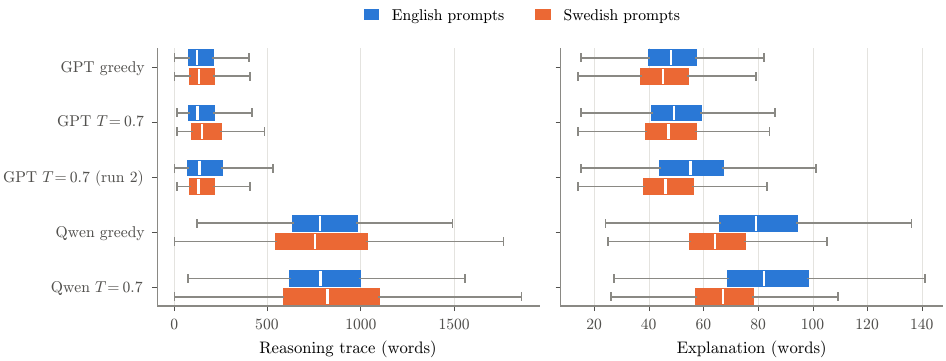}
    \caption{Length of teacher reasoning traces and explanations (words). Boxes: quartiles and median; whiskers: Tukey whiskers (the length limits).}
    \label{fig:length_distributions}
\end{figure}

\paragraph{Filters and set sizes.} We keep a response only if its answer matches the gold label and its explanation is not empty. We also drop length outliers: explanations, and GPT-OSS reasoning traces, longer than the Tukey upper whisker of their own teacher run. The limit is set per run because the teachers write at different lengths; Table~\ref{tab:teacher_stats} gives the teacher runs and their limits, and Figure~\ref{fig:length_distributions} the length distributions. Of 708,675 English responses, 578,151 pass the filters; of 708,140 Swedish responses, 582,206 pass. We use all kept responses, so a question can have up to five valid solutions. The English reasoning-trace set has 549,162 examples from 125,678 questions, and the Swedish set 549,754 examples from 122,521 questions. In both, about 58\% of the examples contain a reasoning trace. The bilingual set is their union of 1,098,916 examples, 85.2\% of them from MedMCQA. Table~\ref{tab:train_composition} gives the reasoning-trace and answer-level data per dataset.

\paragraph{SV-rewritten traces.} GPT-OSS-120B rewrote 320,565 of the 321,195 GPT-OSS traces in the Swedish set into Swedish. We drop the 630 failures, 598 of which reached the length limit, together with their examples. Explanations and answers stay unchanged, and the set has 549,124 examples.

\begin{table}[htb]
    \centering
    \small
    \resizebox{\textwidth}{!}{%
\begin{tabular}{llrrrrrr}
\toprule
& & & & & \multicolumn{2}{c}{\textbf{Upper whisker (words)}} & \\
\cmidrule(lr){6-7}
\textbf{Teacher} & \textbf{Decoding} & \textbf{Responses} & \textbf{Correct (\%)} & \textbf{Kept} & \textbf{Explanation} & \textbf{Reasoning} & \textbf{Median reasoning} \\
\midrule
\multicolumn{8}{l}{\textit{English prompts}} \\
GPT-OSS-120B & greedy & 141,735 & 86.3 & 114,537 & 82 & 401 & 121 \\
GPT-OSS-120B & $T{=}0.7$ & 141,735 & 86.3 & 114,860 & 86 & 414 & 122 \\
GPT-OSS-120B & $T{=}0.7$ (run 2) & 141,735 & 79.5 & 104,456 & 101 & 530 & 135 \\
Qwen3.6-35B-A3B & greedy & 141,735 & 88.2 & 122,242 & 136 & -- & 781 \\
Qwen3.6-35B-A3B & $T{=}0.7$ & 141,735 & 88.0 & 122,056 & 141 & -- & 784 \\
\midrule
\multicolumn{8}{l}{\textit{Swedish prompts}} \\
GPT-OSS-120B & greedy & 141,628 & 84.9 & 112,983 & 79 & 405 & 130 \\
GPT-OSS-120B & $T{=}0.7$ & 141,628 & 84.8 & 113,169 & 84 & 483 & 147 \\
GPT-OSS-120B & $T{=}0.7$ (run 2) & 141,628 & 85.0 & 113,700 & 83 & 405 & 129 \\
Qwen3.6-35B-A3B & greedy & 141,628 & 86.8 & 121,441 & 105 & -- & 754 \\
Qwen3.6-35B-A3B & $T{=}0.7$ & 141,628 & 86.6 & 120,913 & 109 & -- & 821 \\
\bottomrule
\end{tabular}
}
    \caption{Teacher runs on the training questions. Correct: share of responses with the gold answer. Kept: responses that pass all filters. Whiskers are the Tukey upper whiskers used as length limits; the reasoning limit applies only to GPT-OSS, whose traces are kept. Median reasoning: median length of the full teacher trace in words.}
    \label{tab:teacher_stats}
\end{table}

\begin{table}[htb]
    \centering
    \small
\begin{tabular}{lrrrrrr}
\toprule
& \multicolumn{2}{c}{\textbf{Answer-level (questions)}} & \multicolumn{3}{c}{\textbf{Reasoning-trace (examples)}} & \\
\cmidrule(lr){2-3}\cmidrule(lr){4-6}
\textbf{Dataset} & \textbf{EN} & \textbf{SV} & \textbf{EN} & \textbf{SV} & \textbf{SV-RW} & \textbf{EN+SV share (\%)} \\
\midrule
AfriMedQA & 117 & 117 & 455 & 454 & 453 & 0.083 \\
MedBullets & 9,684 & 9,683 & 38,775 & 40,379 & 40,261 & 7.203 \\
MedExQA & 19 & 19 & 82 & 84 & 84 & 0.015 \\
MedMCQA & 115,142 & 115,044 & 469,432 & 466,934 & 466,560 & 85.208 \\
MedQA & 9,685 & 9,684 & 38,731 & 40,246 & 40,109 & 7.187 \\
MedXpertQA-R & 5 & 5 & 8 & 9 & 9 & 0.002 \\
MedXpertQA-U & 5 & 5 & 7 & 8 & 8 & 0.001 \\
MMLU & 4 & 4 & 13 & 14 & 14 & 0.002 \\
MMLU-Pro & 4 & 4 & 13 & 15 & 15 & 0.003 \\
PubMedQA & 428 & 428 & 1,646 & 1,611 & 1,611 & 0.296 \\
\midrule
Total & 135,093 & 134,993 & 549,162 & 549,754 & 549,124 & 100.000 \\
\bottomrule
\end{tabular}

    \caption{Training data per dataset. Answer-level: questions; the Swedish ones passed the translation filter, and the bilingual answer-level set, their union, has 270,086. Reasoning-trace: examples (a question can have up to five). SV-RW: Swedish data with rewritten traces. Last column: share of the bilingual reasoning-trace data.}
    \label{tab:train_composition}
\end{table}

\section{Training and Model Selection}
\label{sec:appendix-training}
\label{sec:appendix-model-selection}

\paragraph{Settings.} We fine-tune all parameters with Unsloth\footnote{\url{https://github.com/unslothai/unsloth}}, TRL\footnote{\url{https://github.com/huggingface/trl}}, and DeepSpeed ZeRO-2~\citep{rajbhandari2020zero} on four NVIDIA H200 GPUs: per-device batch 32, no gradient accumulation (effective batch 128), padding-free packing, gradient checkpointing, bfloat16, learning rate $5\cdot10^{-5}$, cosine schedule, 5\% warm-up, weight decay 0.01, gradient clipping at 1.0, and up to two epochs. Every run in the main text uses seed 13; Appendix~\ref{sec:appendix-seeds} repeats two LFM2.5 runs with seeds 14 and 15. The loss is the negative log-likelihood of the response tokens, and the embeddings are trained. Every 200 steps we compute the loss on a fixed validation set (2,048 English and/or 3,072 Swedish examples; 3,063 for SV-rewritten) and keep the best checkpoint. Early stopping uses a patience of 4 evaluations for reasoning-trace runs. The answer-level runs used a patience of 10, except the LFM2.5 SV and EuroLLM EN+SV runs, which used 2.

\begin{figure}[htb]
    \centering
    \includegraphics[width=\textwidth]{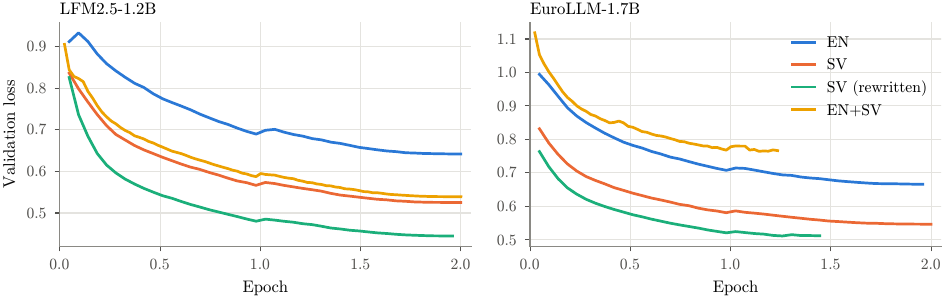}
    \caption{Validation loss of the final reasoning-trace runs. Each training language has its own validation set, so absolute values are comparable only within a language.}
    \label{fig:training_curves}
\end{figure}

\paragraph{Runs.} Table~\ref{tab:training_runs} and Figure~\ref{fig:training_curves} describe the final reasoning-trace runs. Six of them ran the full two epochs. Early stopping first ended both EN+SV runs at step 9,200, after 1.07 epochs. We resumed each from that checkpoint, which starts a new early-stopping counter: LFM2.5 EN+SV then ran to the last step, 17,170, and EuroLLM EN+SV stopped early again at step 10,600, after 1.23 epochs, with its best validation loss at step 9,800. Early stopping ended the EuroLLM SV-rewritten run after 1.44 epochs, and we did not resume it. The answer-level runs have 2,110 steps in English, 2,108 in Swedish, and 4,220 in EN+SV. Five of them ran the full two epochs. Their best validation loss came at step 2,000 for LFM2.5 EN, EuroLLM EN, and EuroLLM SV, at the last step, 2,108, for LFM2.5 SV, and at step 4,200 for LFM2.5 EN+SV. Early stopping ended the EuroLLM EN+SV answer-level run at step 2,400 of 4,220, after 1.14 epochs and two evaluations without improvement; its best step is 2,000.

\begin{table}[htb]
    \centering
    \small
    \resizebox{\textwidth}{!}{%
\begin{tabular}{llrrrrrl}
\toprule
\textbf{Student} & \textbf{Train} & \textbf{Examples} & \textbf{Epochs run} & \textbf{Last step} & \textbf{Best step} & \textbf{Val.\ loss} & \textbf{Note} \\
\midrule
LFM2.5-1.2B & EN & 549,162 & 2.00 & 8,580 & 8,580 & 0.6418 &  \\
LFM2.5-1.2B & SV & 549,754 & 2.00 & 8,588 & 8,588 & 0.5250 &  \\
LFM2.5-1.2B & SV (rewritten) & 549,124 & 2.00 & 8,580 & 8,400 & 0.4448 &  \\
LFM2.5-1.2B & EN+SV & 1,098,916 & 2.00 & 17,170 & 17,170 & 0.5391 & resumed after early stop at step 9,200 \\
EuroLLM-1.7B & EN & 549,162 & 2.00 & 8,580 & 8,400 & 0.6655 &  \\
EuroLLM-1.7B & SV & 549,754 & 2.00 & 8,588 & 8,400 & 0.5461 &  \\
EuroLLM-1.7B & SV (rewritten) & 549,124 & 1.44 & 6,200 & 5,400 & 0.5108 & early stop at epoch 1.44 \\
EuroLLM-1.7B & EN+SV & 1,098,916 & 1.23 & 10,600 & 9,800 & 0.7640 & resumed after early stop at step 9,200; early stop at epoch 1.23 \\
\bottomrule
\end{tabular}
}
    \caption{Final reasoning-trace training runs.}
    \label{tab:training_runs}
\end{table}

\paragraph{Model selection.} We chose the training setup with LFM2.5-1.2B on English reasoning-trace data, evaluated on the 6,642 held-out questions; Table~\ref{tab:model_selection} gives the results. We compared the training methods on an earlier version of the training data. There, full fine-tuning (76.3--76.5\%) is at least as good as the best LoRA run (76.1\%, rank 512), and LoRA gets steadily worse at lower ranks. We also tried the DFT loss, which \citet{wu2026generalization} report to generalize better than standard fine-tuning; with LoRA, it does worse than the standard loss at every rank. We therefore use full fine-tuning, which is affordable at this size and needs no search over ranks. On the final training data, frozen and trained embeddings work equally well in English (80.9\% and 80.8\%). On the Swedish held-out data, trained embeddings are better (78.7\% against 77.7\% with reasoning), so we train embeddings in all final runs. The base model reaches 47.3\% on the same held-out set. Held-out accuracy is much higher than test accuracy because the held-out questions come from the well-covered training datasets.

\begin{table}[htb]
    \centering
    \small
\begin{tabular}{llrrrrlrr}
\toprule
& & & & & & & \multicolumn{2}{c}{\textbf{Accuracy (\%)}} \\
\cmidrule(lr){8-9}
\textbf{Method} & \textbf{Loss} & \textbf{Rank} & \textbf{Batch} & \textbf{LR} & \textbf{Epochs} & \textbf{Embeddings} & \textbf{Reasoning} & \textbf{Direct} \\
\midrule
\multicolumn{7}{l}{Base model (default prompt)} & 47.31 & -- \\
\midrule
\multicolumn{9}{l}{\textit{Final training data}} \\
Full FT & NLL & -- & 128 & 5e-5 & 2 & frozen & 80.94 & 80.40 \\
Full FT$^\star$ & NLL & -- & 128 & 5e-5 & 2 & trained & 80.83 & 80.02 \\
\midrule
\multicolumn{9}{l}{\textit{Earlier version of the training data}} \\
Full FT & NLL & -- & 128 & 5e-5 & 2 & frozen & 76.47 & 75.59 \\
Full FT & NLL & -- & 128 & 5e-5 & 2 & trained & 76.26 & 75.64 \\
LoRA & NLL & 512 & 128 & 5e-5 & 2 & frozen & 76.11 & 74.57 \\
LoRA & NLL & 256 & 128 & 5e-5 & 2 & frozen & 72.39 & 70.75 \\
LoRA & NLL & 128 & 256 & 5e-5 & 2 & frozen & 66.62 & 64.57 \\
LoRA & DFT & 512 & 128 & 5e-5 & 2 & frozen & 61.40 & 61.08 \\
LoRA & DFT & 256 & 128 & 5e-5 & 2 & frozen & 61.11 & 60.95 \\
LoRA & DFT & 256 & 256 & 5e-5 & 2 & frozen & 59.52 & 58.99 \\
LoRA & DFT & 128 & 128 & 5e-5 & 2 & frozen & 59.18 & 58.43 \\
LoRA & DFT & 128 & 256 & 5e-5 & 2 & frozen & 58.31 & 56.68 \\
LoRA & DFT & 128 & 256 & 3e-5 & 2 & frozen & 55.53 & 54.38 \\
LoRA & DFT & 128 & 256 & 5e-5 & 1 & frozen & 54.00 & 53.79 \\
\bottomrule
\end{tabular}

    \caption{Model selection on the English held-out split (6,642 questions): LFM2.5-1.2B fine-tuned on English reasoning-trace data. The top block uses the final training data of the students; the bottom block compares training methods on an earlier version of the training data. Batch: effective batch size. $^\star$Selected setup.}
    \label{tab:model_selection}
\end{table}

\section{Additional Adaptation Results}
\label{sec:appendix-results}

\paragraph{All conditions.} Table~\ref{tab:main_adaptation_full} lists every student condition: direct and reasoning decoding for the reasoning-trace models, the gain over the base model, the share of invalid outputs, and the median number of generated tokens.

\begin{table}[htb]
    \centering
    \small
    \setlength{\tabcolsep}{4pt}
    \resizebox{\textwidth}{!}{%
\begin{tabular}{llrrrrrrrrr}
\toprule
& & \multicolumn{3}{c}{\textbf{Accuracy (\%)}} & \multicolumn{2}{c}{\textbf{Gain vs.\ base}} & \multicolumn{2}{c}{\textbf{Invalid (\%)}} & \multicolumn{2}{c}{\textbf{Median tokens}} \\
\cmidrule(lr){3-5}\cmidrule(lr){6-7}\cmidrule(lr){8-9}\cmidrule(lr){10-11}
\textbf{Model} & \textbf{Train} & \textbf{EN} & \textbf{SV} & $\boldsymbol{\Delta}$ & \textbf{EN} & \textbf{SV} & \textbf{EN} & \textbf{SV} & \textbf{EN} & \textbf{SV} \\
\midrule
\multicolumn{11}{l}{\textbf{LFM2.5-1.2B}} \\
\midrule
\multicolumn{11}{l}{\textit{Explanation prompt}} \\
Base model & -- & 37.24 & 23.49 & 13.76 & -- & -- & 0.9 & 0.4 & 113 & 143 \\
Trace SFT, direct & EN & 47.15 & 27.76 & 19.39 & +9.90 & +4.27 & 1.3 & 0.8 & 155 & 155 \\
Trace SFT, direct & SV & 44.35 & 43.53 & 0.82 & +7.11 & +20.05 & 1.1 & 0.0 & 281 & 293 \\
Trace SFT, direct & SV (rewritten) & 43.77 & 43.85 & -0.08 & +6.52 & +20.36 & 0.1 & 0.0 & 285 & 290 \\
Trace SFT, direct & EN+SV & 48.30 & 45.18 & 3.12 & +11.06 & +21.69 & 0.8 & 0.0 & 155 & 293 \\
\addlinespace[2pt]
Trace SFT, reasoning & EN & 49.05 & 14.52 & 34.53 & +11.81 & -8.97 & 4.6 & 57.2 & 485 & 11,375 \\
Trace SFT, reasoning & SV & 50.08 & 49.79 & 0.29 & +12.83 & +26.30 & 1.4 & 2.3 & 557 & 595 \\
Trace SFT, reasoning & SV (rewritten) & 43.99 & 44.32 & -0.33 & +6.75 & +20.84 & 0.4 & 0.4 & 584 & 590 \\
Trace SFT, reasoning & EN+SV & 50.51 & 51.97 & -1.46 & +13.26 & +28.48 & 6.0 & 2.1 & 486 & 603 \\
\addlinespace[2pt]
\multicolumn{11}{l}{\textit{Answer-only prompt}} \\
Base model & -- & 36.29 & 23.85 & 12.44 & -- & -- & 0.0 & 0.0 & 11 & 11 \\
Answer SFT & EN & 35.24 & 24.82 & 10.42 & -1.05 & +0.98 & 0.0 & 0.0 & 8 & 8 \\
Answer SFT & SV & 27.40 & 27.80 & -0.40 & -8.89 & +3.96 & 0.0 & 0.0 & 8 & 8 \\
Answer SFT & EN+SV & 33.17 & 30.67 & 2.50 & -3.12 & +6.82 & 0.0 & 0.0 & 8 & 8 \\
\midrule
\multicolumn{11}{l}{\textbf{EuroLLM-1.7B}} \\
\midrule
\multicolumn{11}{l}{\textit{Explanation prompt}} \\
Base model & -- & 17.48 & 23.49 & -6.01 & -- & -- & 29.3 & 2.7 & 77 & 48 \\
Trace SFT, direct & EN & 44.73 & 38.96 & 5.77 & +27.26 & +15.47 & 0.0 & 0.1 & 158 & 163 \\
Trace SFT, direct & SV & 41.49 & 40.91 & 0.58 & +24.01 & +17.43 & 0.6 & 0.1 & 193 & 298 \\
Trace SFT, direct & SV (rewritten) & 40.01 & 39.38 & 0.63 & +22.53 & +15.89 & 0.2 & 0.1 & 288 & 292 \\
Trace SFT, direct & EN+SV & 44.69 & 40.97 & 3.72 & +27.22 & +17.49 & 0.2 & 0.4 & 156 & 303 \\
\addlinespace[2pt]
Trace SFT, reasoning & EN & 43.19 & 39.44 & 3.75 & +25.72 & +15.96 & 11.6 & 12.0 & 478 & 505 \\
Trace SFT, reasoning & SV & 44.32 & 44.53 & -0.21 & +26.85 & +21.04 & 2.4 & 2.7 & 496 & 588 \\
Trace SFT, reasoning & SV (rewritten) & 39.43 & 38.96 & 0.47 & +21.96 & +15.47 & 0.9 & 1.0 & 558 & 568 \\
Trace SFT, reasoning & EN+SV & 41.90 & 44.46 & -2.56 & +24.42 & +20.97 & 16.5 & 7.8 & 470 & 585 \\
\addlinespace[2pt]
\multicolumn{11}{l}{\textit{Answer-only prompt}} \\
Base model & -- & 14.70 & 21.33 & -6.63 & -- & -- & 46.5 & 13.7 & 15 & 15 \\
Answer SFT & EN & 33.66 & 30.04 & 3.62 & +18.96 & +8.71 & 0.0 & 0.0 & 12 & 12 \\
Answer SFT & SV & 30.88 & 30.74 & 0.14 & +16.18 & +9.41 & 0.0 & 0.0 & 12 & 12 \\
Answer SFT & EN+SV & 31.63 & 29.44 & 2.20 & +16.93 & +8.11 & 0.0 & 0.0 & 12 & 12 \\
\bottomrule
\end{tabular}
}
    \caption{All adaptation results on the test set (9,733 core items per language). Accuracy is weighted by items; invalid outputs count as wrong. $\Delta=\mathrm{EN}-\mathrm{SV}$. Gains are relative to the base model with the same prompt.}
    \label{tab:main_adaptation_full}
\end{table}

\begin{figure}[htb]
    \centering
    \includegraphics[width=\textwidth]{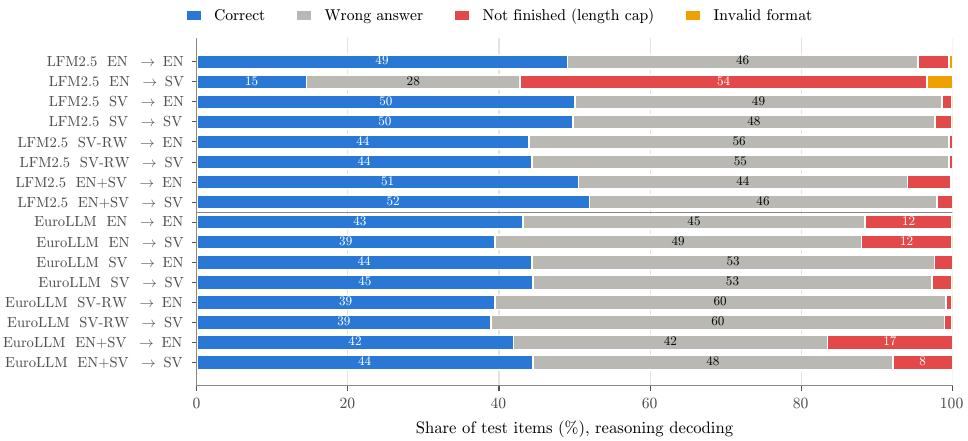}
    \caption{Outcome of every test item with reasoning decoding, per student, training language, and test language (``EN $\rightarrow$ SV'': trained on English, tested on Swedish). ``Not finished'': the length limit was reached before a valid answer.}
    \label{fig:output_status}
\end{figure}

\paragraph{Reasoning at inference.} Section~\ref{sec:analysis-language} compares direct and reasoning decoding, and Section~\ref{sec:analysis-english} discusses the unfinished outputs of the English-trained LFM2.5-1.2B and the bilingual EuroLLM-1.7B. Figure~\ref{fig:output_status} shows the outcome of every reasoning output. Besides these cases, the English-only EuroLLM-1.7B leaves 11.6\% of its English and 12.0\% of its Swedish reasoning outputs unfinished, against at most 2.7\% for the Swedish-trained model; unlike LFM2.5-1.2B, it fails at similar rates in both languages. Appendix~\ref{sec:appendix-eval} gives the token limits of the students.

\paragraph{Language of the reasoning.} Table~\ref{tab:reasoning_language} shows the language in which the students reason. Section~\ref{sec:analysis-language} shows that the SV-rewritten students, the only ones that reason almost entirely in Swedish, are less accurate in both languages than those trained on the original Swedish traces, while the students trained on the original Swedish or bilingual traces reason in English as their teacher did. Besides the second-hand copy of the teacher's reasoning noted there, a 1--2B model may simply know more medicine through English.

\begin{table}[htb]
    \centering
    \small
\begin{tabular}{llrr}
\toprule
& & \multicolumn{2}{c}{\textbf{Swedish reasoning traces (\%)}} \\
\cmidrule(lr){3-4}
\textbf{Student} & \textbf{Train} & \textbf{English test} & \textbf{Swedish test} \\
\midrule
LFM2.5-1.2B & EN & 0.0 & 17.4 \\
LFM2.5-1.2B & SV & 0.0 & 0.1 \\
LFM2.5-1.2B & SV (rewritten) & 100.0 & 100.0 \\
LFM2.5-1.2B & EN+SV & 0.0 & 0.1 \\
EuroLLM-1.7B & EN & 0.0 & 0.0 \\
EuroLLM-1.7B & SV & 0.0 & 0.3 \\
EuroLLM-1.7B & SV (rewritten) & 99.8 & 100.0 \\
EuroLLM-1.7B & EN+SV & 0.0 & 0.2 \\
\bottomrule
\end{tabular}

    \caption{Share of reasoning traces written in Swedish (reasoning decoding), decided by counting common Swedish and English function words.}
    \label{tab:reasoning_language}
\end{table}

\paragraph{Item-level agreement.} Table~\ref{tab:paired_stats} gives bootstrap intervals, the number of items solved in only one language, and McNemar tests for every student. For the models trained only on Swedish, the gap stays within about one point.

\paragraph{Comparisons between models.} Each condition in Table~\ref{tab:headline_tests} is one training run with seed 13, so its intervals measure the sampling of test items, not the variation between training runs; the next paragraphs repeat the comparisons that rest on the bilingual LFM2.5-1.2B student over three seeds. The comparisons are also post hoc: the bilingual LFM2.5-1.2B student is the best of the conditions on these items, and the bootstrap treats the 536 near-duplicate MMLU and MMLU-Pro pairs as independent. Table~\ref{tab:headline_tests} tests the comparisons that the main text relies on, with paired bootstrap intervals and McNemar tests on the same core items and the Holm correction over the 12 rows. The main comparison, in which the bilingual LFM2.5-1.2B student exceeds GPT-OSS-20B without reasoning in Swedish by 1.3 points, has a Holm-corrected $p$ of 0.064 over the 12 rows of this table and of 0.097 over the 12 rows of Table~\ref{tab:seeds_tests}, so it is not significant after the correction, and neither is the lead with seed 14 or 15.

\begin{table}[htb]
    \centering
    \footnotesize
    \setlength{\tabcolsep}{3pt}
    \resizebox{\textwidth}{!}{%
\begin{tabular}{llcrrrrrr}
\toprule
\textbf{A} & \textbf{B} & \textbf{Lang.} & \textbf{A} & \textbf{B} & \textbf{A $-$ B} & \textbf{95\% CI} & \textbf{McNemar $p$} & \textbf{Holm $p$} \\
\midrule
LFM2.5-1.2B, EN+SV traces & GPT-OSS-20B, reasoning off & SV & 51.97 & 50.64 & +1.33 & [0.3, 2.4] & 0.016 & 0.064 \\
LFM2.5-1.2B, EN+SV traces & Qwen3-4B-Thinking, reasoning on & SV & 51.97 & 57.43 & -5.47 & [-6.5, -4.4] & $<$0.001 & $<$0.001 \\
\midrule
LFM2.5-1.2B, EN+SV traces & LFM2.5-1.2B, SV traces & SV & 51.97 & 49.79 & +2.18 & [1.3, 3.1] & $<$0.001 & $<$0.001 \\
LFM2.5-1.2B, EN+SV traces & LFM2.5-1.2B, SV traces & EN & 50.51 & 50.08 & +0.43 & [-0.5, 1.3] & 0.359 & 0.980 \\
LFM2.5-1.2B, SV traces & LFM2.5-1.2B, SV (rewritten) traces & SV & 49.79 & 44.32 & +5.47 & [4.5, 6.4] & $<$0.001 & $<$0.001 \\
LFM2.5-1.2B, SV traces & LFM2.5-1.2B, SV (rewritten) traces & EN & 50.08 & 43.99 & +6.08 & [5.1, 7.1] & $<$0.001 & $<$0.001 \\
EuroLLM-1.7B, SV traces & EuroLLM-1.7B, SV (rewritten) traces & SV & 44.53 & 38.96 & +5.57 & [4.5, 6.6] & $<$0.001 & $<$0.001 \\
EuroLLM-1.7B, SV traces & EuroLLM-1.7B, SV (rewritten) traces & EN & 44.32 & 39.43 & +4.89 & [3.9, 5.9] & $<$0.001 & $<$0.001 \\
EuroLLM-1.7B, SV traces & EuroLLM-1.7B, EN+SV traces & SV & 44.53 & 44.46 & +0.07 & [-0.8, 1.0] & 0.896 & 0.980 \\
LFM2.5-1.2B, EN+SV traces & LFM2.5-1.2B, EN+SV answer SFT & SV & 51.97 & 30.67 & +21.30 & [20.1, 22.5] & $<$0.001 & $<$0.001 \\
\midrule
LFM2.5-1.2B, SV traces, reasoning & same model, direct & SV & 49.79 & 43.53 & +6.26 & [5.3, 7.3] & $<$0.001 & $<$0.001 \\
LFM2.5-1.2B, SV (rewritten), reasoning & same model, direct & SV & 44.32 & 43.85 & +0.47 & [-0.4, 1.4] & 0.327 & 0.980 \\
\bottomrule
\end{tabular}
}
    \caption{Paired comparisons behind the main-text claims, on the 9,733 core items: accuracy (\%) of models A and B, their difference, its 95\% paired bootstrap interval (10,000 resamples of items), the exact McNemar test, and the Holm-corrected $p$ over the rows of this table. Students use reasoning decoding unless stated.}
    \label{tab:headline_tests}
\end{table}

\paragraph{Variation between training seeds.}\label{sec:appendix-seeds} We repeated the two best LFM2.5-1.2B trace conditions, Swedish and bilingual traces, with seeds 14 and 15, which change the data order and the other random state of training; every other setting is that of seed 13. All three seeds start from the same pretrained weights. Seeds 14 and 15 are evaluated in English with a later version of our evaluation code with the same settings, while seed 13 keeps its original English evaluation; all Swedish and exam-set results of the three seeds use the same code, so the English differences between seeds also include this change. The Swedish runs ran the full two epochs. Like seed 13, both bilingual runs were first ended by early stopping at step 9,200. We evaluated these early-stopped checkpoints, which reached 46.3\% on the core Swedish items with both seeds, and then resumed both runs from step 9,200 to follow the procedure of seed 13. Unlike seed 13, which ran to step 17,170, they stopped early again: seed 14 at step 16,600, after 1.93 epochs, with its best checkpoint at step 15,800, and seed 15 at step 15,800, after 1.84 epochs, with its best checkpoint at step 15,000. At these steps the cosine schedule has lowered the learning rate below 5\% of its peak. Table~\ref{tab:seeds} gives the accuracy of the six runs. With reasoning decoding, the three seeds of each condition differ by at most 0.4 points on the core items in Swedish and 1.0 point in English, and the seed-13 runs are within 0.2 points of the mean of their condition in Swedish. The exam sets written in Swedish vary more, by up to 1.5 points on MedQA-SWE and 4.4 points on SMDT, and direct decoding varies by up to 6.9 points on SMDT.

Table~\ref{tab:seeds_tests} repeats, for each seed, the comparisons of the main text that rest on this student, with the Holm correction over its 12 rows. Bilingual traces are above the Swedish traces of the same seed on the core Swedish items with every seed, by 1.9 to 2.5 points, and the difference is significant after the correction; in English, they are 0.4 to 1.2 points above, and no seed is significant after the correction. On MedQA-SWE, the lead of bilingual traces is 3.2 and 2.8 points with seeds 13 and 14, both significant, but 0.5 points with seed 15, and on SMDT, Swedish traces are above bilingual traces with seed 14, with 56.0\% against 50.9\%. The bilingual student is above GPT-OSS-20B without reasoning in Swedish with every seed, by 1.0 to 1.3 points, but the paired interval excludes zero only for seeds 13 and 14, and no seed is significant after the correction. The Holm correction multiplies the $k$-th smallest of 12 $p$-values by $13-k$: the uncorrected $p$ of 0.016 for seed 13 is multiplied by 4 in Table~\ref{tab:headline_tests}, where eight tests have smaller $p$-values, and by 6 here, where six do, which gives 0.064 and 0.097. The lead has the same sign with every seed, and the three seeds differ by at most 0.3 points in Swedish, so its uncertainty comes mostly from the sampling of test items, not from training.

\begin{table}[htb]
    \centering
    \small
\begin{tabular}{llrrrrrr}
\toprule
\textbf{Trained on} & \textbf{Test set} & \textbf{Seed 13} & \textbf{Seed 14} & \textbf{Seed 15} & \textbf{Mean} & \textbf{SD} & \textbf{Range} \\
\midrule
\multicolumn{8}{l}{\textit{Reasoning decoding}} \\
SV & Core EN & 50.1 & 50.2 & 49.8 & 50.0 & 0.2 & 0.4 \\
 & Core SV & 49.8 & 49.4 & 49.8 & 49.6 & 0.2 & 0.4 \\
 & MedQA-SWE & 61.5 & 61.8 & 62.8 & 62.0 & 0.6 & 1.2 \\
 & SMDT & 51.6 & 56.0 & 51.6 & 53.0 & 2.5 & 4.4 \\
EN+SV & Core EN & 50.5 & 51.3 & 50.3 & 50.7 & 0.5 & 1.0 \\
 & Core SV & 52.0 & 51.8 & 51.7 & 51.8 & 0.1 & 0.3 \\
 & MedQA-SWE & 64.8 & 64.6 & 63.2 & 64.2 & 0.9 & 1.5 \\
 & SMDT & 54.1 & 50.9 & 52.8 & 52.6 & 1.6 & 3.1 \\
\midrule
\multicolumn{8}{l}{\textit{Direct decoding}} \\
SV & Core EN & 44.4 & 45.0 & 45.5 & 44.9 & 0.6 & 1.1 \\
 & Core SV & 43.5 & 44.5 & 43.6 & 43.9 & 0.5 & 0.9 \\
 & MedQA-SWE & 53.9 & 54.2 & 55.0 & 54.3 & 0.6 & 1.1 \\
 & SMDT & 45.3 & 46.5 & 39.6 & 43.8 & 3.7 & 6.9 \\
EN+SV & Core EN & 48.3 & 48.9 & 49.4 & 48.9 & 0.6 & 1.1 \\
 & Core SV & 45.2 & 45.6 & 45.6 & 45.5 & 0.3 & 0.5 \\
 & MedQA-SWE & 56.6 & 55.9 & 55.7 & 56.1 & 0.4 & 0.8 \\
 & SMDT & 49.1 & 47.8 & 43.4 & 46.8 & 3.0 & 5.7 \\
\bottomrule
\end{tabular}

    \caption{Accuracy (\%) of the LFM2.5-1.2B trace students on Swedish (SV) and bilingual (EN+SV) traces for three training seeds. Seed 13 is the run reported everywhere else. Core: the 9,733 core items per language; MedQA-SWE and SMDT: the exam sets written in Swedish. SD: sample standard deviation over the three seeds; Range: highest minus lowest accuracy.}
    \label{tab:seeds}
\end{table}

\begin{table}[htb]
    \centering
    \footnotesize
    \setlength{\tabcolsep}{3pt}
    \resizebox{\textwidth}{!}{%
\begin{tabular}{llcrrrrrr}
\toprule
\textbf{Compared with (B)} & \textbf{Test set} & \textbf{Seed} & \textbf{A} & \textbf{B} & \textbf{A $-$ B} & \textbf{95\% CI} & \textbf{McNemar $p$} & \textbf{Holm $p$} \\
\midrule
SV traces, same seed & Core SV & 13 & 51.97 & 49.79 & +2.18 & [1.3, 3.1] & $<$0.001 & $<$0.001 \\
 &  & 14 & 51.84 & 49.37 & +2.48 & [1.6, 3.4] & $<$0.001 & $<$0.001 \\
 &  & 15 & 51.68 & 49.79 & +1.89 & [1.0, 2.8] & $<$0.001 & $<$0.001 \\
\midrule
SV traces, same seed & Core EN & 13 & 50.51 & 50.08 & +0.43 & [-0.5, 1.3] & 0.359 & 0.719 \\
 &  & 14 & 51.33 & 50.16 & +1.17 & [0.3, 2.1] & 0.012 & 0.083 \\
 &  & 15 & 50.34 & 49.77 & +0.58 & [-0.3, 1.5] & 0.219 & 0.658 \\
\midrule
SV traces, same seed & MedQA-SWE & 13 & 64.79 & 61.54 & +3.25 & [1.6, 4.9] & $<$0.001 & 0.001 \\
 &  & 14 & 64.63 & 61.82 & +2.81 & [1.1, 4.5] & 0.001 & 0.011 \\
 &  & 15 & 63.24 & 62.77 & +0.47 & [-1.2, 2.1] & 0.602 & 0.719 \\
\midrule
GPT-OSS-20B, reasoning off & Core SV & 13 & 51.97 & 50.64 & +1.33 & [0.3, 2.4] & 0.016 & 0.097 \\
 &  & 14 & 51.84 & 50.64 & +1.20 & [0.1, 2.3] & 0.029 & 0.147 \\
 &  & 15 & 51.68 & 50.64 & +1.04 & [-0.1, 2.1] & 0.064 & 0.256 \\
\bottomrule
\end{tabular}
}
    \caption{The paired comparisons of the main text that rest on the bilingual LFM2.5-1.2B student, repeated for each training seed with reasoning decoding. A: the bilingual trace student of the given seed; B: the model named in the first column, the Swedish trace student of the same seed or GPT-OSS-20B without reasoning, the same run for every seed. Accuracy (\%) of A and B, their difference, its unadjusted 95\% paired bootstrap interval (10,000 resamples of items), the exact McNemar test, and the Holm-corrected $p$ over the 12 rows of this table.}
    \label{tab:seeds_tests}
\end{table}

\paragraph{Valid outputs and chance.} An invalid output counts as wrong, so a low score can mean that a model does not finish or does not follow the format, not that it lacks the knowledge. Table~\ref{tab:validity} separates the two. The base EuroLLM-1.7B leaves 29.3\% of its English outputs invalid with the explanation prompt and 46.5\% with the answer-only prompt, so its English accuracy, 17.5\% and 14.7\%, is below the chance level of 20.7\%; on its valid outputs it reaches 24.7\% and 27.5\%. Part of the gain of the trained EuroLLM-1.7B students is therefore learning to answer in the format. LFM2.5-1.2B trained on English traces leaves 57.2\% of its Swedish reasoning outputs invalid and is correct on 33.9\% of its valid ones. The answer-level students produce no invalid outputs.

\begin{table}[htb]
    \centering
    \small
\begin{tabular}{llrrrrrr}
\toprule
& & \multicolumn{3}{c}{\textbf{EN}} & \multicolumn{3}{c}{\textbf{SV}} \\
\cmidrule(lr){3-5}\cmidrule(lr){6-8}
\textbf{Model} & \textbf{Train} & \textbf{Acc.} & \textbf{Invalid} & \textbf{Valid acc.} & \textbf{Acc.} & \textbf{Invalid} & \textbf{Valid acc.} \\
\midrule
\multicolumn{8}{l}{\textbf{LFM2.5-1.2B}} \\
Base, explanation prompt & -- & 37.2 & 0.9 & 37.6 & 23.5 & 0.4 & 23.6 \\
Trace SFT, direct & EN & 47.1 & 1.3 & 47.8 & 27.8 & 0.8 & 28.0 \\
Trace SFT, direct & SV & 44.4 & 1.1 & 44.9 & 43.5 & 0.0 & 43.5 \\
Trace SFT, direct & SV (rewritten) & 43.8 & 0.1 & 43.8 & 43.9 & 0.0 & 43.9 \\
Trace SFT, direct & EN+SV & 48.3 & 0.8 & 48.7 & 45.2 & 0.0 & 45.2 \\
Trace SFT, reasoning & EN & 49.0 & 4.6 & 51.4 & 14.5 & 57.2 & 33.9 \\
Trace SFT, reasoning & SV & 50.1 & 1.4 & 50.8 & 49.8 & 2.3 & 51.0 \\
Trace SFT, reasoning & SV (rewritten) & 44.0 & 0.4 & 44.2 & 44.3 & 0.4 & 44.5 \\
Trace SFT, reasoning & EN+SV & 50.5 & 6.0 & 53.7 & 52.0 & 2.1 & 53.1 \\
Base, answer-only prompt & -- & 36.3 & 0.0 & 36.3 & 23.8 & 0.0 & 23.8 \\
Answer SFT & EN & 35.2 & 0.0 & 35.2 & 24.8 & 0.0 & 24.8 \\
Answer SFT & SV & 27.4 & 0.0 & 27.4 & 27.8 & 0.0 & 27.8 \\
Answer SFT & EN+SV & 33.2 & 0.0 & 33.2 & 30.7 & 0.0 & 30.7 \\
\midrule
\multicolumn{8}{l}{\textbf{EuroLLM-1.7B}} \\
Base, explanation prompt & -- & 17.5 & 29.3 & 24.7 & 23.5 & 2.7 & 24.1 \\
Trace SFT, direct & EN & 44.7 & 0.0 & 44.7 & 39.0 & 0.1 & 39.0 \\
Trace SFT, direct & SV & 41.5 & 0.6 & 41.8 & 40.9 & 0.1 & 41.0 \\
Trace SFT, direct & SV (rewritten) & 40.0 & 0.2 & 40.1 & 39.4 & 0.1 & 39.4 \\
Trace SFT, direct & EN+SV & 44.7 & 0.2 & 44.8 & 41.0 & 0.4 & 41.1 \\
Trace SFT, reasoning & EN & 43.2 & 11.6 & 48.8 & 39.4 & 12.0 & 44.8 \\
Trace SFT, reasoning & SV & 44.3 & 2.4 & 45.4 & 44.5 & 2.7 & 45.8 \\
Trace SFT, reasoning & SV (rewritten) & 39.4 & 0.9 & 39.8 & 39.0 & 1.0 & 39.4 \\
Trace SFT, reasoning & EN+SV & 41.9 & 16.5 & 50.2 & 44.5 & 7.8 & 48.2 \\
Base, answer-only prompt & -- & 14.7 & 46.5 & 27.5 & 21.3 & 13.7 & 24.7 \\
Answer SFT & EN & 33.7 & 0.0 & 33.7 & 30.0 & 0.0 & 30.0 \\
Answer SFT & SV & 30.9 & 0.0 & 30.9 & 30.7 & 0.0 & 30.7 \\
Answer SFT & EN+SV & 31.6 & 0.0 & 31.6 & 29.4 & 0.0 & 29.4 \\
\midrule
Chance (uniform guess) & & 20.7 & & & 20.7 & & \\
\bottomrule
\end{tabular}

    \caption{Accuracy (\%), share of invalid outputs (\%), unfinished at the token limit or not matching the schema, and accuracy on the valid outputs only, for every student condition and base model. Chance: the mean of $1/k$ over the core items, with $k$ options.}
    \label{tab:validity}
\end{table}

\begin{figure}[htb]
    \centering
    \includegraphics[width=\textwidth]{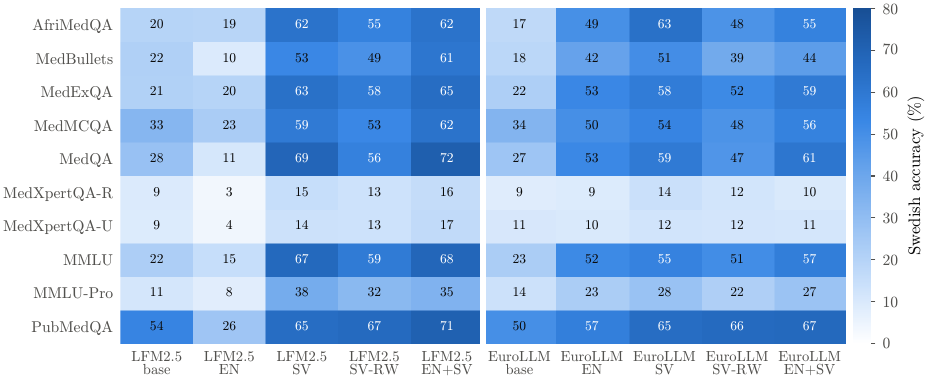}
    \caption{Swedish accuracy (\%) per dataset for the base explanation prompt and each reasoning-trace checkpoint (reasoning decoding).}
    \label{fig:dataset_accuracy}
\end{figure}

\paragraph{Hard questions and datasets.} Section~\ref{sec:analysis-gains} shows that the gains of the bilingual LFM2.5-1.2B model reach test sets with almost no training data but not the hard items. Table~\ref{tab:hard_subset} gives the hard items for all students. Figure~\ref{fig:dataset_accuracy} shows Swedish accuracy per dataset, and Tables~\ref{tab:per_dataset_lfm_reasoning} to~\ref{tab:per_dataset_eurollm_answer} give both languages for every condition. The students trained on Swedish or bilingual traces improve in Swedish on every dataset, but on MedXpertQA all of them stay below 17\%. Table~\ref{tab:train_weighted} weights each dataset by its share of the training data instead of the test data; with this weighting, the bilingual LFM2.5-1.2B model reaches 60.4\% in English and 62.8\% in Swedish.

\begin{table}[htb]
    \centering
    \resizebox{\textwidth}{!}{%
\begin{tabular}{llrrrrrrrr}
\toprule
& & \multicolumn{2}{c}{\textbf{Accuracy, 95\% CI}} & \multicolumn{2}{c}{\textbf{EN $-$ SV}} & \multicolumn{2}{c}{\textbf{Solved in only one language}} & & \\
\cmidrule(lr){3-4}\cmidrule(lr){5-6}\cmidrule(lr){7-8}
\textbf{Model} & \textbf{Train} & \textbf{EN} & \textbf{SV} & \textbf{Gap} & \textbf{95\% CI} & \textbf{EN only} & \textbf{SV only} & \textbf{McNemar $p$} & \textbf{Holm $p$} \\
\midrule
\multicolumn{10}{l}{\textbf{LFM2.5-1.2B}} \\
Base, explanation prompt & -- & [36.3, 38.2] & [22.7, 24.3] & +13.76 & [12.6, 14.9] & 2,419 & 1,080 & $<$0.001 & $<$0.001 \\
Trace SFT, direct & EN & [46.2, 48.1] & [26.9, 28.7] & +19.39 & [18.2, 20.5] & 2,781 & 894 & $<$0.001 & $<$0.001 \\
Trace SFT, direct & SV & [43.3, 45.3] & [42.6, 44.5] & +0.82 & [-0.2, 1.8] & 1,191 & 1,111 & 0.100 & 0.996 \\
Trace SFT, direct & SV (rewritten) & [42.8, 44.7] & [42.9, 44.8] & -0.08 & [-1.0, 0.8] & 1,073 & 1,081 & 0.880 & 1.000 \\
Trace SFT, direct & EN+SV & [47.3, 49.3] & [44.2, 46.2] & +3.12 & [2.2, 4.1] & 1,273 & 969 & $<$0.001 & $<$0.001 \\
Trace SFT, reasoning & EN & [48.1, 50.1] & [13.8, 15.2] & +34.53 & [33.4, 35.6] & 3,720 & 359 & $<$0.001 & $<$0.001 \\
Trace SFT, reasoning & SV & [49.1, 51.1] & [48.8, 50.8] & +0.29 & [-0.6, 1.2] & 1,024 & 996 & 0.548 & 1.000 \\
Trace SFT, reasoning & SV (rewritten) & [43.0, 45.0] & [43.3, 45.3] & -0.33 & [-1.3, 0.6] & 1,113 & 1,145 & 0.514 & 1.000 \\
Trace SFT, reasoning & EN+SV & [49.5, 51.5] & [51.0, 53.0] & -1.46 & [-2.4, -0.6] & 876 & 1,018 & 0.001 & 0.013 \\
Base, answer-only prompt & -- & [35.4, 37.2] & [23.0, 24.7] & +12.44 & [11.3, 13.6] & 2,263 & 1,052 & $<$0.001 & $<$0.001 \\
Answer SFT & EN & [34.3, 36.2] & [24.0, 25.7] & +10.42 & [9.3, 11.5] & 2,029 & 1,015 & $<$0.001 & $<$0.001 \\
Answer SFT & SV & [26.5, 28.3] & [26.9, 28.7] & -0.40 & [-1.4, 0.6] & 1,143 & 1,182 & 0.431 & 1.000 \\
Answer SFT & EN+SV & [32.2, 34.1] & [29.8, 31.6] & +2.50 & [1.5, 3.5] & 1,293 & 1,050 & $<$0.001 & $<$0.001 \\
\midrule
\multicolumn{10}{l}{\textbf{EuroLLM-1.7B}} \\
Base, explanation prompt & -- & [16.7, 18.2] & [22.7, 24.3] & -6.01 & [-6.7, -5.3] & 375 & 960 & $<$0.001 & $<$0.001 \\
Trace SFT, direct & EN & [43.8, 45.7] & [38.0, 39.9] & +5.77 & [4.8, 6.8] & 1,556 & 994 & $<$0.001 & $<$0.001 \\
Trace SFT, direct & SV & [40.5, 42.5] & [39.9, 41.9] & +0.58 & [-0.4, 1.6] & 1,243 & 1,187 & 0.265 & 1.000 \\
Trace SFT, direct & SV (rewritten) & [39.0, 41.0] & [38.4, 40.3] & +0.63 & [-0.4, 1.6] & 1,201 & 1,140 & 0.215 & 1.000 \\
Trace SFT, direct & EN+SV & [43.7, 45.7] & [40.0, 42.0] & +3.72 & [2.7, 4.7] & 1,413 & 1,051 & $<$0.001 & $<$0.001 \\
Trace SFT, reasoning & EN & [42.2, 44.2] & [38.5, 40.4] & +3.75 & [2.8, 4.7] & 1,399 & 1,034 & $<$0.001 & $<$0.001 \\
Trace SFT, reasoning & SV & [43.3, 45.3] & [43.5, 45.5] & -0.21 & [-1.2, 0.7] & 1,071 & 1,091 & 0.683 & 1.000 \\
Trace SFT, reasoning & SV (rewritten) & [38.5, 40.4] & [38.0, 39.9] & +0.47 & [-0.6, 1.5] & 1,336 & 1,290 & 0.380 & 1.000 \\
Trace SFT, reasoning & EN+SV & [40.9, 42.9] & [43.5, 45.4] & -2.56 & [-3.5, -1.6] & 923 & 1,172 & $<$0.001 & $<$0.001 \\
Base, answer-only prompt & -- & [14.0, 15.4] & [20.5, 22.1] & -6.63 & [-7.2, -6.0] & 160 & 805 & $<$0.001 & $<$0.001 \\
Answer SFT & EN & [32.7, 34.6] & [29.1, 31.0] & +3.62 & [2.7, 4.6] & 1,274 & 922 & $<$0.001 & $<$0.001 \\
Answer SFT & SV & [30.0, 31.8] & [29.8, 31.7] & +0.14 & [-0.8, 1.1] & 1,107 & 1,093 & 0.782 & 1.000 \\
Answer SFT & EN+SV & [30.7, 32.6] & [28.5, 30.4] & +2.20 & [1.3, 3.1] & 1,085 & 871 & $<$0.001 & $<$0.001 \\
\bottomrule
\end{tabular}
}
    \caption{Paired statistics for each student configuration: 95\% bootstrap intervals for accuracy (\%) and for the English--Swedish gap, the number of items solved in only one language, and the exact McNemar test, with the Holm correction over all student tests in the last column.}
    \label{tab:paired_stats}
\end{table}

\begin{table}[htb]
    \centering
    \resizebox{\textwidth}{!}{%
\begin{tabular}{llrrrrrr}
\toprule
& & \multicolumn{2}{c}{\textbf{Hard (840)}} & \multicolumn{2}{c}{\textbf{Remainder (8,893)}} & \multicolumn{2}{c}{\textbf{Hard $-$ remainder}} \\
\cmidrule(lr){3-4}\cmidrule(lr){5-6}\cmidrule(lr){7-8}
\textbf{Model} & \textbf{Train} & \textbf{EN} & \textbf{SV} & \textbf{EN} & \textbf{SV} & \textbf{EN} & \textbf{SV} \\
\midrule
\multicolumn{8}{l}{\textbf{LFM2.5-1.2B}} \\
Base, explanation prompt & -- & 17.38 & 21.31 & 39.12 & 23.69 & -21.74 & -2.38 \\
Trace SFT, direct & EN & 18.81 & 19.52 & 49.83 & 28.54 & -31.02 & -9.02 \\
Trace SFT, direct & SV & 17.50 & 19.17 & 46.89 & 45.83 & -29.39 & -26.67 \\
Trace SFT, direct & SV (rewritten) & 18.21 & 19.52 & 46.18 & 46.15 & -27.97 & -26.62 \\
Trace SFT, direct & EN+SV & 19.52 & 17.98 & 51.02 & 47.75 & -31.49 & -29.77 \\
Trace SFT, reasoning & EN & 18.10 & 8.10 & 51.97 & 15.12 & -33.88 & -7.03 \\
Trace SFT, reasoning & SV & 21.55 & 19.52 & 52.77 & 52.65 & -31.22 & -33.12 \\
Trace SFT, reasoning & SV (rewritten) & 16.43 & 17.74 & 46.60 & 46.83 & -30.17 & -29.10 \\
Trace SFT, reasoning & EN+SV & 18.69 & 21.90 & 53.51 & 54.81 & -34.82 & -32.90 \\
Base, answer-only prompt & -- & 15.48 & 18.21 & 38.25 & 24.38 & -22.78 & -6.16 \\
Answer SFT & EN & 20.00 & 17.98 & 36.68 & 25.47 & -16.68 & -7.49 \\
Answer SFT & SV & 18.69 & 19.29 & 28.22 & 28.61 & -9.53 & -9.32 \\
Answer SFT & EN+SV & 19.76 & 20.48 & 34.43 & 31.63 & -14.67 & -11.16 \\
\midrule
\multicolumn{8}{l}{\textbf{EuroLLM-1.7B}} \\
Base, explanation prompt & -- & 15.48 & 22.02 & 17.67 & 23.63 & -2.19 & -1.60 \\
Trace SFT, direct & EN & 20.36 & 19.05 & 47.04 & 40.84 & -26.68 & -21.79 \\
Trace SFT, direct & SV & 18.10 & 18.57 & 43.70 & 43.02 & -25.60 & -24.45 \\
Trace SFT, direct & SV (rewritten) & 18.81 & 20.71 & 42.01 & 41.14 & -23.20 & -20.43 \\
Trace SFT, direct & EN+SV & 19.52 & 20.48 & 47.07 & 42.91 & -27.55 & -22.43 \\
Trace SFT, reasoning & EN & 18.45 & 15.36 & 45.53 & 41.72 & -27.08 & -26.36 \\
Trace SFT, reasoning & SV & 18.93 & 18.10 & 46.72 & 47.03 & -27.79 & -28.93 \\
Trace SFT, reasoning & SV (rewritten) & 18.45 & 17.14 & 41.41 & 41.02 & -22.96 & -23.88 \\
Trace SFT, reasoning & EN+SV & 14.76 & 16.79 & 44.46 & 47.07 & -29.70 & -30.29 \\
Base, answer-only prompt & -- & 11.07 & 19.17 & 15.05 & 21.53 & -3.97 & -2.37 \\
Answer SFT & EN & 19.52 & 18.81 & 34.99 & 31.10 & -15.47 & -12.29 \\
Answer SFT & SV & 20.12 & 19.52 & 31.90 & 31.80 & -11.78 & -12.28 \\
Answer SFT & EN+SV & 17.86 & 19.05 & 32.94 & 30.42 & -15.08 & -11.37 \\
\bottomrule
\end{tabular}
}
    \caption{Accuracy (\%) on the 840 hard items and the 8,893 remaining core items.}
    \label{tab:hard_subset}
\end{table}

\begin{table}[htb]
    \centering
    \small
\begin{tabular}{llrrrr}
\toprule
& & \multicolumn{2}{c}{\textbf{Test-item weighted}} & \multicolumn{2}{c}{\textbf{Training-share weighted}} \\
\cmidrule(lr){3-4}\cmidrule(lr){5-6}
\textbf{Model} & \textbf{Train} & \textbf{EN} & \textbf{SV} & \textbf{EN} & \textbf{SV} \\
\midrule
\multicolumn{6}{l}{\textbf{LFM2.5-1.2B}} \\
Base, explanation prompt & -- & 37.24 & 23.49 & 40.48 & 31.81 \\
Trace SFT, direct & EN & 47.15 & 27.76 & 57.26 & 35.34 \\
Trace SFT, direct & SV & 44.35 & 43.53 & 52.78 & 53.01 \\
Trace SFT, direct & SV (rewritten) & 43.77 & 43.85 & 51.54 & 52.93 \\
Trace SFT, direct & EN+SV & 48.30 & 45.18 & 58.35 & 55.29 \\
Trace SFT, reasoning & EN & 49.05 & 14.52 & 58.55 & 21.29 \\
Trace SFT, reasoning & SV & 50.08 & 49.79 & 59.62 & 59.13 \\
Trace SFT, reasoning & SV (rewritten) & 43.99 & 44.32 & 52.19 & 53.03 \\
Trace SFT, reasoning & EN+SV & 50.51 & 51.97 & 60.40 & 62.84 \\
Base, answer-only prompt & -- & 36.29 & 23.85 & 38.89 & 28.88 \\
Answer SFT & EN & 35.24 & 24.82 & 46.33 & 32.02 \\
Answer SFT & SV & 27.40 & 27.80 & 34.95 & 36.92 \\
Answer SFT & EN+SV & 33.17 & 30.67 & 45.97 & 41.51 \\
\midrule
\multicolumn{6}{l}{\textbf{EuroLLM-1.7B}} \\
Base, explanation prompt & -- & 17.48 & 23.49 & 24.46 & 32.24 \\
Trace SFT, direct & EN & 44.73 & 38.96 & 54.44 & 47.12 \\
Trace SFT, direct & SV & 41.49 & 40.91 & 49.12 & 49.46 \\
Trace SFT, direct & SV (rewritten) & 40.01 & 39.38 & 48.40 & 48.22 \\
Trace SFT, direct & EN+SV & 44.69 & 40.97 & 54.97 & 49.92 \\
Trace SFT, reasoning & EN & 43.19 & 39.44 & 52.59 & 49.41 \\
Trace SFT, reasoning & SV & 44.32 & 44.53 & 54.23 & 54.44 \\
Trace SFT, reasoning & SV (rewritten) & 39.43 & 38.96 & 47.35 & 47.59 \\
Trace SFT, reasoning & EN+SV & 41.90 & 44.46 & 51.13 & 55.16 \\
Base, answer-only prompt & -- & 14.70 & 21.33 & 27.03 & 31.04 \\
Answer SFT & EN & 33.66 & 30.04 & 44.58 & 38.45 \\
Answer SFT & SV & 30.88 & 30.74 & 38.03 & 40.77 \\
Answer SFT & EN+SV & 31.63 & 29.44 & 40.53 & 38.72 \\
\bottomrule
\end{tabular}

    \caption{Accuracy (\%) weighted by the number of core items per dataset (as in the main results) and by each dataset's share of the training data (bilingual reasoning-trace data for trace models and base explanation prompt; answer-level data for answer models and base answer-only prompt).}
    \label{tab:train_weighted}
\end{table}

\begin{table}[htb]
    \centering
    \resizebox{\textwidth}{!}{%
\begin{tabular}{lrrrrrrrrrrr}
\toprule
& & \multicolumn{2}{c}{\textbf{Base}} & \multicolumn{2}{c}{\textbf{EN}} & \multicolumn{2}{c}{\textbf{SV}} & \multicolumn{2}{c}{\textbf{SV-RW}} & \multicolumn{2}{c}{\textbf{EN+SV}} \\
\cmidrule(lr){3-4}\cmidrule(lr){5-6}\cmidrule(lr){7-8}\cmidrule(lr){9-10}\cmidrule(lr){11-12}
\textbf{Dataset} & \textbf{Train \%} & EN & SV & EN & SV & EN & SV & EN & SV & EN & SV \\
\midrule
AfriMedQA & 0.083 & 41.57 & 19.88 & 63.25 & 19.28 & 60.24 & 62.05 & 55.42 & 55.42 & 62.05 & 62.05 \\
MedBullets & 7.203 & 35.69 & 21.55 & 49.12 & 9.54 & 55.83 & 53.00 & 44.88 & 48.76 & 55.12 & 60.78 \\
MedExQA & 0.015 & 58.41 & 21.06 & 64.50 & 19.98 & 66.02 & 63.30 & 59.17 & 58.31 & 63.08 & 64.71 \\
MedMCQA & 85.208 & 40.74 & 32.90 & 58.59 & 23.14 & 59.11 & 58.81 & 52.18 & 53.11 & 59.90 & 62.23 \\
MedQA & 7.187 & 41.22 & 28.41 & 67.09 & 10.94 & 69.13 & 68.70 & 59.03 & 55.81 & 71.16 & 71.84 \\
MedXpertQA-R & 0.002 & 11.17 & 9.43 & 14.23 & 3.48 & 15.02 & 15.02 & 12.73 & 13.15 & 15.02 & 15.92 \\
MedXpertQA-U & 0.001 & 11.58 & 9.19 & 11.95 & 3.86 & 12.32 & 13.60 & 11.76 & 13.24 & 15.62 & 16.73 \\
MMLU & 0.002 & 53.58 & 22.36 & 65.85 & 15.09 & 66.60 & 66.98 & 58.02 & 58.58 & 65.00 & 68.21 \\
MMLU-Pro & 0.003 & 27.97 & 11.49 & 34.48 & 7.92 & 36.02 & 37.55 & 28.74 & 32.06 & 36.53 & 34.74 \\
PubMedQA & 0.296 & 66.31 & 54.08 & 67.17 & 26.39 & 66.95 & 65.24 & 67.60 & 66.52 & 70.82 & 70.82 \\
\bottomrule
\end{tabular}
}
    \caption{LFM2.5-1.2B, reasoning-trace SFT with reasoning decoding: accuracy (\%) per dataset. Train \%: share of the bilingual reasoning-trace data.}
    \label{tab:per_dataset_lfm_reasoning}
\end{table}

\begin{table}[htb]
    \centering
    \resizebox{\textwidth}{!}{%
\begin{tabular}{lrrrrrrrrrrr}
\toprule
& & \multicolumn{2}{c}{\textbf{Base}} & \multicolumn{2}{c}{\textbf{EN}} & \multicolumn{2}{c}{\textbf{SV}} & \multicolumn{2}{c}{\textbf{SV-RW}} & \multicolumn{2}{c}{\textbf{EN+SV}} \\
\cmidrule(lr){3-4}\cmidrule(lr){5-6}\cmidrule(lr){7-8}\cmidrule(lr){9-10}\cmidrule(lr){11-12}
\textbf{Dataset} & \textbf{Train \%} & EN & SV & EN & SV & EN & SV & EN & SV & EN & SV \\
\midrule
AfriMedQA & 0.083 & 41.57 & 19.88 & 57.83 & 28.31 & 53.01 & 57.23 & 49.40 & 54.82 & 54.82 & 55.42 \\
MedBullets & 7.203 & 35.69 & 21.55 & 50.53 & 27.92 & 43.11 & 48.41 & 45.58 & 45.23 & 52.65 & 49.47 \\
MedExQA & 0.015 & 58.41 & 21.06 & 60.15 & 33.33 & 59.93 & 56.13 & 59.39 & 57.65 & 63.19 & 58.85 \\
MedMCQA & 85.208 & 40.74 & 32.90 & 57.31 & 36.05 & 53.08 & 52.93 & 51.46 & 53.11 & 58.21 & 55.40 \\
MedQA & 7.187 & 41.22 & 28.41 & 62.93 & 33.84 & 58.35 & 58.02 & 57.76 & 57.68 & 65.22 & 59.20 \\
MedXpertQA-R & 0.002 & 11.17 & 9.43 & 14.23 & 11.05 & 13.81 & 14.59 & 12.91 & 11.77 & 15.44 & 13.09 \\
MedXpertQA-U & 0.001 & 11.58 & 9.19 & 14.89 & 9.38 & 11.03 & 10.48 & 13.97 & 14.34 & 15.62 & 14.15 \\
MMLU & 0.002 & 53.58 & 22.36 & 61.70 & 30.28 & 57.08 & 54.53 & 57.36 & 57.08 & 61.60 & 58.58 \\
MMLU-Pro & 0.003 & 27.97 & 11.49 & 30.78 & 15.07 & 30.27 & 25.80 & 28.74 & 27.33 & 30.52 & 26.05 \\
PubMedQA & 0.296 & 66.31 & 54.08 & 67.17 & 50.43 & 68.67 & 67.17 & 69.74 & 70.60 & 69.53 & 70.39 \\
\bottomrule
\end{tabular}
}
    \caption{LFM2.5-1.2B, reasoning-trace SFT with direct decoding: accuracy (\%) per dataset.}
    \label{tab:per_dataset_lfm_direct}
\end{table}

\begin{table}[htb]
    \centering
    \resizebox{\textwidth}{!}{%
\begin{tabular}{lrrrrrrrrr}
\toprule
& & \multicolumn{2}{c}{\textbf{Base}} & \multicolumn{2}{c}{\textbf{EN}} & \multicolumn{2}{c}{\textbf{SV}} & \multicolumn{2}{c}{\textbf{EN+SV}} \\
\cmidrule(lr){3-4}\cmidrule(lr){5-6}\cmidrule(lr){7-8}\cmidrule(lr){9-10}
\textbf{Dataset} & \textbf{Train \%} & EN & SV & EN & SV & EN & SV & EN & SV \\
\midrule
AfriMedQA & 0.087 & 40.36 & 29.52 & 43.37 & 24.70 & 33.73 & 32.53 & 35.54 & 29.52 \\
MedBullets & 7.168 & 27.21 & 25.80 & 38.87 & 25.44 & 25.80 & 26.86 & 39.58 & 32.51 \\
MedExQA & 0.014 & 55.27 & 25.73 & 42.45 & 26.60 & 32.36 & 29.64 & 37.79 & 32.57 \\
MedMCQA & 85.232 & 39.65 & 29.07 & 47.49 & 32.78 & 36.01 & 38.11 & 47.19 & 42.72 \\
MedQA & 7.169 & 40.29 & 28.67 & 39.36 & 29.26 & 30.79 & 32.40 & 37.57 & 35.79 \\
MedXpertQA-R & 0.004 & 12.61 & 10.39 & 13.81 & 11.17 & 10.21 & 10.09 & 11.95 & 12.61 \\
MedXpertQA-U & 0.004 & 11.76 & 9.93 & 9.56 & 11.58 & 9.19 & 10.48 & 11.95 & 10.66 \\
MMLU & 0.003 & 51.70 & 25.94 & 39.53 & 28.40 & 33.30 & 32.17 & 36.13 & 33.96 \\
MMLU-Pro & 0.003 & 27.08 & 13.67 & 16.73 & 12.39 & 12.52 & 13.41 & 10.86 & 14.18 \\
PubMedQA & 0.317 & 67.17 & 51.50 & 63.30 & 41.20 & 52.79 & 50.21 & 59.23 & 52.36 \\
\bottomrule
\end{tabular}
}
    \caption{LFM2.5-1.2B, answer-level SFT: accuracy (\%) per dataset. Train \%: share of the answer-level data.}
    \label{tab:per_dataset_lfm_answer}
\end{table}

\begin{table}[htb]
    \centering
    \resizebox{\textwidth}{!}{%
\begin{tabular}{lrrrrrrrrrrr}
\toprule
& & \multicolumn{2}{c}{\textbf{Base}} & \multicolumn{2}{c}{\textbf{EN}} & \multicolumn{2}{c}{\textbf{SV}} & \multicolumn{2}{c}{\textbf{SV-RW}} & \multicolumn{2}{c}{\textbf{EN+SV}} \\
\cmidrule(lr){3-4}\cmidrule(lr){5-6}\cmidrule(lr){7-8}\cmidrule(lr){9-10}\cmidrule(lr){11-12}
\textbf{Dataset} & \textbf{Train \%} & EN & SV & EN & SV & EN & SV & EN & SV & EN & SV \\
\midrule
AfriMedQA & 0.083 & 15.66 & 17.47 & 57.23 & 49.40 & 53.61 & 63.25 & 44.58 & 48.19 & 48.80 & 55.42 \\
MedBullets & 7.203 & 18.37 & 18.37 & 43.46 & 41.70 & 44.52 & 50.53 & 34.28 & 38.52 & 41.34 & 43.82 \\
MedExQA & 0.015 & 18.24 & 22.04 & 59.07 & 52.99 & 58.96 & 58.09 & 54.94 & 51.90 & 57.33 & 58.85 \\
MedMCQA & 85.208 & 25.28 & 33.83 & 52.89 & 49.77 & 54.61 & 54.35 & 48.27 & 48.31 & 51.54 & 55.59 \\
MedQA & 7.187 & 20.70 & 26.80 & 57.68 & 52.59 & 58.95 & 58.86 & 48.69 & 47.33 & 55.39 & 60.90 \\
MedXpertQA-R & 0.002 & 5.89 & 8.65 & 11.17 & 9.25 & 12.85 & 14.17 & 12.79 & 11.89 & 11.53 & 10.21 \\
MedXpertQA-U & 0.001 & 5.33 & 10.85 & 10.48 & 10.29 & 11.21 & 12.32 & 9.19 & 11.76 & 11.76 & 11.40 \\
MMLU & 0.002 & 18.40 & 22.83 & 55.66 & 51.89 & 57.74 & 55.00 & 51.70 & 50.75 & 53.30 & 57.45 \\
MMLU-Pro & 0.003 & 8.68 & 13.67 & 27.46 & 22.86 & 26.69 & 28.10 & 22.61 & 21.71 & 23.63 & 27.46 \\
PubMedQA & 0.296 & 31.55 & 49.79 & 65.24 & 56.87 & 66.31 & 64.81 & 66.95 & 66.31 & 68.45 & 67.17 \\
\bottomrule
\end{tabular}
}
    \caption{EuroLLM-1.7B, reasoning-trace SFT with reasoning decoding: accuracy (\%) per dataset.}
    \label{tab:per_dataset_eurollm_reasoning}
\end{table}

\begin{table}[htb]
    \centering
    \resizebox{\textwidth}{!}{%
\begin{tabular}{lrrrrrrrrrrr}
\toprule
& & \multicolumn{2}{c}{\textbf{Base}} & \multicolumn{2}{c}{\textbf{EN}} & \multicolumn{2}{c}{\textbf{SV}} & \multicolumn{2}{c}{\textbf{SV-RW}} & \multicolumn{2}{c}{\textbf{EN+SV}} \\
\cmidrule(lr){3-4}\cmidrule(lr){5-6}\cmidrule(lr){7-8}\cmidrule(lr){9-10}\cmidrule(lr){11-12}
\textbf{Dataset} & \textbf{Train \%} & EN & SV & EN & SV & EN & SV & EN & SV & EN & SV \\
\midrule
AfriMedQA & 0.083 & 15.66 & 17.47 & 51.20 & 48.19 & 50.00 & 53.61 & 42.77 & 40.36 & 48.19 & 47.59 \\
MedBullets & 7.203 & 18.37 & 18.37 & 47.35 & 40.64 & 39.58 & 43.46 & 39.93 & 37.46 & 50.88 & 40.64 \\
MedExQA & 0.015 & 18.24 & 22.04 & 57.55 & 49.51 & 57.00 & 54.29 & 56.46 & 53.85 & 59.93 & 56.24 \\
MedMCQA & 85.208 & 25.28 & 33.83 & 54.54 & 47.41 & 49.51 & 49.70 & 48.91 & 48.87 & 54.99 & 50.53 \\
MedQA & 7.187 & 20.70 & 26.80 & 59.80 & 49.28 & 53.27 & 51.82 & 50.04 & 50.64 & 58.35 & 51.23 \\
MedXpertQA-R & 0.002 & 5.89 & 8.65 & 12.91 & 11.89 & 13.15 & 11.17 & 10.57 & 10.93 & 13.99 & 12.37 \\
MedXpertQA-U & 0.001 & 5.33 & 10.85 & 14.71 & 13.97 & 13.05 & 14.34 & 11.95 & 10.66 & 12.87 & 11.76 \\
MMLU & 0.002 & 18.40 & 22.83 & 55.66 & 48.58 & 53.02 & 54.43 & 49.91 & 49.72 & 54.43 & 49.25 \\
MMLU-Pro & 0.003 & 8.68 & 13.67 & 31.03 & 24.01 & 25.03 & 22.99 & 26.56 & 24.90 & 28.61 & 26.56 \\
PubMedQA & 0.296 & 31.55 & 49.79 & 68.24 & 68.45 & 69.10 & 67.17 & 68.24 & 64.81 & 67.81 & 69.74 \\
\bottomrule
\end{tabular}
}
    \caption{EuroLLM-1.7B, reasoning-trace SFT with direct decoding: accuracy (\%) per dataset.}
    \label{tab:per_dataset_eurollm_direct}
\end{table}

\begin{table}[htb]
    \centering
    \resizebox{\textwidth}{!}{%
\begin{tabular}{lrrrrrrrrr}
\toprule
& & \multicolumn{2}{c}{\textbf{Base}} & \multicolumn{2}{c}{\textbf{EN}} & \multicolumn{2}{c}{\textbf{SV}} & \multicolumn{2}{c}{\textbf{EN+SV}} \\
\cmidrule(lr){3-4}\cmidrule(lr){5-6}\cmidrule(lr){7-8}\cmidrule(lr){9-10}
\textbf{Dataset} & \textbf{Train \%} & EN & SV & EN & SV & EN & SV & EN & SV \\
\midrule
AfriMedQA & 0.087 & 14.46 & 21.08 & 32.53 & 31.33 & 31.33 & 29.52 & 33.73 & 32.53 \\
MedBullets & 7.168 & 4.59 & 18.37 & 33.92 & 31.10 & 31.80 & 37.46 & 28.62 & 27.21 \\
MedExQA & 0.014 & 21.06 & 20.52 & 41.04 & 37.02 & 41.37 & 40.50 & 41.26 & 39.20 \\
MedMCQA & 85.232 & 30.38 & 32.45 & 45.99 & 39.35 & 38.75 & 41.82 & 41.75 & 40.17 \\
MedQA & 7.169 & 10.94 & 26.72 & 37.83 & 34.18 & 35.11 & 31.21 & 37.40 & 32.91 \\
MedXpertQA-R & 0.004 & 0.00 & 5.17 & 11.47 & 10.51 & 10.87 & 11.05 & 11.47 & 11.35 \\
MedXpertQA-U & 0.004 & 0.18 & 5.33 & 11.40 & 10.11 & 9.74 & 10.29 & 11.03 & 9.56 \\
MMLU & 0.003 & 18.58 & 21.89 & 38.02 & 33.02 & 40.66 & 35.57 & 35.85 & 34.06 \\
MMLU-Pro & 0.003 & 7.79 & 11.11 & 15.96 & 15.20 & 15.33 & 15.20 & 14.81 & 14.30 \\
PubMedQA & 0.317 & 0.43 & 39.91 & 63.30 & 62.66 & 53.86 & 52.58 & 56.01 & 42.92 \\
\bottomrule
\end{tabular}
}
    \caption{EuroLLM-1.7B, answer-level SFT: accuracy (\%) per dataset.}
    \label{tab:per_dataset_eurollm_answer}
\end{table}

\section{Native Swedish Exam Sets}
\label{sec:appendix-native}

\paragraph{Data.} Both sets come from the repository of the Swedish Medical LLM Benchmark\footnote{\url{https://github.com/BirgerMoell/swedish-medical-benchmark}, commit \texttt{f6297d5}.}, which also holds a copy of MedQA-SWE. Their questions come from the theoretical part of the Swedish knowledge test (\textit{kunskapsprov}) that physicians trained outside the EU and EEA take for a Swedish license, and were written in Swedish. We parse the options into A--E and drop 8 of the 3,180 MedQA-SWE questions whose options are malformed: in 7, two options are merged on one line, and in 1, an option is empty. Of the 535 SMDT questions, 28 are malformed: they have no options or not exactly the options a--e, or an answer that cannot be parsed or whose text differs from the option it names. The two sets overlap, so we drop the 348 SMDT questions that repeat a MedQA-SWE question, 221 with the same normalized question text and 127 with a question similarity of at least 0.9; in both cases the options must also have a similarity of at least 0.5. We keep 7 SMDT questions that share a case with a MedQA-SWE question but ask something else. The 159 remaining SMDT questions come from exams held between 2016 and 2024. Every question has five options and one correct answer. The prompts are built by the same code as the Swedish benchmark prompts, so their template, option format, and answer schema are the same. In both sets, a question that follows up on a case can name the answer to the previous question; we keep the questions as released and did not count such questions or measure their effect, which also makes the intervals somewhat optimistic.

\paragraph{Overlap and contamination.} Our training questions are translations of the English datasets of Section~\ref{sec:datasets} and are not drawn from either set; we did not search them for questions that resemble a native exam question. The exams are public, so the base models and the teachers may have seen them during pretraining, and teacher exposure could reach the students through the traces; we cannot measure the effect of such exposure on the gains.

\paragraph{Results.} Table~\ref{tab:native_swedish_full} gives every student configuration on both sets, and Table~\ref{tab:native_swedish_tests} the paired comparisons. LFM2.5-1.2B trained on English traces, which leaves most of its reasoning outputs unfinished, gives an invalid output on 56.8\% of MedQA-SWE and 66.7\% of SMDT. On SMDT, none of the three tested differences below 10 points is significant there.

\begin{table}[htb]
    \centering
    \small
    \setlength{\tabcolsep}{4pt}
    \resizebox{\textwidth}{!}{%
\begin{tabular}{lllrrrrr}
\toprule
& & & & \multicolumn{2}{c}{\textbf{MedQA-SWE}} & \multicolumn{2}{c}{\textbf{SMDT}} \\
\cmidrule(lr){5-6}\cmidrule(lr){7-8}
\textbf{Student} & \textbf{Trained on} & \textbf{Decoding} & \textbf{Core SV} & \textbf{Acc. [95\% CI]} & \textbf{Inv.} & \textbf{Acc. [95\% CI]} & \textbf{Inv.} \\
\midrule
LFM2.5-1.2B & None, answer-only prompt & -- & 23.8 & 24.5 [23.0, 26.0] & 0.1 & 23.9 [17.6, 30.8] & 0.0 \\
 & EN answers & -- & 24.8 & 23.7 [22.3, 25.2] & 0.0 & 25.8 [18.9, 32.7] & 0.0 \\
 & SV answers & -- & 27.8 & 22.9 [21.4, 24.3] & 0.0 & 24.5 [18.2, 31.4] & 0.0 \\
 & EN+SV answers & -- & 30.7 & 25.5 [24.0, 27.0] & 0.0 & 17.0 [11.3, 23.3] & 0.0 \\
 & None, explanation prompt & -- & 23.5 & 21.3 [19.9, 22.7] & 0.6 & 22.6 [16.4, 28.9] & 0.6 \\
 & EN traces & Reasoning & 14.5 & 16.0 [14.8, 17.3] & 56.8 & 8.2 [4.4, 12.6] & 66.7 \\
 & EN traces & Direct & 27.8 & 29.7 [28.2, 31.3] & 0.6 & 25.8 [18.9, 32.7] & 1.9 \\
 & SV traces & Reasoning & 49.8 & 61.5 [59.8, 63.3] & 1.2 & 51.6 [44.0, 59.1] & 1.9 \\
 & SV traces & Direct & 43.5 & 53.9 [52.1, 55.6] & 0.0 & 45.3 [37.7, 52.8] & 0.0 \\
 & SV (rewritten) traces & Reasoning & 44.3 & 54.3 [52.6, 56.1] & 0.3 & 49.7 [42.1, 57.9] & 0.0 \\
 & SV (rewritten) traces & Direct & 43.9 & 53.9 [52.1, 55.6] & 0.0 & 47.8 [40.3, 55.3] & 0.0 \\
 & EN+SV traces & Reasoning & 52.0 & 64.8 [63.1, 66.5] & 1.3 & 54.1 [46.5, 61.6] & 1.9 \\
 & EN+SV traces & Direct & 45.2 & 56.6 [54.9, 58.3] & 0.0 & 49.1 [41.5, 56.6] & 0.0 \\
\midrule
EuroLLM-1.7B & None, answer-only prompt & -- & 21.3 & 19.8 [18.4, 21.2] & 6.7 & 19.5 [13.8, 25.8] & 5.7 \\
 & EN answers & -- & 30.0 & 28.3 [26.7, 29.9] & 0.0 & 28.9 [22.0, 35.8] & 0.0 \\
 & SV answers & -- & 30.7 & 28.1 [26.5, 29.6] & 0.0 & 24.5 [18.2, 31.4] & 0.0 \\
 & EN+SV answers & -- & 29.4 & 26.5 [25.0, 28.1] & 0.0 & 18.2 [12.6, 24.5] & 0.0 \\
 & None, explanation prompt & -- & 23.5 & 21.0 [19.6, 22.4] & 3.8 & 17.0 [11.3, 22.6] & 6.9 \\
 & EN traces & Reasoning & 39.4 & 48.3 [46.6, 50.1] & 8.5 & 44.0 [36.5, 51.6] & 8.8 \\
 & EN traces & Direct & 39.0 & 47.3 [45.6, 49.0] & 0.1 & 42.1 [34.6, 49.7] & 0.0 \\
 & SV traces & Reasoning & 44.5 & 56.7 [54.9, 58.4] & 1.5 & 50.9 [43.4, 59.1] & 2.5 \\
 & SV traces & Direct & 40.9 & 49.6 [47.9, 51.3] & 0.0 & 50.9 [43.4, 58.5] & 0.0 \\
 & SV (rewritten) traces & Reasoning & 39.0 & 48.4 [46.6, 50.1] & 0.2 & 37.1 [29.6, 44.7] & 0.0 \\
 & SV (rewritten) traces & Direct & 39.4 & 48.8 [47.1, 50.5] & 0.1 & 46.5 [39.0, 54.1] & 0.0 \\
 & EN+SV traces & Reasoning & 44.5 & 55.1 [53.4, 56.9] & 4.9 & 45.3 [37.7, 52.8] & 4.4 \\
 & EN+SV traces & Direct & 41.0 & 50.7 [49.0, 52.5] & 0.1 & 40.3 [32.7, 47.8] & 0.0 \\
\bottomrule
\end{tabular}
}
    \caption{Accuracy (\%) of every student configuration on the core Swedish items of the benchmark (Core SV) and on the two native Swedish sets, with 95\% bootstrap intervals over questions (10,000 resamples) and the share of invalid outputs (Inv., \%), which count as wrong. Answer-level models and the base model with the answer-only prompt answer directly; reasoning-trace models are evaluated with reasoning and with direct decoding. Chance is 20\%.}
    \label{tab:native_swedish_full}
\end{table}

\begin{table}[htb]
    \centering
    \small
    \setlength{\tabcolsep}{4pt}
    \resizebox{\textwidth}{!}{%
\begin{tabular}{lllcrrrrr}
\toprule
\textbf{Student} & \textbf{A} & \textbf{B} & \textbf{Set} & \textbf{A} & \textbf{B} & \textbf{A $-$ B [95\% CI]} & \textbf{McNemar $p$} & \textbf{Holm $p$} \\
\midrule
LFM2.5-1.2B & EN+SV traces & base model & MedQA-SWE & 64.8 & 21.3 & +43.5 [+41.4, +45.6] & $<$0.001 & $<$0.001 \\
 & EN+SV traces & base model & SMDT & 54.1 & 22.6 & +31.4 [+22.0, +40.9] & $<$0.001 & $<$0.001 \\
 & SV traces & base model & MedQA-SWE & 61.5 & 21.3 & +40.3 [+38.1, +42.4] & $<$0.001 & $<$0.001 \\
 & SV traces & base model & SMDT & 51.6 & 22.6 & +28.9 [+18.9, +39.0] & $<$0.001 & $<$0.001 \\
 & EN+SV traces & SV traces & MedQA-SWE & 64.8 & 61.5 & +3.2 [+1.6, +4.9] & $<$0.001 & $<$0.001 \\
 & EN+SV traces & SV traces & SMDT & 54.1 & 51.6 & +2.5 [-5.7, +10.7] & 0.659 & 0.659 \\
 & SV traces & EN traces & MedQA-SWE & 61.5 & 16.0 & +45.5 [+43.5, +47.4] & $<$0.001 & $<$0.001 \\
 & SV traces & EN traces & SMDT & 51.6 & 8.2 & +43.4 [+34.6, +52.2] & $<$0.001 & $<$0.001 \\
 & EN+SV traces & EN+SV answer SFT & MedQA-SWE & 64.8 & 25.5 & +39.2 [+37.1, +41.4] & $<$0.001 & $<$0.001 \\
 & EN+SV traces & EN+SV answer SFT & SMDT & 54.1 & 17.0 & +37.1 [+27.0, +47.2] & $<$0.001 & $<$0.001 \\
\midrule
EuroLLM-1.7B & EN+SV traces & base model & MedQA-SWE & 55.1 & 21.0 & +34.1 [+32.0, +36.3] & $<$0.001 & $<$0.001 \\
 & EN+SV traces & base model & SMDT & 45.3 & 17.0 & +28.3 [+19.5, +37.1] & $<$0.001 & $<$0.001 \\
 & SV traces & base model & MedQA-SWE & 56.7 & 21.0 & +35.7 [+33.5, +37.8] & $<$0.001 & $<$0.001 \\
 & SV traces & base model & SMDT & 50.9 & 17.0 & +34.0 [+25.2, +43.4] & $<$0.001 & $<$0.001 \\
 & EN+SV traces & SV traces & MedQA-SWE & 55.1 & 56.7 & -1.5 [-3.3, +0.2] & 0.093 & 0.374 \\
 & EN+SV traces & SV traces & SMDT & 45.3 & 50.9 & -5.7 [-14.5, +3.1] & 0.272 & 0.555 \\
 & SV traces & EN traces & MedQA-SWE & 56.7 & 48.3 & +8.4 [+6.5, +10.2] & $<$0.001 & $<$0.001 \\
 & SV traces & EN traces & SMDT & 50.9 & 44.0 & +6.9 [-1.9, +16.4] & 0.185 & 0.555 \\
 & EN+SV traces & EN+SV answer SFT & MedQA-SWE & 55.1 & 26.5 & +28.6 [+26.5, +30.8] & $<$0.001 & $<$0.001 \\
 & EN+SV traces & EN+SV answer SFT & SMDT & 45.3 & 18.2 & +27.0 [+18.2, +35.8] & $<$0.001 & $<$0.001 \\
\bottomrule
\end{tabular}
}
    \caption{Paired comparisons of student configurations on the native Swedish sets: accuracy of A and B (\%), the difference with its paired 95\% bootstrap interval, the exact McNemar test, and the Holm-adjusted $p$-value over all 20 rows. Trace models use reasoning decoding; the base model uses the explanation prompt.}
    \label{tab:native_swedish_tests}
\end{table}

\section{Reasoning Traces: Examples}
\label{sec:appendix-examples}

Even a student that is equally accurate in both languages does not solve the same items in both. Table~\ref{tab:item_counts} gives the paired outcomes of every reasoning-trace student. The boxes below show the bilingual LFM2.5-1.2B model with reasoning decoding on seven test items: five that it answers wrongly in English but correctly in Swedish, and two of the opposite kind. Each box shows the English and the Swedish item and the model's reasoning and answer for each; long traces are shortened in the middle (``[\ldots]''). The model reasons in English for both inputs, as Section~\ref{sec:analysis-language} discusses; only its final explanation, not shown here, is in Swedish. We show the examples as they are and do not interpret them.

\begin{table}[htb]
    \centering
    \small
\begin{tabular}{lllrrrr}
\toprule
\textbf{Student} & \textbf{Train} & \textbf{Decoding} & \textbf{Both} & \textbf{EN only} & \textbf{SV only} & \textbf{Neither} \\
\midrule
LFM2.5-1.2B & EN & Reasoning & 1,054 & 3,720 & 359 & 4,600 \\
LFM2.5-1.2B & SV & Reasoning & 3,850 & 1,024 & 996 & 3,863 \\
LFM2.5-1.2B & SV (rewritten) & Reasoning & 3,169 & 1,113 & 1,145 & 4,306 \\
LFM2.5-1.2B & EN+SV & Reasoning & 4,040 & 876 & 1,018 & 3,799 \\
LFM2.5-1.2B & EN & Direct & 1,808 & 2,781 & 894 & 4,250 \\
LFM2.5-1.2B & SV & Direct & 3,126 & 1,191 & 1,111 & 4,305 \\
LFM2.5-1.2B & SV (rewritten) & Direct & 3,187 & 1,073 & 1,081 & 4,392 \\
LFM2.5-1.2B & EN+SV & Direct & 3,428 & 1,273 & 969 & 4,063 \\
\midrule
EuroLLM-1.7B & EN & Reasoning & 2,805 & 1,399 & 1,034 & 4,495 \\
EuroLLM-1.7B & SV & Reasoning & 3,243 & 1,071 & 1,091 & 4,328 \\
EuroLLM-1.7B & SV (rewritten) & Reasoning & 2,502 & 1,336 & 1,290 & 4,605 \\
EuroLLM-1.7B & EN+SV & Reasoning & 3,155 & 923 & 1,172 & 4,483 \\
EuroLLM-1.7B & EN & Direct & 2,798 & 1,556 & 994 & 4,385 \\
EuroLLM-1.7B & SV & Direct & 2,795 & 1,243 & 1,187 & 4,508 \\
EuroLLM-1.7B & SV (rewritten) & Direct & 2,693 & 1,201 & 1,140 & 4,699 \\
EuroLLM-1.7B & EN+SV & Direct & 2,937 & 1,413 & 1,051 & 4,332 \\
\bottomrule
\end{tabular}

    \caption{Paired outcomes of the reasoning-trace students on the 9,733 core test items. Both: correct in English and in Swedish. EN only: correct only in English. SV only: correct only in Swedish. Neither: wrong in both.}
    \label{tab:item_counts}
\end{table}

\begin{figure}[htb]
\begin{tcolorbox}[enhanced, colback=white, colframe=black!55, colbacktitle=black!8, coltitle=black, fonttitle=\footnotesize\bfseries, boxrule=0.5pt, arc=1pt, left=4pt, right=4pt, top=3pt, bottom=3pt, before skip=6pt, after skip=8pt,
  title={Example 1: MedQA \quad\textmd{(English $\times$\,/ Swedish $\checkmark$)}\hfill Gold answer: A}]
\footnotesize
\textit{Statin pharmacology; which substance decreases after starting atorvastatin.}\par\smallskip
\begin{minipage}[t]{0.485\linewidth}\raggedright
\textbf{English item}\par\smallskip
A 58-year-old male with a history of obesity and hypertension presents to his primary care physician for a follow-up visit. He reports that he feels well and has no complaints. He currently takes hydrochlorothiazide. His temperature is 98.6\textdegree{}F (37\textdegree{}C), blood pressure is 135/80 mmHg, pulse is 86/min, and respirations are 17/min. His BMI is 31 kg/m2. Results of a lipid panel are: Total cholesterol is 280 mg/dl, triglycerides are 110 mg/dl, HDL cholesterol is 40 mg/dl, and LDL cholesterol is 195 mg/dl. Her physician considers starting her on atorvastatin. Which of the following will most likely decrease after initiating this medication?\par\smallskip
\textbf{A.}~Mevalonate\par \textbf{B.}~Acetoacetyl-CoA\par \textbf{C.}~Diacyglycerol\par \textbf{D.}~High-density lipoprotein
\end{minipage}\hfill
\begin{minipage}[t]{0.485\linewidth}\raggedright
\textbf{Swedish item}\par\smallskip
En 58-årig man med anamnes på obesitas och hypertoni kommer till sin allmänläkare för ett uppföljningsbesök. Han uppger att han mår bra och inte har några besvär. Han behandlas för närvarande med hydroklortiazid. Hans temperatur är 98,6\textdegree{}F (37\textdegree{}C), blodtrycket 135/80 mmHg, pulsen 86/min och andningsfrekvensen 17/min. Hans BMI är 31 kg/m2. Resultat från en lipidprofil är: totalkolesterol 280 mg/dl, triglycerider 110 mg/dl, HDL-kolesterol 40 mg/dl och LDL-kolesterol 195 mg/dl. Hans läkare överväger att sätta in atorvastatin. Vilket av följande kommer mest sannolikt att minska efter insättning av detta läkemedel?\par\smallskip
\textbf{A.}~mevalonat\par \textbf{B.}~acetoacetyl-CoA\par \textbf{C.}~diacylglycerol\par \textbf{D.}~high-density-lipoprotein (HDL)
\end{minipage}
\tcbline
\begin{minipage}[t]{0.485\linewidth}\raggedright\scriptsize
\textbf{English input $\rightarrow$ model output}\par\smallskip
\textit{Reasoning:} We need to parse the question: 58-year-old male, obesity, hypertension, lipid panel: total cholesterol 280, TG 110, HDL 40, LDL 195. The physician considers starting atorvastatin (a statin). Which of the following will most likely decrease after initiating this medication? Options: A) Mevalonate, B) Acetoacetyl-CoA, C) Diacylglycerol, D) High-density lipoprotein. We need to know statin mechanism: HMG-CoA reductase inhibitor reduces cholesterol synthesis, leading to upregulation of LDL receptors, increased clearance of LDL from plasma. So LDL levels will decrease. The question asks which will most likely decrease after initiating statin. Among options, LDL (high-density lipoprotein) would decrease. [\ldots] Indeed, statins lower LDL cholesterol. So D. Provide explanation: Statins inhibit HMG-CoA reductase, decreasing hepatic cholesterol synthesis, upregulating LDL receptors, increasing clearance of LDL, thus plasma LDL falls. The other molecules are not directly reduced. Thus answer index D. Return JSON.\par\smallskip
\textit{Final answer:} \textbf{D.}~High-density lipoprotein~\textcolor{red!70!black}{$\times$}
\end{minipage}\hfill
\begin{minipage}[t]{0.485\linewidth}\raggedright\scriptsize
\textbf{Swedish input $\rightarrow$ model output}\par\smallskip
\textit{Reasoning:} We need to parse the question: 58-year-old man with obesity, hypertension, on hydrochlorothiazide, lipid profile: total cholesterol 280, TG 110, HDL 40, LDL 195. Doctor considers starting atorvastatin. Which of the following will most likely decrease after starting this drug? Options: mevalonate, acetoacetyl-CoA, diacylglycerol, HDL. We need to know mechanism of statins: they inhibit HMG-CoA reductase, decreasing cholesterol synthesis, leading to upregulation of LDL receptors, increased clearance of LDL from plasma, also modestly increase HDL. They also increase levels of mevalonate? Actually HMG-CoA reductase converts HMG-CoA to mevalonate. Inhibition reduces mevalonate production. So mevalonate would decrease. [\ldots] So answer A. But need to ensure nuance: Statins also increase levels of HMG-CoA reductase activity? No, they inhibit. So mevalonate production decreases. So answer A. Provide explanation. Return JSON with explanation and answer\_index "A".\par\smallskip
\textit{Final answer:} \textbf{A.}~mevalonat~\textcolor{green!45!black}{$\checkmark$}
\end{minipage}
\end{tcolorbox}
\end{figure}

\begin{figure}[htb]
\begin{tcolorbox}[enhanced, colback=white, colframe=black!55, colbacktitle=black!8, coltitle=black, fonttitle=\footnotesize\bfseries, boxrule=0.5pt, arc=1pt, left=4pt, right=4pt, top=3pt, bottom=3pt, before skip=6pt, after skip=8pt,
  title={Example 2: PubMedQA \quad\textmd{(English $\times$\,/ Swedish $\checkmark$)}\hfill Gold answer: B}]
\footnotesize
\textit{Abstract-based yes/no/maybe question on serovar-specific immunity to Neisseria gonorrhoeae.}\par\smallskip
\begin{minipage}[t]{0.485\linewidth}\raggedright
\textbf{English item}\par\smallskip
To determine whether the host immune response to gonorrhoea provides limited serovar specific protection from reinfection. 508 episodes of gonorrhoea diagnosed at a city centre genitourinary medicine clinic including 22 patients with multiple infections over a 4 year period. Patients with recurrent gonococcal infection were analysed with respect to the initial and subsequent serovars isolated. No significant difference was seen in the prevalence of serovars isolated following a repeat infection compared with those without repeat infections. The site of the initial infection did not appear to influence the subsequent serovar isolated. Serovar specific immunity to Neisseria gonorrhoeae: does it exist?\par\smallskip
\textbf{A.}~yes\par \textbf{B.}~no\par \textbf{C.}~maybe
\end{minipage}\hfill
\begin{minipage}[t]{0.485\linewidth}\raggedright
\textbf{Swedish item}\par\smallskip
För att avgöra om värdens immunsvar mot gonorré ger ett begränsat serovarspecifikt skydd mot återinfektion. 508 fall av gonorré diagnostiserades vid en genitourinärmedicinsk klinik i stadskärnan, inklusive 22 patienter med multipla infektioner under en 4-årsperiod. Patienter med recidiverande gonokockinfektion analyserades med avseende på de serovarianter som isolerades initialt och därefter. Ingen signifikant skillnad sågs i prevalensen av serovarianter som isolerades efter en upprepad infektion jämfört med hos dem utan upprepade infektioner. Lokalen för den initiala infektionen verkade inte påverka vilken serovariant som därefter isolerades. Serovarspecifik immunitet mot Neisseria gonorrhoeae: finns den?\par\smallskip
\textbf{A.}~ja\par \textbf{B.}~nej\par \textbf{C.}~kanske
\end{minipage}
\tcbline
\begin{minipage}[t]{0.485\linewidth}\raggedright\scriptsize
\textbf{English input $\rightarrow$ model output}\par\smallskip
\textit{Reasoning:} We need to interpret the question. It describes a study: 508 episodes of gonorrhoea diagnosed at a city centre genitourinary medicine clinic including 22 patients with multiple infections over a 4 year period. Patients with recurrent gonococcal infection were analyzed with respect to the initial and subsequent serovars isolated. No significant difference was seen in the prevalence of serovars isolated following a repeat infection compared with those without repeat infections. The site of the initial infection did not appear to influence the subsequent serovar isolated. Serovar specific immunity to Neisseria gonorrhoeae: does it exist? [\ldots] That suggests that immunity is not serovar-specific. So the answer: yes, serovar specific immunity exists. So answer A. We need to produce JSON with explanation and answer\_index.\par\smallskip
\textit{Final answer:} \textbf{A.}~yes~\textcolor{red!70!black}{$\times$}
\end{minipage}\hfill
\begin{minipage}[t]{0.485\linewidth}\raggedright\scriptsize
\textbf{Swedish input $\rightarrow$ model output}\par\smallskip
\textit{Reasoning:} We need to parse the question: It's Swedish. It describes a study: 508 cases of gonorrhea diagnosed at a genitourinary clinic in city, including 22 patients with multiple infections over 4-year period. Patients with recurrent gonococcal infection analyzed with respect to serovars isolated initially and then. No significant difference seen in prevalence of serovars isolated after repeated infection compared to those without repeated infections. The site of the initial infection did not affect which serovar was then isolated. Serovarspecific immunity to N. gonorrhoeae: does it exist? Options: A) ja (yes), B) nej (no), C) kanske (maybe). [\ldots] So answer B. Provide explanation: The study shows no significant difference in serovar prevalence after repeated infections, indicating that immunity is not serovar-specific; thus immunity does not exist in a limited manner. Return JSON with explanation and answer\_index "B".\par\smallskip
\textit{Final answer:} \textbf{B.}~nej~\textcolor{green!45!black}{$\checkmark$}
\end{minipage}
\end{tcolorbox}
\end{figure}

\begin{figure}[htb]
\begin{tcolorbox}[enhanced, colback=white, colframe=black!55, colbacktitle=black!8, coltitle=black, fonttitle=\footnotesize\bfseries, boxrule=0.5pt, arc=1pt, left=4pt, right=4pt, top=3pt, bottom=3pt, before skip=6pt, after skip=8pt,
  title={Example 3: MedExQA \quad\textmd{(English $\times$\,/ Swedish $\checkmark$)}\hfill Gold answer: D}]
\footnotesize
\textit{Immunoglobulin structure; location of the antigen-binding site.}\par\smallskip
\begin{minipage}[t]{0.485\linewidth}\raggedright
\textbf{English item}\par\smallskip
In the structure of immunoglobulins, where is the antigen-binding site?\par\smallskip
\textbf{A.}~Fc\par \textbf{B.}~Carbohydrate\par \textbf{C.}~Hinge\par \textbf{D.}~Fab
\end{minipage}\hfill
\begin{minipage}[t]{0.485\linewidth}\raggedright
\textbf{Swedish item}\par\smallskip
Var finns antigenbindningsstället i immunoglobulinernas struktur?\par\smallskip
\textbf{A.}~Fc\par \textbf{B.}~Kolhydrat\par \textbf{C.}~hinge-regionen\par \textbf{D.}~Fab
\end{minipage}
\tcbline
\begin{minipage}[t]{0.485\linewidth}\raggedright\scriptsize
\textbf{English input $\rightarrow$ model output}\par\smallskip
\textit{Reasoning:} We need to answer: In the structure of immunoglobulins, where is the antigen-binding site? Options: A) Fc, B) Carbohydrate, C) Hinge, D) Fab. Immunoglobulin structure: heavy chain constant region (CH) forms the Fc region, which is the antigen-binding site. Light chain variable region (VH, VL) forms the Fab region, which includes antigen-binding site. The hinge region is between Fab and Fc, not antigen-binding. Carbohydrate? Not relevant. So answer: Fc. Option A. Explain: The antigen-binding site is formed by the variable domains of the heavy and light chains, located in the Fab region. The Fc region is the constant region, not antigen-binding. The hinge is a flexible region connecting Fab and Fc. Carbohydrate is not part of immunoglobulin structure. So answer A. Return JSON.\par\smallskip
\textit{Final answer:} \textbf{A.}~Fc~\textcolor{red!70!black}{$\times$}
\end{minipage}\hfill
\begin{minipage}[t]{0.485\linewidth}\raggedright\scriptsize
\textbf{Swedish input $\rightarrow$ model output}\par\smallskip
\textit{Reasoning:} We need to answer: "Var finns antigenbindningsstället i immunoglobulinernas struktur?" Swedish: "Where is the antigen-binding site in the immunoglobulin structure?" Options: A) Fc, B) Carbohydrate, C) hinge region, D) Fab. Antigen-binding site is located in the variable region of the heavy and light chains, specifically the Fab region (Fab = antigen-binding fragment). So answer D) Fab. Explain: antigen-binding site is formed by variable domains of heavy and light chains (VH, VL). Fc is constant region, not binding. Carbohydrate not relevant. Hinge region is flexible region between Fab and Fc, not binding. Thus JSON with explanation and answer\_index "D".\par\smallskip
\textit{Final answer:} \textbf{D.}~Fab~\textcolor{green!45!black}{$\checkmark$}
\end{minipage}
\end{tcolorbox}
\end{figure}

\begin{figure}[htb]
\begin{tcolorbox}[enhanced, colback=white, colframe=black!55, colbacktitle=black!8, coltitle=black, fonttitle=\footnotesize\bfseries, boxrule=0.5pt, arc=1pt, left=4pt, right=4pt, top=3pt, bottom=3pt, before skip=6pt, after skip=8pt,
  title={Example 4: MMLU \quad\textmd{(English $\times$\,/ Swedish $\checkmark$)}\hfill Gold answer: D}]
\footnotesize
\textit{Next management step for penile lesions that turn white with acetic acid.}\par\smallskip
\begin{minipage}[t]{0.485\linewidth}\raggedright
\textbf{English item}\par\smallskip
A previously healthy 27-year-old man comes to the physician 4 weeks after noticing three nontender lesions on his penis. He says they have not changed in size. He is sexually active with multiple male and female partners and uses condoms inconsistently. He takes no medications. He drinks two to five beers on social occasions. He occasionally smokes marijuana. His temperature is 36.9\textdegree{}C (98.4\textdegree{}F). There is no lymphadenopathy. Examination shows three sessile, flesh-colored lesions on the shaft of the penis that are 10 mm in diameter. On application of a dilute solution of acetic acid, the lesions turn white. The remainder of the examination shows no abnormalities. Which of the following is the most appropriate next step in management?\par\smallskip
\textbf{A.}~Oral acyclovir therapy\par \textbf{B.}~Oral doxycycline therapy\par \textbf{C.}~Intramuscular penicillin therapy\par \textbf{D.}~Cryotherapy
\end{minipage}\hfill
\begin{minipage}[t]{0.485\linewidth}\raggedright
\textbf{Swedish item}\par\smallskip
En tidigare frisk 27-årig man kommer till läkaren 4 veckor efter att ha upptäckt tre oömma lesioner på penis. Han säger att de inte har förändrats i storlek. Han är sexuellt aktiv med flera manliga och kvinnliga partner och använder kondom inkonsekvent. Han tar inga läkemedel. Han dricker två till fem öl vid sociala tillfällen. Han röker ibland marijuana. Hans temperatur är 36,9\textdegree{}C (98,4\textdegree{}F). Ingen lymfadenopati föreligger. Vid undersökning ses tre bredbasiga, hudfärgade lesioner på penisskaftet, vardera 10 mm i diameter. Vid applicering av en utspädd lösning av ättiksyra blir lesionerna vita. I övrigt visar undersökningen inga avvikelser. Vilket av följande är det mest lämpliga nästa steget i handläggningen?\par\smallskip
\textbf{A.}~Oral behandling med aciklovir\par \textbf{B.}~Oral behandling med doxycyklin\par \textbf{C.}~Intramuskulär behandling med penicillin\par \textbf{D.}~Kryoterapi
\end{minipage}
\tcbline
\begin{minipage}[t]{0.485\linewidth}\raggedright\scriptsize
\textbf{English input $\rightarrow$ model output}\par\smallskip
\textit{Reasoning:} We need to analyze the case. 27-year-old man, 4 weeks after noticing three nontender lesions on penis. Lesions are sessile, flesh-colored, 10 mm diameter. On application of dilute acetic acid, lesions turn white. That is a classic test for syphilis: the acetic acid test (also called the "acetic acid test" or "acetic acid test for syphilis" or "acetic acid test for syphilis" is used to detect syphilis lesions). The lesions are likely syphilitic chancres. The patient is previously healthy, but has multiple partners, inconsistent condom use. So likely primary syphilis. The next step: treat with penicillin. [\ldots] The appropriate treatment is benzathine penicillin G IM. Acyclovir is for herpes; doxycycline is for chlamydia; cryotherapy is for condyloma. So answer C. Return JSON.\par\smallskip
\textit{Final answer:} \textbf{C.}~Intramuscular penicillin therapy~\textcolor{red!70!black}{$\times$}
\end{minipage}\hfill
\begin{minipage}[t]{0.485\linewidth}\raggedright\scriptsize
\textbf{Swedish input $\rightarrow$ model output}\par\smallskip
\textit{Reasoning:} We need to analyze the case. 27-year-old man with three painless lesions on penis, 10 mm each, broad-based, skin-colored, appear after 4 weeks, no lymphadenopathy. Application of acetic acid (Vinegar) makes lesions turn white (acetowhite). This suggests condyloma acuminata (genital warts) caused by HPV (low-risk types 6,11). The lesions are painless, broad-based, cauliflower-like. Acetowhite reaction is typical for HPV lesions. Management: first-line for external genital warts: patient-applied topical therapy (imiquimod, podofilox, sinecatechins) or provider-administered destructive methods (cryotherapy, surgical excision, laser). Oral antivirals (acyclovir) are for HSV, not HPV. Doxycycline is for chlamydia, not appropriate. Penicillin is for syphilis (chancre). So best next step: Cryotherapy (option D). So answer D. Explain: The lesions are condyloma acuminata; treat with cryotherapy. Provide reasoning. Return JSON with explanation and answer\_index "D".\par\smallskip
\textit{Final answer:} \textbf{D.}~Kryoterapi~\textcolor{green!45!black}{$\checkmark$}
\end{minipage}
\end{tcolorbox}
\end{figure}

\begin{figure}[htb]
\begin{tcolorbox}[enhanced, colback=white, colframe=black!55, colbacktitle=black!8, coltitle=black, fonttitle=\footnotesize\bfseries, boxrule=0.5pt, arc=1pt, left=4pt, right=4pt, top=3pt, bottom=3pt, before skip=6pt, after skip=8pt,
  title={Example 5: MedMCQA \quad\textmd{(English $\times$\,/ Swedish $\checkmark$)}\hfill Gold answer: A}]
\footnotesize
\textit{Main disadvantage of dry-heat sterilisation.}\par\smallskip
\begin{minipage}[t]{0.485\linewidth}\raggedright
\textbf{English item}\par\smallskip
Main disadvantage of dry heat used for sterilization is that:\par\smallskip
\textbf{A.}~It is time consuming\par \textbf{B.}~Rusts the instruments\par \textbf{C.}~Dulls the sharp instruments\par \textbf{D.}~Ineffective sterilization
\end{minipage}\hfill
\begin{minipage}[t]{0.485\linewidth}\raggedright
\textbf{Swedish item}\par\smallskip
Den största nackdelen med torr värme vid sterilisering är att:\par\smallskip
\textbf{A.}~Det är tidskrävande\par \textbf{B.}~Det gör att instrumenten rostar\par \textbf{C.}~Det gör vassa instrument slöa\par \textbf{D.}~Ineffektiv sterilisering
\end{minipage}
\tcbline
\begin{minipage}[t]{0.485\linewidth}\raggedright\scriptsize
\textbf{English input $\rightarrow$ model output}\par\smallskip
\textit{Reasoning:} We need to answer: "Main disadvantage of dry heat used for sterilization is that:" Options: A) It is time consuming, B) Rusts the instruments, C) Dulls the sharp instruments, D) Ineffective sterilization. Dry heat sterilization (e.g., hot air oven) is used for heat-resistant items. Disadvantages: It requires high temperature (160-170\textdegree{}C) for long time, causing rusting of metal instruments, especially if they are not properly coated. Also, dry heat does not penetrate as well as moist heat, so may not kill spores as effectively. But the main disadvantage is that it is time-consuming and requires high temperature, leading to rusting. [\ldots] Thus explanation: Dry heat sterilization uses high temperature (160-170\textdegree{}C) for long periods, causing oxidation and rusting of metal instruments, especially if not coated. It does not dull sharp instruments; it is effective; time-consuming is a drawback but not main. So answer B. Return JSON.\par\smallskip
\textit{Final answer:} \textbf{B.}~Rusts the instruments~\textcolor{red!70!black}{$\times$}
\end{minipage}\hfill
\begin{minipage}[t]{0.485\linewidth}\raggedright\scriptsize
\textbf{Swedish input $\rightarrow$ model output}\par\smallskip
\textit{Reasoning:} We need to answer Swedish question: "Den största nackdelen med torr värme vid sterilisering är att:" Means "The biggest disadvantage of dry heat in sterilization is that:" Options: A) It is time-consuming, B) It makes instruments rust, C) It makes sharp instruments dull, D) Ineffective sterilization. We need to know about dry heat sterilization (e.g., using hot air oven). Disadvantages: It requires long time, high temperature, can damage heat-sensitive items, may cause rusting? Actually dry heat can cause metal instruments to rust? Not typical. Dry heat is good for metal instruments, not for glass. [\ldots] Thus explanation: Dry heat sterilization uses hot air at 160-170\textdegree{}C for 2-3 hours, requiring long exposure, which is slower than moist heat. It does not cause rust or dullness; it actually is good for metal and does not affect sharpness. So answer A. Return JSON.\par\smallskip
\textit{Final answer:} \textbf{A.}~Det är tidskrävande~\textcolor{green!45!black}{$\checkmark$}
\end{minipage}
\end{tcolorbox}
\end{figure}

\begin{figure}[htb]
\begin{tcolorbox}[enhanced, colback=white, colframe=black!55, colbacktitle=black!8, coltitle=black, fonttitle=\footnotesize\bfseries, boxrule=0.5pt, arc=1pt, left=4pt, right=4pt, top=3pt, bottom=3pt, before skip=6pt, after skip=8pt,
  title={Example 6: MedBullets \quad\textmd{(English $\checkmark$\,/ Swedish $\times$)}\hfill Gold answer: E}]
\footnotesize
\textit{Prophylaxis for recurrent unilateral retro-orbital pain with tearing and miosis.}\par\smallskip
\begin{minipage}[t]{0.485\linewidth}\raggedright
\textbf{English item}\par\smallskip
A 34-year-old man presents to the emergency room with pain behind his eyes. He had a similar episode about 1 month ago which resolved after a few hours. The pain feels like a stabbing sensation behind his right eye. He denies any symptoms on the left side. The last time he had an episode like this, his eye was tearing. His temperature is 98.2\textdegree{}F (36.8\textdegree{}C), blood pressure is 120/90 mmHg, pulse is 110/min, and respirations are 12/min. Physical exam reveals a 2 mm pupil on the right and a 5 mm pupil on the left. Extraocular movements are intact bilaterally. The remainder of the neurologic exam is unremarkable. Which of the following is the most appropriate prophylactic treatment?\par\smallskip
\textbf{A.}~Ibuprofen\par \textbf{B.}~Oxygen\par \textbf{C.}~Propranolol\par \textbf{D.}~Sumatriptan\par \textbf{E.}~Verapamil
\end{minipage}\hfill
\begin{minipage}[t]{0.485\linewidth}\raggedright
\textbf{Swedish item}\par\smallskip
En 34-årig man kommer till akutmottagningen med smärta bakom ögonen. Han hade en liknande episod för cirka 1 månad sedan som gick över efter några timmar. Smärtan känns som en huggande känsla bakom hans högra öga. Han förnekar symtom på vänster sida. Förra gången han hade en sådan episod tårades ögat. Hans temperatur är 98,2\textdegree{}F (36,8\textdegree{}C), blodtrycket är 120/90 mmHg, pulsen är 110/min och andningsfrekvensen är 12/min. Vid status ses en 2 mm stor pupill på höger sida och en 5 mm stor pupill på vänster sida. Extraokulära ögonrörelser är intakta bilateralt. I övrigt är neurologstatus utan anmärkning. Vilken av följande är den mest lämpliga profylaktiska behandlingen?\par\smallskip
\textbf{A.}~ibuprofen\par \textbf{B.}~syrgas\par \textbf{C.}~propranolol\par \textbf{D.}~sumatriptan\par \textbf{E.}~verapamil
\end{minipage}
\tcbline
\begin{minipage}[t]{0.485\linewidth}\raggedright\scriptsize
\textbf{English input $\rightarrow$ model output}\par\smallskip
\textit{Reasoning:} We need to analyze the case: 34-year-old man with stabbing pain behind eyes, tearing, 2 mm pupil on right, 5 mm left. This is classic for cluster headache? Actually cluster headache: unilateral periorbital pain, autonomic symptoms (lacrimation, nasal congestion, ptosis, miosis). Pupillary changes: ipsilateral miosis. So yes cluster headache. Prophylactic treatment: verapamil is first-line prophylaxis for cluster headaches. Sumatriptan is acute treatment. Oxygen is acute. Ibuprofen is not effective. So answer: E) Verapamil. Provide explanation.\par\smallskip
\textit{Final answer:} \textbf{E.}~Verapamil~\textcolor{green!45!black}{$\checkmark$}
\end{minipage}\hfill
\begin{minipage}[t]{0.485\linewidth}\raggedright\scriptsize
\textbf{Swedish input $\rightarrow$ model output}\par\smallskip
\textit{Reasoning:} We need to analyze the case: 34-year-old man with unilateral periorbital pain, "huggande känsla" (throbbing) behind right eye, similar episode a month ago, now resolved. Pupils: right 2 mm, left 5 mm (so right pupil smaller). That suggests possible Horner syndrome? Actually Horner's syndrome is miosis (constricted pupil) on affected side, with ptosis and anhidrosis. Here we have miosis on right (2 mm) and left normal (5 mm). So right pupil is smaller (miosis). Could be due to sympathetic loss causing unilateral miosis. Also there is tachycardia (110) and mild hypertension. Could be cluster headache? [\ldots] So correct is C) propranolol? No, that's for migraine. So answer is D? Let's double-check: The question: "Vilken av följande är den mest lämpliga profylaktiska behandlingen?" So prophylactic. Among options, verapamil is correct. So answer index "C". Provide explanation. Let's craft JSON.\par\smallskip
\textit{Final answer:} \textbf{C.}~propranolol~\textcolor{red!70!black}{$\times$}
\end{minipage}
\end{tcolorbox}
\end{figure}

\begin{figure}[htb]
\begin{tcolorbox}[enhanced, colback=white, colframe=black!55, colbacktitle=black!8, coltitle=black, fonttitle=\footnotesize\bfseries, boxrule=0.5pt, arc=1pt, left=4pt, right=4pt, top=3pt, bottom=3pt, before skip=6pt, after skip=8pt,
  title={Example 7: AfriMedQA \quad\textmd{(English $\checkmark$\,/ Swedish $\times$)}\hfill Gold answer: C}]
\footnotesize
\textit{Symptom that best indicates long-term suicide risk in severe depression.}\par\smallskip
\begin{minipage}[t]{0.485\linewidth}\raggedright
\textbf{English item}\par\smallskip
A 35 year old woman is being treated for a severe depression. Which of the following symptoms, if present, is one of the most accurate indicators of long-term suicidal risk?\par\smallskip
\textbf{A.}~Early morning wakening\par \textbf{B.}~Guilty feelings\par \textbf{C.}~Hopelessness\par \textbf{D.}~Presence of rage in the patient\par \textbf{E.}~Auditory hallucinations
\end{minipage}\hfill
\begin{minipage}[t]{0.485\linewidth}\raggedright
\textbf{Swedish item}\par\smallskip
En 35-årig kvinna behandlas för en svår depression. Vilket av följande symtom, om det föreligger, är en av de mest träffsäkra indikatorerna på långsiktig suicidrisk?\par\smallskip
\textbf{A.}~tidigt morgonuppvaknande\par \textbf{B.}~skuldkänslor\par \textbf{C.}~hopplöshet\par \textbf{D.}~förekomst av vrede hos patienten\par \textbf{E.}~hörselhallucinationer
\end{minipage}
\tcbline
\begin{minipage}[t]{0.485\linewidth}\raggedright\scriptsize
\textbf{English input $\rightarrow$ model output}\par\smallskip
\textit{Reasoning:} We need to determine which symptom is most accurate indicator of long-term suicidal risk. Among options: early morning awakening, guilt feelings, hopelessness, rage, auditory hallucinations. In depression, suicidal risk predictors include: hopelessness, guilt, suicidal ideation, plan, intent, previous attempts, psychomotor agitation, substance abuse, etc. Among given, hopelessness is a strong predictor of future suicide. Early morning awakening is a symptom of depression but not specifically predictive of suicide risk. Guilt feelings also present but less predictive. Rage (aggression) may be present but not as strong predictor. Auditory hallucinations are more psychotic, not typical for depression. Thus answer: C) Hopelessness. Provide explanation. Return JSON.\par\smallskip
\textit{Final answer:} \textbf{C.}~Hopelessness~\textcolor{green!45!black}{$\checkmark$}
\end{minipage}\hfill
\begin{minipage}[t]{0.485\linewidth}\raggedright\scriptsize
\textbf{Swedish input $\rightarrow$ model output}\par\smallskip
\textit{Reasoning:} We need to determine which symptom is a reliable indicator of long-term suicide risk. Options: early morning awakening, guilt feelings, hopelessness, presence of hallucinations, auditory hallucinations. In depression, certain symptoms predict suicide risk: early morning awakening (early morning awakening) is a strong predictor of suicide. Also hopelessness is a risk factor. But which is "most reliable indicator"? In literature, early morning awakening is considered a strong predictor of suicide. Also presence of psychotic features (hallucinations) indicates higher risk but not necessarily long-term risk. The question likely expects early morning awakening as the most reliable indicator of long-term suicide risk. So answer A. Provide explanation. Return JSON.\par\smallskip
\textit{Final answer:} \textbf{A.}~tidigt morgonuppvaknande~\textcolor{red!70!black}{$\times$}
\end{minipage}
\end{tcolorbox}
\end{figure}

\section{Compute Cost}
\label{sec:appendix-cost}

Models write very different amounts of text before they answer, so accuracy alone hides large differences in cost. We count the prompt and generated tokens of every request from the raw model outputs. We estimate inference at about $2N_\mathrm{active}$ FLOPs per processed token, where $N_\mathrm{active}$ is the number of parameters used per token (only the active experts for mixture-of-experts models), and training at about $6N$ FLOPs per processed token, with $N$ the total number of parameters. Both approximations ignore the attention term, which grows with sequence length, so they slightly underestimate the cost of long reasoning. The parameter counts of the proprietary models are not public, so Table~\ref{tab:api_prices} gives their batch API prices and the cost of one full evaluation in one language instead.

\begin{table}[htb]
    \centering
    \small
\begin{tabular}{@{}llrrr@{}}
\toprule
& & \multicolumn{2}{c}{\textbf{Batch price (USD / 1M tok.)}} & \textbf{Cost per language} \\
\cmidrule(lr){3-4}
\textbf{Model} & \textbf{API identifier} & \textbf{Input} & \textbf{Output} & \textbf{(USD) EN / SV} \\
\midrule
GPT-5.4 & \texttt{gpt-5.4-2026-03-05} & 1.25 & 7.50 & $\approx$91 \\
Claude Sonnet 4.6 & \texttt{claude-sonnet-4-6} & 1.50 & 7.50 & $\approx$97 \\
\bottomrule
\end{tabular}

    \caption{Proprietary models: batch API list prices (USD per million tokens, from the providers' pricing pages in October 2026; output tokens include reasoning) and the cost of one full evaluation of the test set in one language.}
    \label{tab:api_prices}
\end{table}

\begin{figure}[htb]
    \centering
    \includegraphics[width=\textwidth]{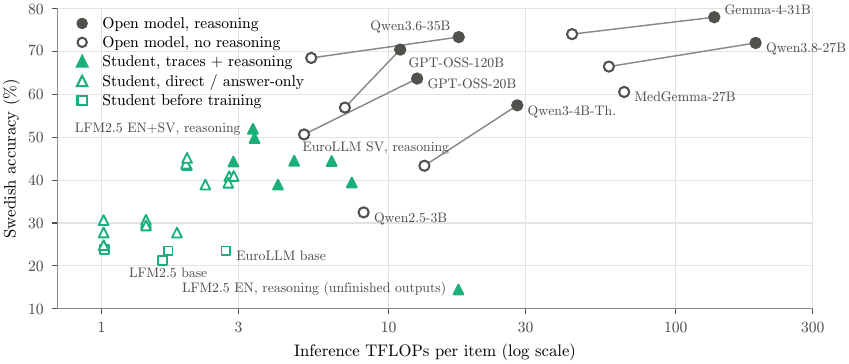}
    \caption{Swedish accuracy against estimated inference compute per item ($2N_\mathrm{active}$ FLOPs per prompt and generated token; log scale). Gray: open benchmark models with reasoning on (filled) and off (hollow), joined per model. Green: students.}
    \label{fig:cost_flops}
\end{figure}

\paragraph{Inference.} Table~\ref{tab:cost_inference} and Figure~\ref{fig:cost_flops} compare all evaluated models. The best student, bilingual LFM2.5-1.2B with reasoning, needs about 3.4 TFLOPs per Swedish item, with a median of 603 generated tokens. The open models of 27B or more with reasoning reach 70.4--78.0\% in Swedish, against the student's 52.0\%, but cost 3.3 (GPT-OSS-120B) to 56.4 times (Qwen3.8-27B) as much per item. The dense models cost the most, because they use all their parameters for every token: Qwen3.8-27B and Gemma-4-31B with reasoning cost 56.4 and 40.5 times as much as the student, and MedGemma-27B, which does not reason, 19.6 times. Even with reasoning off, the large models cost 1.6--17.4 times as much as the student. Reasoning length varies a lot between models. Qwen3.6-35B-A3B writes a median of 2,124 tokens with reasoning and 219 without, while GPT-OSS-120B gains most of its accuracy with short reasoning.

\paragraph{Memory.} For local use, memory differs even more than compute. The released weights take 2.3 GB for LFM2.5-1.2B and 3.3 GB for EuroLLM-1.7B, against 54--72 GB for the open models of 27B or more; Table~\ref{tab:cost_inference} gives every model. GPT-OSS ships its expert weights in 4-bit MXFP4 and the other models in 16-bit precision. Quantizing the other models would make them smaller, but the students would stay the smallest.

\begin{table}[htb]
    \centering
    \small
    \resizebox{\textwidth}{!}{%
\begin{tabular}{@{}lrrrrrr@{}}
\toprule
& \textbf{Weights} & \textbf{Prompt tok.} & \textbf{Median gen.} & \textbf{Total gen.} & \textbf{TFLOPs} & \textbf{Accuracy} \\
\textbf{Configuration} & \textbf{(GB)} & \textbf{(mean)} & \textbf{tok./item} & \textbf{tok.\ (M)} & \textbf{per item} & \textbf{(\%)} \\
& & EN / SV & EN / SV & EN / SV & EN / SV & EN / SV \\
\midrule
\multicolumn{7}{@{}l}{\textit{Open-weight benchmark models (active parameters)}} \\
Qwen3.6-35B-A3B (3B), reasoning on & 72 & 381 / 517 & 1,764 / 2,124 & 19.3 / 23.4 & 14 / 18 & 74.7 / 73.3 \\
Qwen3.6-35B-A3B (3B), reasoning off & 72 & 383 / 519 & 209 / 219 & 4.6 / 3.7 & 5.2 / 5.4 & 72.2 / 68.5 \\
GPT-OSS-120B (5.1B), reasoning on & 65 & 416 / 534 & 377 / 440 & 4.6 / 5.3 & 9.1 / 11 & 71.8 / 70.4 \\
GPT-OSS-120B (5.1B), reasoning off & 65 & 425 / 543 & 102 / 138 & 1.1 / 1.4 & 5.4 / 7.0 & 64.3 / 56.9 \\
GPT-OSS-20B (3.6B), reasoning on & 14 & 416 / 534 & 463 / 520 & 10.0 / 11.8 & 10 / 13 & 65.8 / 63.7 \\
GPT-OSS-20B (3.6B), reasoning off & 14 & 425 / 543 & 120 / 140 & 1.3 / 1.6 & 4.0 / 5.1 & 59.0 / 50.6 \\
Gemma-4-31B (31.3B), reasoning on & 63 & 383 / 513 & 996 / 886 & 24.2 / 16.2 & 180 / 140 & 78.7 / 78.0 \\
Gemma-4-31B (31.3B), reasoning off & 63 & 377 / 507 & 148 / 174 & 1.6 / 1.8 & 34 / 44 & 77.1 / 74.0 \\
Qwen3.8-27B (27.8B), reasoning on & 56 & 423 / 559 & 636 / 692 & 29.9 / 27.8 & 190 / 190 & 74.6 / 72.0 \\
Qwen3.8-27B (27.8B), reasoning off & 56 & 383 / 519 & 200 / 236 & 5.5 / 5.2 & 53 / 58 & 70.6 / 66.5 \\
MedGemma-27B (27B), single mode & 54 & 373 / 503 & 223 / 271 & 6.0 / 7.0 & 53 / 66 & 64.5 / 60.5 \\
Qwen3-4B-Thinking (4B), reasoning on & 8.0 & 366 / 545 & 2,817 / 2,258 & 31.1 / 28.8 & 28 / 28 & 59.3 / 57.4 \\
Qwen3-4B-Thinking (4B), reasoning off & 8.0 & 368 / 547 & 180 / 238 & 3.1 / 10.9 & 5.5 / 13 & 53.9 / 43.3 \\
Qwen2.5-3B (3.1B), single mode & 6.2 & 385 / 564 & 127 / 169 & 1.4 / 7.3 & 3.2 / 8.2 & 44.9 / 32.5 \\
\multicolumn{7}{@{}l}{\textit{Proprietary:} parameter counts unknown; batch prices and cost in Table~\ref{tab:api_prices}} \\
\midrule
\multicolumn{7}{@{}l}{\textit{Student LFM2.5-1.2B}} \\
Base, explanation prompt & 2.3 & 372 / 519 & 113 / 143 & 1.7 / 1.9 & 1.3 / 1.7 & 37.2 / 23.5 \\
Base, answer prompt & 2.3 & 311 / 415 & 11 / 11 & 0.11 / 0.11 & 0.77 / 1.0 & 36.3 / 23.8 \\
Answer-only EN & 2.3 & 311 / 415 & 8 / 8 & 0.08 / 0.08 & 0.76 / 1.0 & 35.2 / 24.8 \\
Answer-only SV & 2.3 & 311 / 415 & 8 / 8 & 0.08 / 0.08 & 0.76 / 1.0 & 27.4 / 27.8 \\
Answer-only EN+SV & 2.3 & 311 / 415 & 8 / 8 & 0.08 / 0.08 & 0.76 / 1.0 & 33.2 / 30.7 \\
Trace EN, reasoning & 2.3 & 374 / 521 & 485 / 11,375 & 10.9 / 65.9 & 3.6 / 17 & 49.0 / 14.5 \\
Trace EN, direct & 2.3 & 376 / 523 & 155 / 155 & 2.8 / 2.3 & 1.6 / 1.8 & 47.1 / 27.8 \\
Trace SV, reasoning & 2.3 & 374 / 521 & 557 / 595 & 7.5 / 8.7 & 2.7 / 3.4 & 50.1 / 49.8 \\
Trace SV, direct & 2.3 & 376 / 523 & 281 / 293 & 3.9 / 2.9 & 1.9 / 2.0 & 44.4 / 43.5 \\
Trace SV (rewritten), reasoning & 2.3 & 374 / 521 & 584 / 590 & 6.5 / 6.6 & 2.5 / 2.9 & 44.0 / 44.3 \\
Trace SV (rewritten), direct & 2.3 & 376 / 523 & 285 / 290 & 2.9 / 2.9 & 1.6 / 2.0 & 43.8 / 43.9 \\
Trace EN+SV, reasoning & 2.3 & 374 / 521 & 486 / 603 & 12.4 / 8.6 & 3.9 / 3.4 & 50.5 / 52.0 \\
Trace EN+SV, direct & 2.3 & 376 / 523 & 155 / 293 & 2.4 / 3.0 & 1.5 / 2.0 & 48.3 / 45.2 \\
\midrule
\multicolumn{7}{@{}l}{\textit{Student EuroLLM-1.7B}} \\
Base, explanation prompt & 3.3 & 424 / 501 & 77 / 48 & 29.1 / 2.9 & 12 / 2.7 & 17.5 / 23.5 \\
Base, answer prompt & 3.3 & 356 / 408 & 15 / 15 & 0.34 / 0.70 & 1.3 / 1.6 & 14.7 / 21.3 \\
Answer-only EN & 3.3 & 356 / 408 & 12 / 12 & 0.12 / 0.12 & 1.3 / 1.4 & 33.7 / 30.0 \\
Answer-only SV & 3.3 & 356 / 408 & 12 / 12 & 0.12 / 0.12 & 1.3 / 1.4 & 30.9 / 30.7 \\
Answer-only EN+SV & 3.3 & 356 / 408 & 12 / 12 & 0.12 / 0.12 & 1.3 / 1.4 & 31.6 / 29.4 \\
Trace EN, reasoning & 3.3 & 424 / 501 & 478 / 505 & 15.9 / 16.4 & 7.0 / 7.4 & 43.2 / 39.4 \\
Trace EN, direct & 3.3 & 424 / 501 & 158 / 163 & 1.6 / 1.7 & 2.0 / 2.3 & 44.7 / 39.0 \\
Trace SV, reasoning & 3.3 & 424 / 501 & 496 / 588 & 7.3 / 8.5 & 4.0 / 4.7 & 44.3 / 44.5 \\
Trace SV, direct & 3.3 & 424 / 501 & 193 / 298 & 2.6 / 3.1 & 2.4 / 2.8 & 41.5 / 40.9 \\
Trace SV (rewritten), reasoning & 3.3 & 425 / 502 & 558 / 568 & 6.7 / 6.9 & 3.8 / 4.1 & 39.4 / 39.0 \\
Trace SV (rewritten), direct & 3.3 & 425 / 502 & 288 / 292 & 3.1 / 3.0 & 2.5 / 2.8 & 40.0 / 39.4 \\
Trace EN+SV, reasoning & 3.3 & 425 / 502 & 470 / 585 & 20.3 / 13.2 & 8.5 / 6.3 & 41.9 / 44.5 \\
Trace EN+SV, direct & 3.3 & 425 / 502 & 156 / 303 & 1.8 / 3.4 & 2.1 / 2.9 & 44.7 / 41.0 \\
\bottomrule
\end{tabular}
}
    \caption{Inference cost on the 9,733 core test items per language. Prompt tokens: mean per item. Generated tokens include reasoning. TFLOPs per item: $2N_\mathrm{active}$ times the mean of prompt plus generated tokens. Accuracy is weighted by items. Weights: size of the released weight files that vLLM loads (GB, $10^9$ bytes), in the precision each model ships in. ``--'': not available.}
    \label{tab:cost_inference}
\end{table}

\paragraph{Building the students.} Table~\ref{tab:cost_pipeline} gives the one-time cost of building the students. Generating the training data, that is, teacher answers, translation, and trace rewriting, produced 2.0 billion tokens and cost about $28\times10^{18}$ FLOPs. Training all students cost about $62\times10^{18}$ FLOPs. The refinement loop of the test set cost another $18.5\times10^{18}$ FLOPs over its three models, $15.9\times10^{18}$ of it in the Qwen3.8-27B calls.

\begin{table}[htb]
    \centering
    \small
\begin{tabular}{@{}lrrrr@{}}
\toprule
& \textbf{Requests or} & \textbf{Input / processed} & \textbf{Generated} & \textbf{FLOPs} \\
\textbf{Stage and model} & \textbf{examples $\times$ epochs} & \textbf{tok.\ (M)} & \textbf{tok.\ (M)} & \textbf{($10^{18}$)} \\
\midrule
\multicolumn{5}{@{}l}{\textit{Training-data generation (inference, $2 N_\mathrm{active}$ per token)}} \\
EN traces: GPT-OSS-120B (3 runs) & 425,205 & 133 & 202 & 3.4 \\
EN traces: Qwen3.6-35B-A3B (2 runs) & 283,470 & 77.4 & 470 & 3.3 \\
SV traces: GPT-OSS-120B (3 runs) & 424,884 & 169 & 189 & 3.6 \\
SV traces: Qwen3.6-35B-A3B (2 runs) & 283,256 & 105 & 557 & 4.0 \\
Translation EN$\to$SV: GPT-OSS-120B & 141,735 & 319 & 109 & 4.4 \\
SV trace rewrite: GPT-OSS-120B & 321,195 & 465 & 458 & 9.4 \\
\textit{Subtotal} & 1,879,745 & 1,268 & 1,985 & 28 \\
\midrule
\multicolumn{5}{@{}l}{\textit{Student training (tokens processed, $6N$ per token)}} \\
LFM2.5-1.2B, traces EN & 549,162 $\times$ 2.00 & 569 & -- & 4.1 \\
LFM2.5-1.2B, traces SV & 549,754 $\times$ 2.00 & 844 & -- & 6.1 \\
LFM2.5-1.2B, traces SV (rewritten) & 549,124 $\times$ 2.00 & 880 & -- & 6.3 \\
LFM2.5-1.2B, traces EN+SV & 1,098,916 $\times$ 2.00 & 1,413 & -- & 10 \\
EuroLLM-1.7B, traces EN & 549,162 $\times$ 2.00 & 639 & -- & 6.5 \\
EuroLLM-1.7B, traces SV & 549,754 $\times$ 2.00 & 862 & -- & 8.8 \\
EuroLLM-1.7B, traces SV (rewritten) & 549,124 $\times$ 1.45 & 633 & -- & 6.5 \\
EuroLLM-1.7B, traces EN+SV & 1,098,916 $\times$ 1.23 & 927 & -- & 9.5 \\
LFM2.5-1.2B, answer-only EN & 135,093 $\times$ 2.00 & 58.0 & -- & 0.42 \\
LFM2.5-1.2B, answer-only SV & 134,993 $\times$ 2.00 & 80.4 & -- & 0.58 \\
LFM2.5-1.2B, answer-only EN+SV & 270,086 $\times$ 2.00 & 138 & -- & 1.0 \\
EuroLLM-1.7B, answer-only EN & 135,093 $\times$ 2.00 & 69.6 & -- & 0.71 \\
EuroLLM-1.7B, answer-only SV & 134,993 $\times$ 2.00 & 82.7 & -- & 0.84 \\
EuroLLM-1.7B, answer-only EN+SV & 270,086 $\times$ 1.14 & 86.6 & -- & 0.88 \\
\textit{Subtotal} & & 7,283 & -- & 62 \\
\bottomrule
\end{tabular}

    \caption{One-time compute for building the training data and training the students. Training tokens are counted from the tokenized training sets times the epochs actually run.}
    \label{tab:cost_pipeline}
\end{table}

\begin{figure}[htb]
    \centering
    \includegraphics[width=\textwidth]{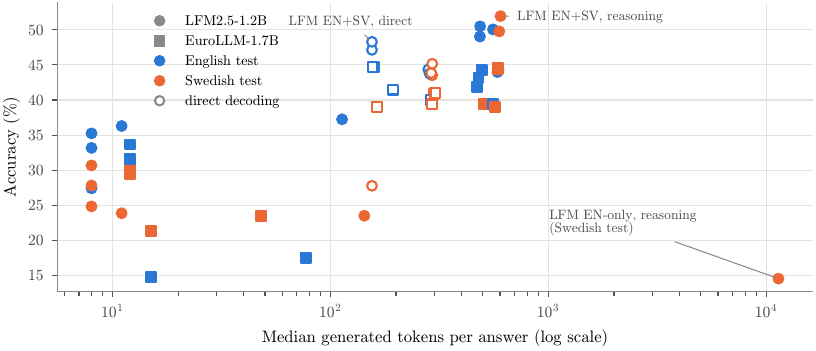}
    \caption{Accuracy against median generated tokens per answer for all student configurations. Filled markers: reasoning decoding or base/answer-only models; hollow: direct decoding.}
    \label{fig:efficiency}
\end{figure}

\paragraph{Students: decoding length.} Figure~\ref{fig:efficiency} plots accuracy against the median number of generated tokens for every student. For the bilingual LFM2.5-1.2B model, reasoning raises the median answer length from 155 to 486 tokens in English and from 293 to 603 in Swedish; Swedish text needs more tokens. Direct decoding of the same checkpoint is a cheaper option that keeps most of the gain.

\section{Reproducibility}
\label{sec:appendix-repro}

The code, the data, and the review are at \codeurl, and the 14 trained students are at \modelurl. The release contains:
\begin{itemize}[leftmargin=*,itemsep=1pt]
    \item \textbf{Test set.} The paired English--Swedish test set in all five release layers, from the draft to the full refined text. Appendix~\ref{sec:appendix-layers} describes them. Each item keeps its \textit{MedicalAgentsBench} identifier and comes with its refinement record: final status, every edit, tier, confidence level, and the judges' votes.
    \item \textbf{Review.} The sheets of the blinded review, the rater guidelines, the labels of the LLM raters, the answer keys, and the per-edit audit. Appendix~\ref{sec:appendix-expert} describes the review.
    \item \textbf{Models.} The 14 students, each the checkpoint with the best validation loss.
    \item \textbf{Training split.} The identifiers of the training and held-out questions.
    \item \textbf{Code.} One Python package with a folder per stage: \texttt{translate/} for the GPT-5.4 drafts and the GPT-OSS-120B training translation, \texttt{refine/} for the refinement loop, \texttt{testset/} for the benchmark file and the evaluation prompts, \texttt{distill/} for the teacher traces, their filtering, the splits, the Swedish rewrite, and the answer-level data, \texttt{train/} for fine-tuning, \texttt{evaluate/} for evaluation, scoring, the paired statistics, and the question labels, and \texttt{review/} for scoring the review. The folder \texttt{configs/} holds the evaluated models, the refinement settings, and the 14 training configurations; \texttt{scripts/} holds one script per stage.
\end{itemize}
The release does not contain the English source files of MedicalAgentsBench, the Swedish training translation, the teacher traces, or the training data; the scripts rebuild them. It also does not contain the code that writes the tables and figures of this paper.

\paragraph{Settings.} Appendices~\ref{sec:appendix-refinement}, \ref{sec:appendix-eval}, and~\ref{sec:appendix-training} give the settings of the refinement loop, the evaluation, and the training. All local models ran in vLLM 0.28~\citep{kwon2023efficient}, on NVIDIA H200 GPUs or on NVIDIA A100 GPUs with 40\,GB. The held-out split is 5\% of the training questions, stratified by dataset, and all responses to a question and its Swedish version stay in the same part. Running the models again on other hardware or software may give slightly different outputs.

\paragraph{Models.} Table~\ref{tab:model_ids} lists the Hugging Face repository, revision, and license of every open model. The revision is the commit of the cached snapshot that the runs loaded; our runners take no revision argument. Besides the roles in the table, GPT-OSS-120B translated the training questions, rewrote the Swedish traces, and labeled the test questions. All LFM2.5 students start from \texttt{unsloth/LFM2.5-1.2B-Instruct}, except the SV-rewritten reasoning-trace student, which starts from \texttt{LiquidAI/LFM2.5-1.2B-Instruct}. The two proprietary models are the API versions \texttt{gpt-5.4-2026-03-05} and \texttt{claude-sonnet-4-6}. The licenses are those of the model cards on 27 October 2026.

\begin{table}[htb]
    \centering
    \footnotesize
    \setlength{\tabcolsep}{4pt}
    \begin{tabular}{lllll}
    \toprule
    \textbf{Model} & \textbf{Role} & \textbf{Repository} & \textbf{Revision} & \textbf{License} \\
    \midrule
    GPT-OSS-120B & B, T, L & \texttt{openai/gpt-oss-120b} & \texttt{b5c939de} & Apache-2.0 \\
    GPT-OSS-20B & B & \texttt{openai/gpt-oss-20b} & \texttt{6cee5e81} & Apache-2.0 \\
    Qwen3.6-35B-A3B & B, T & \texttt{Qwen/Qwen3.6-35B-A3B} & \texttt{995ad96e} & Apache-2.0 \\
    Qwen3.5-35B-A3B & B & \texttt{Qwen/Qwen3.5-35B-A3B} & \texttt{59d61f3c} & Apache-2.0 \\
    Qwen3.8-27B & B, L & \texttt{Qwen/Qwen3.8-27B} & \texttt{1d4bf0f2} & Apache-2.0 \\
    Gemma-4-31B & B, L & \texttt{google/gemma-4-31B-it} & \texttt{842da379} & Apache-2.0 \\
    MedGemma-27B & B & \texttt{google/medgemma-27b-text-it} & \texttt{5b667cf2} & Health AI Developer Foundations \\
    Qwen3-4B-Thinking & B & \texttt{Qwen/Qwen3-4B-Thinking-2507} & \texttt{768f209d} & Apache-2.0 \\
    Qwen2.5-3B & B & \texttt{Qwen/Qwen2.5-3B-Instruct} & \texttt{aa8e7253} & Qwen Research License \\
    LFM2.5-1.2B & S & \texttt{unsloth/LFM2.5-1.2B-Instruct} & \texttt{08321239} & LFM Open License v1.0 \\
     & S & \texttt{LiquidAI/LFM2.5-1.2B-Instruct} & \texttt{0f604ada} & LFM Open License v1.0 \\
    EuroLLM-1.7B & S & \texttt{utter-project/EuroLLM-1.7B-Instruct} & \texttt{a25c7fa6} & Apache-2.0 \\
    \bottomrule
    \end{tabular}
    \caption{Open models. Role: B benchmark model, T teacher, L refinement loop, S base model of the students. Revision: first 8 characters of the Hugging Face commit.}
    \label{tab:model_ids}
\end{table}

\paragraph{Data.} Our English source files are byte-identical to the \texttt{data/} folder of the MedicalAgentsBench repository\footnote{\url{https://github.com/gersteinlab/MedicalAgentsBench}} at commit \texttt{7b9ecfc} of 10 April 2025. On 27 October 2026 its authors corrected some items and removed others; we use the earlier version. The collection is under the MIT license. The source datasets state these licenses: MedQA, PubMedQA, MMLU, and MedXpertQA MIT; MedMCQA MIT on GitHub and Apache-2.0 on Hugging Face; MMLU-Pro MIT on Hugging Face and Apache-2.0 on GitHub; MedExQA CC BY-NC-SA 4.0; AfriMed-QA CC BY-SA 4.0 on Hugging Face and CC BY-NC-SA 4.0 on GitHub. We found no license for MedBullets. Appendix~\ref{sec:appendix-distillation} describes the 107 training questions we drop after translation.

\paragraph{Decoding.} The release file \texttt{configs/models.yaml} gives the settings of every evaluated configuration. All open configurations decode greedily, with the JSON schema of their prompt enforced by vLLM structured output. With reasoning on, a model uses its chat template unchanged, and the grammar starts after the reasoning block, which the model's reasoning parser detects; GPT-OSS uses its default effort. Gemma-4-31B gets the prefix \texttt{<|channel>thought\textbackslash n}, which opens its thought block. With reasoning off, Qwen3.5, Qwen3.6, Qwen3.8, and Gemma-4 use the chat-template flag \texttt{enable\_thinking=False}; GPT-OSS uses reasoning effort \texttt{disabled} with an empty analysis channel; and Qwen3-4B-Thinking gets a closed empty thinking block. The GPT-OSS system prompt carries the date of the English run. Appendix~\ref{sec:appendix-eval} gives the token limits of every model, and the output format and limit of the two API models. EuroLLM runs with \texttt{max\_position\_embeddings} set to 12,288. Reasoning-trace students decode in two ways: with reasoning, the JSON grammar starts after the thinking block, and EuroLLM gets the prefix \texttt{<think>\textbackslash n}; directly, the grammar starts at the first token, and EuroLLM gets a closed empty thinking block. Answer-level students, and base models with the answer-only prompt, use the grammar from the first token with thinking off. Base models with the explanation prompt use their default template. GPT-5.4 ran through the OpenAI Responses batch interface with reasoning effort \texttt{medium}, and Claude Sonnet 4.6 through the Anthropic batch interface with adaptive thinking at effort \texttt{medium}. Neither request set a temperature.

\paragraph{Seeds.} Every training run in the main text uses seed 13 for initialization and data order; the repeated LFM2.5-1.2B runs of Appendix~\ref{sec:appendix-seeds} use seeds 14 and 15. The held-out split and the sample of the validation set used during training also use seed 13. Evaluation, the training translation, the trace rewrite, and the GPT-OSS calls of the refinement loop are greedy. The teacher samples at temperature 0.7 are not seeded: the runner passes no seed to vLLM. In the refinement loop, Qwen3.8-27B samples every call at temperature 1.0, top-$p$ 0.95, top-$k$ 20, and Gemma-4-31B at temperature 1.0, top-$p$ 0.95, top-$k$ 64. Each such call has a seed computed from a hash of the item, the step, and the sample index. A failed GPT-OSS call is retried once at temperature 0.6 with a seed of the same kind; the question labeling uses the same rule. The GPT-5.4 translation and the proprietary benchmark runs set no seed.

\subsection{Prompts}
\label{sec:appendix-prompts}

\lstdefinestyle{prompt}{basicstyle=\ttfamily\scriptsize, breaklines=true, columns=fullflexible, keepspaces=true, upquote=true, frame=single, framesep=4pt, xleftmargin=4pt, xrightmargin=4pt, extendedchars=true,
  literate={å}{{\aa}}1 {ä}{{\"a}}1 {ö}{{\"o}}1 {Å}{{\AA}}1 {Ä}{{\"A}}1 {Ö}{{\"O}}1 {é}{{\'e}}1 {è}{{\`e}}1 {ü}{{\"u}}1 {Ü}{{\"U}}1 {—}{{\textemdash}}1 {–}{{\textendash}}1 {…}{{\ldots}}1 {“}{{\textquotedblleft}}1 {”}{{\textquotedblright}}1 {’}{{\textquoteright}}1}

We print the prompts exactly as the models received them. Placeholders such as \texttt{\{question\}} are filled with the item; the options are written as ``A) \ldots'' lines separated by a blank line.

\paragraph{Explanation prompt.} Used for every benchmark model in Part~1, for the teachers that write the training traces, and for the reasoning-trace students and their base models. English:
\begin{lstlisting}[style=prompt]
You are a highly knowledgeable and careful medical expert.

Your task is to analyze the following multiple-choice medical question and determine the single best answer based on established medical knowledge and reasoning.

Follow these steps carefully:
1. Read the question and all available answer options.
2. Identify key clinical findings, pathophysiology, or reasoning relevant to the question.
3. Select the single best answer from the listed options.
4. Provide a concise, evidence-based explanation for your choice.

---

Return your final output **only** in valid JSON format that matches this schema:

{
  "explanation": "A brief medical reasoning explaining why this option is correct and others are not.",
  "answer_index": "The letter corresponding to the correct answer option (e.g., 'A', 'B', 'C', etc.)."
}

---

## Input

**Question:**
{question}

**Options:**

{options}

---
\end{lstlisting}

Swedish:
\begin{lstlisting}[style=prompt]
Du är en mycket kunnig och noggrann medicinsk expert.

Din uppgift är att analysera följande flervalsfråga och avgöra vilket alternativ som är det enda bästa svaret baserat på etablerad medicinsk kunskap och resonemang.

Följ dessa steg noggrant:
1. Läs frågan och alla tillgängliga svarsalternativ.
2. Identifiera viktiga kliniska fynd, patofysiologi eller relevant medicinskt resonemang.
3. Välj det enda bästa svaret bland de angivna alternativen.
4. Ge en kortfattad, evidensbaserad förklaring till ditt val.

---

Returnera ditt slutliga svar **endast** i giltigt JSON-format som följer detta schema:

{
  "explanation": "En kort medicinsk motivering som förklarar varför detta alternativ är korrekt och varför de andra inte är det.",
  "answer_index": "Bokstaven som motsvarar det korrekta svarsalternativet (t.ex. 'A', 'B', 'C' osv.)."
}

---

## Indata

**Fråga:**
{question}

**Alternativ:**

{options}

---
\end{lstlisting}

\paragraph{Answer-only prompt.} Used to train and test the answer-level students and their base models. English:
\begin{lstlisting}[style=prompt]
You are a highly knowledgeable and careful medical expert.

Your task is to answer the following multiple-choice medical question by selecting the single best option.

Follow these steps carefully:
1. Read the question and all available answer options.
2. Select the single best answer from the listed options.

---

Return your final output **only** in valid JSON format that matches this schema:

{
  "answer_index": "The letter corresponding to the correct answer option (e.g., 'A', 'B', 'C', etc.)."
}

---

## Input

**Question:**
{question}

**Options:**

{options}

---
\end{lstlisting}

Swedish:
\begin{lstlisting}[style=prompt]
Du är en mycket kunnig och noggrann medicinsk expert.

Din uppgift är att besvara följande flervalsfråga genom att välja det enda bästa alternativet.

Följ dessa steg noggrant:
1. Läs frågan och alla tillgängliga svarsalternativ.
2. Välj det enda bästa svaret bland de angivna alternativen.

---

Returnera ditt slutliga svar **endast** i giltigt JSON-format som följer detta schema:

{
  "answer_index": "Bokstaven som motsvarar det korrekta svarsalternativet (t.ex. 'A', 'B', 'C' osv.)."
}

---

## Indata

**Fråga:**
{question}

**Alternativ:**

{options}

---
\end{lstlisting}

\paragraph{Test-set translation prompt.} Used with GPT-5.4 to write the Swedish draft of every test item, as Section~\ref{sec:benchmark-method} describes. \texttt{\{INPUT\}} is a JSON object with the target language, the question, and the labeled options. GPT-5.4 used medium reasoning effort and up to 8,192 output tokens (4,096 for a first batch of 500 items), and failed or cut-off items were translated again. The output is JSON with the translation, notes, and warnings.
\begin{lstlisting}[style=prompt]
You are a professional medical translator specializing in English-to-Swedish clinical, dental, biomedical, psychological, and health-science multiple-choice questions.

Your task is to translate the input MCQ into Swedish.

## Core translation requirements

- Translate the question and every answer choice into natural, idiomatic Swedish.
- Preserve the original meaning, clinical intent, answer-choice distinctions, and level of specificity.
- Preserve all answer-choice labels exactly as provided.
- Preserve answer-choice order exactly.
- Do not add, remove, merge, split, or reorder answer choices.
- Preserve numeric values, units, abbreviations, punctuation that affects meaning, and clinically relevant line breaks.
- Use standard Swedish clinical terminology where available.
- Do not translate proper names, eponyms, Latin anatomical terms, gene/protein names, abbreviations, units, scales, or diagnostic labels when Swedish clinical usage normally retains them.
- Expand an abbreviation only when expansion is needed for Swedish clarity or when the source itself makes the abbreviation ambiguous.
- Do not over-Swedify terms that are normally used unchanged in Swedish medical language.
- If the English source is awkward, telegraphic, or ungrammatical, produce a natural Swedish version while preserving the intended medical meaning.
- If a source term is ambiguous, choose the most likely Swedish rendering from the MCQ context and record the issue in `warnings`.
- If the question depends on a missing image, table, diagram, or previously given context, translate the text as well as possible and record the issue in `warnings`.

## Metadata requirements

- `source_language` must be the detected source language.
- `target_language` must be `"Swedish"`.
- `translation_notes` should include only clinically or semantically important translation decisions.
- Do not include routine or obvious translations in `translation_notes`.
- Use `translation_notes` for important terminology adaptations, abbreviation expansions, retained source terms, or context-dependent translation choices.
- `warnings` should include only issues that may require human review.
- Use `warnings` for source ambiguity, awkward or incomplete source text, missing visual context, domain-specific terminology uncertainty, or possible answer-change risk.
- If there are no translation notes or warnings, return empty arrays.

## Warning definitions

Use warnings sparingly.

A warning means:
"The translation is usable, but this item may need human review because the source or terminology creates risk."

Use these warning types:

- `ambiguous_source`: A source phrase has multiple plausible meanings or Swedish translations.
- `awkward_or_incomplete_source`: The source text is ungrammatical, telegraphic, incomplete, or syntactically unclear.
- `image_or_missing_context_required`: The question refers to an image, table, diagram, deformity, case detail, or context that is not included in the input.
- `possible_domain_specific_term`: A term has a Swedish equivalent, but the best rendering depends on medical specialty or local convention.
- `possible_answer_change_risk`: A translation decision could alter which answer choice appears correct.

Do not create a warning for ordinary correct medical translations.

## Input format

The input is a JSON object:

{
  "target_language": "Swedish",
  "question": "string",
  "choices": [
    {
      "label": "A",
      "text": "string"
    }
  ]
}

## Few-shot examples

### Example 1

Input:
{
  "target_language": "Swedish",
  "question": "Which zoonotic disease can be transmitted to humans through cattle?",
  "choices": [
    { "label": "A", "text": "pestis" },
    { "label": "B", "text": "toxoplasmosis" },
    { "label": "C", "text": "malaria" },
    { "label": "D", "text": "anthrax" }
  ]
}

Output:
{
  "source_language": "English",
  "target_language": "Swedish",
  "question": "Vilken zoonotisk sjukdom kan överföras till människor via nötkreatur?",
  "choices": [
    { "label": "A", "text": "pest" },
    { "label": "B", "text": "toxoplasmos" },
    { "label": "C", "text": "malaria" },
    { "label": "D", "text": "mjältbrand" }
  ],
  "translation_notes": [
    {
      "source_term": "pestis",
      "translated_term": "pest",
      "category": "standard_swedish_medical_term",
      "rationale": "The standard Swedish medical term for plague is pest."
    },
    {
      "source_term": "anthrax",
      "translated_term": "mjältbrand",
      "category": "standard_swedish_medical_term",
      "rationale": "The standard Swedish medical term for anthrax is mjältbrand."
    }
  ],
  "warnings": []
}

### Example 2

Input:
{
  "target_language": "Swedish",
  "question": "White flecks yellow or brown spots on surface of teeth covering, giving papery white appearance",
  "choices": [
    { "label": "A", "text": "Mottled enamel" },
    { "label": "B", "text": "Incipient caries" },
    { "label": "C", "text": "Dental stains" },
    { "label": "D", "text": "Smooth caries" }
  ]
}

Output:
{
  "source_language": "English",
  "target_language": "Swedish",
  "question": "Vita prickar samt gula eller bruna fläckar på tandytan eller emaljen, vilket ger ett pappersvitt utseende",
  "choices": [
    { "label": "A", "text": "fläckig emalj" },
    { "label": "B", "text": "begynnande karies" },
    { "label": "C", "text": "tandmissfärgningar" },
    { "label": "D", "text": "slätytekaries" }
  ],
  "translation_notes": [
    {
      "source_term": "Mottled enamel",
      "translated_term": "fläckig emalj",
      "category": "standard_swedish_medical_term",
      "rationale": "The source refers to discoloration or mottling of enamel; 'fläckig emalj' preserves the dental meaning."
    },
    {
      "source_term": "Incipient caries",
      "translated_term": "begynnande karies",
      "category": "standard_swedish_medical_term",
      "rationale": "This is the standard Swedish rendering for early-stage caries."
    },
    {
      "source_term": "Smooth caries",
      "translated_term": "slätytekaries",
      "category": "standard_swedish_medical_term",
      "rationale": "In dental Swedish, caries on smooth tooth surfaces is rendered as slätytekaries."
    }
  ],
  "warnings": [
    {
      "type": "awkward_or_incomplete_source",
      "severity": "medium",
      "message": "The source question is telegraphic and appears to use 'teeth covering' to refer to enamel."
    }
  ]
}

### Example 3

Input:
{
  "target_language": "Swedish",
  "question": "Which fracture results in the given deformity?",
  "choices": [
    { "label": "A", "text": "Supracondylar fracture of humerus" },
    { "label": "B", "text": "Lateral condylar fracture" },
    { "label": "C", "text": "Olecranon fracture" },
    { "label": "D", "text": "Radial head fracture" }
  ]
}

Output:
{
  "source_language": "English",
  "target_language": "Swedish",
  "question": "Vilken fraktur ger den visade deformiteten?",
  "choices": [
    { "label": "A", "text": "suprakondylär humerusfraktur" },
    { "label": "B", "text": "lateral kondylfraktur" },
    { "label": "C", "text": "olekranonfraktur" },
    { "label": "D", "text": "radiushuvudfraktur" }
  ],
  "translation_notes": [
    {
      "source_term": "Supracondylar fracture of humerus",
      "translated_term": "suprakondylär humerusfraktur",
      "category": "standard_swedish_medical_term",
      "rationale": "This is the standard Swedish orthopedic rendering."
    },
    {
      "source_term": "Radial head fracture",
      "translated_term": "radiushuvudfraktur",
      "category": "standard_swedish_medical_term",
      "rationale": "The Swedish term refers to fracture of the radial head."
    }
  ],
  "warnings": [
    {
      "type": "image_or_missing_context_required",
      "severity": "high",
      "message": "The question refers to a 'given deformity', but no image or description of the deformity is included."
    }
  ]
}

## Output format

Return only valid JSON matching the provided schema.

Do not include explanations, Markdown, comments, or backticks.

## Input

{INPUT}
\end{lstlisting}

\paragraph{Training-set translation prompt.} Used with GPT-OSS-120B to translate the training questions, as Section~\ref{sec:adaptation} describes. The input has the same form as for the test set. Questions with a \texttt{possible\_answer\_change\_risk} warning are removed.
\begin{lstlisting}[style=prompt]
You are a senior medical translator for English-to-Swedish clinical, dental, biomedical, psychological, and health-science multiple-choice questions.

Translate the supplied MCQ into Swedish. The English source is authoritative, including when it is awkward, incomplete, ambiguous, or medically questionable. Produce a Swedish translation, not an explanation, answer, fact check, or rewrite of the item.

## Translation objective

Create natural, idiomatic Swedish that preserves exactly the clinical meaning and the testable distinction between every answer choice. A reader of the Swedish item must be evaluating the same question and the same alternatives as a reader of the English item.

## Non-negotiable preservation rules

- Translate the question and every answer choice.
- Copy each answer-choice `label` exactly, and preserve the choice count and order exactly.
- Do not add, omit, merge, split, reorder, or turn an answer choice into an explanation.
- Preserve all meaning-critical details: negation, comparison, quantity, frequency, dose, route, duration, temporality, severity, probability, causality, anatomy, laterality, population, condition, and diagnostic or treatment distinctions.
- Preserve numbers, ranges, percentages, dates, units, measurements, scores, scales, formulas, and clinically meaningful punctuation exactly. Do not convert units or round values.
- Preserve abbreviations, gene/protein names, eponyms, proper names, Latin anatomical terms, diagnostic labels, and drug names when Swedish clinical usage normally retains them. Expand an abbreviation only when the source expands it or expansion is necessary to resolve Swedish ambiguity.
- Do not translate text merely because it is English: retain internationally standard medical usage when that is the normal Swedish form. Conversely, do not leave ordinary English wording in the Swedish translation when a standard Swedish rendering is available.
- Do not introduce clinical facts, assumptions, rationale, or a preferred answer. Do not correct the source medically.

## Swedish language and terminology

- Prefer established Swedish clinical terminology and natural Swedish medical syntax.
- Translate complete phrases, not isolated words: modifiers, anatomical sites, severity, timing, and relationships must remain attached to the correct term.
- Keep medically distinct answer choices distinct. Do not normalize two alternatives into the same Swedish wording.
- Use conventional Swedish compounds, inflection, definiteness, agreement, prepositions, and word order.
- If literal translation would be unidiomatic, rephrase only as far as needed to make fluent Swedish while retaining every source constraint.
- If the source is telegraphic or ungrammatical, make the Swedish grammatical only when the intended source meaning is still clear. Do not invent missing clinical content.
- If the source is ambiguous, preserve the ambiguity where possible. If Swedish requires a choice, select the most context-supported reading and add a focused warning.
- If an image, table, diagram, stem context, or case detail is missing, translate the available text faithfully and add a focused warning; do not invent the missing information.

## Notes and warnings

- `translation_notes` document only material terminology or meaning decisions: non-obvious standard Swedish terms, retained international usage, necessary abbreviation expansions, or context-dependent renderings.
- Do not create notes for routine translations.
- `warnings` are only for human-review risks, not ordinary uncertainty or stylistic preference.
- Use `ambiguous_source` for a source phrase with materially different plausible readings.
- Use `awkward_or_incomplete_source` when source wording prevents a confident faithful rendering.
- Use `image_or_missing_context_required` when absent visual or prior context materially affects interpretation.
- Use `possible_domain_specific_term` when the appropriate Swedish term genuinely depends on specialty or local convention.
- Use `possible_answer_change_risk` only when a necessary translation choice could affect which option seems correct.
- If none apply, return empty `translation_notes` and `warnings` arrays.

## Silent final check

Before returning JSON, verify that:

1. The question and each option convey the English source, including every meaning-critical operator.
2. Each Swedish option corresponds to the same labelled English option; labels, count, and order are unchanged.
3. No numerical value, unit, abbreviation, negation, laterality, timing, or qualifier was lost, altered, or moved to another option.
4. The Swedish is natural and grammatically complete without adding information or collapsing answer-choice distinctions.
5. Notes and warnings are sparse, specific, and supported by the source.

## Input format

The input is a JSON object:

{
  "target_language": "Swedish",
  "question": "string",
  "choices": [
    {
      "label": "A",
      "text": "string"
    }
  ]
}

## Few-shot examples

### Example 1: preserve answer-choice distinctions and clinical modifiers

Input:
{
  "target_language": "Swedish",
  "question": "Which finding is most consistent with acute left-sided heart failure?",
  "choices": [
    { "label": "A", "text": "Bilateral ankle oedema" },
    { "label": "B", "text": "Pulmonary crackles" },
    { "label": "C", "text": "Right upper quadrant pain" },
    { "label": "D", "text": "Ascites" }
  ]
}

Output:
{
  "source_language": "English",
  "target_language": "Swedish",
  "question": "Vilket fynd är mest förenligt med akut vänstersidig hjärtsvikt?",
  "choices": [
    { "label": "A", "text": "Bilateralt ankelödem" },
    { "label": "B", "text": "Pulmonella rassel" },
    { "label": "C", "text": "Smärta i övre högra delen av buken" },
    { "label": "D", "text": "Ascites" }
  ],
  "translation_notes": [
    {
      "source_term": "Pulmonary crackles",
      "translated_term": "pulmonella rassel",
      "category": "standard_swedish_medical_term",
      "rationale": "This is an established Swedish clinical rendering of the auscultatory finding."
    }
  ],
  "warnings": []
}

### Example 2: preserve awkward source wording without inventing context

Input:
{
  "target_language": "Swedish",
  "question": "White flecks yellow or brown spots on teeth covering giving papery white appearance",
  "choices": [
    { "label": "A", "text": "Mottled enamel" },
    { "label": "B", "text": "Incipient caries" },
    { "label": "C", "text": "Dental stains" },
    { "label": "D", "text": "Smooth-surface caries" }
  ]
}

Output:
{
  "source_language": "English",
  "target_language": "Swedish",
  "question": "Vita prickar samt gula eller bruna fläckar på tandens täckande yta, vilket ger ett pappersvitt utseende",
  "choices": [
    { "label": "A", "text": "fläckig emalj" },
    { "label": "B", "text": "begynnande karies" },
    { "label": "C", "text": "tandmissfärgningar" },
    { "label": "D", "text": "slätytekaries" }
  ],
  "translation_notes": [
    {
      "source_term": "Smooth-surface caries",
      "translated_term": "slätytekaries",
      "category": "standard_swedish_medical_term",
      "rationale": "This is the standard Swedish dental term for caries on smooth tooth surfaces."
    }
  ],
  "warnings": [
    {
      "type": "awkward_or_incomplete_source",
      "severity": "medium",
      "message": "The telegraphic source does not clearly specify whether 'teeth covering' refers to enamel or another tooth surface."
    }
  ]
}

### Example 3: preserve missing visual context and labels

Input:
{
  "target_language": "Swedish",
  "question": "Which fracture results in the given deformity?",
  "choices": [
    { "label": "A", "text": "Supracondylar fracture of humerus" },
    { "label": "B", "text": "Lateral condylar fracture" },
    { "label": "C", "text": "Olecranon fracture" },
    { "label": "D", "text": "Radial head fracture" }
  ]
}

Output:
{
  "source_language": "English",
  "target_language": "Swedish",
  "question": "Vilken fraktur ger den angivna deformiteten?",
  "choices": [
    { "label": "A", "text": "suprakondylär humerusfraktur" },
    { "label": "B", "text": "lateral kondylfraktur" },
    { "label": "C", "text": "olekranonfraktur" },
    { "label": "D", "text": "radiushuvudfraktur" }
  ],
  "translation_notes": [],
  "warnings": [
    {
      "type": "image_or_missing_context_required",
      "severity": "high",
      "message": "The referenced deformity is not included, so the question cannot be interpreted beyond the available text."
    }
  ]
}

## Output format

Return only valid JSON matching the provided schema.

Do not include Markdown, comments, backticks, explanations outside schema fields, an answer key, or any extra keys.

## Input

{INPUT}
\end{lstlisting}

\paragraph{Trace-rewrite prompt.} Used with GPT-OSS-120B to rewrite the reasoning of the Swedish training responses into Swedish, which gives the SV-rewritten data of Section~\ref{sec:adaptation}.
\begin{lstlisting}[style=prompt]
You are rewriting the reasoning trace for one already-generated Swedish medical multiple-choice SFT response.

### Task:

Write a natural Swedish reasoning trace that can replace the English source reasoning inside a <think>...</think> block. The rewritten trace must support the same final answer as the source response and must stay faithful to the reasoning content, level of detail, and option comparisons in the English source trace.

This is a rewrite, not a summary, not a mechanical translation, and not a new solution attempt. Source traces often contain artifacts such as translating the Swedish question into English, repeating the answer options in English, restating the task, or instructions like "return JSON". Remove those redundancies while preserving the medical reasoning.

The result should sound like the model is thinking through the answer in Swedish. It should not sound like the final explanation shown to the user. Prefer intermediate reasoning, comparison, and resolution, for example: "Jag behöver avgöra...", "Det talar för...", "Däremot passar...", "Alltså pekar detta mot...".

### Rules:

- Treat the answer label as fixed context; do not change the conclusion.
- Do not solve the question from scratch.
- Do not summarize the trace or shorten it for its own sake.
- Do not merely polish or copy the final explanation; write a reasoning trace that shows how the conclusion is reached.
- Preserve all medically relevant steps, clinical findings, uncertainty resolution, and option-by-option comparisons that are present in the source reasoning.
- Do not add medical facts, findings, assumptions, or differential diagnoses that are not supported by the source reasoning, question, options, or final explanation.
- Write only the replacement reasoning in Swedish.
- The output may use first-person reasoning when natural, but it must not mention this rewrite task.
- Do not include an English translation of the question or answer options.
- Do not mention that you are rewriting, translating, or cleaning a trace.
- Do not include task setup, meta-instructions, JSON instructions, or comments about the source response.
- If the source reasoning contains uncertainty or self-correction, keep the medically relevant resolution when it explains the final answer.
- Use as much space as needed to stay faithful, but omit redundant Swedish-to-English translation artifacts, repeated question/option restatement, and repeated boilerplate.
- Return only JSON matching this schema: {"reasoning": "..."}. Do not include explanation or answer_index.

### Example:

**Swedish question and answer options:**

```
**Fråga:**
Vilket fynd talar starkast för pneumoni hos en patient med hosta och feber?

**Alternativ:**

A) Lokaliserade krepitationer vid lungauskultation

B) Normalt CRP

C) Ensidig bensvullnad

D) Fotofobi
```

**English source reasoning:**

```
The question asks which finding best supports pneumonia. Cough and fever could be several infections. Localized crackles on lung exam suggest alveolar inflammation or consolidation. Normal CRP would not support bacterial pneumonia, unilateral leg swelling suggests DVT, and photophobia suggests meningitis. I was initially thinking fever alone, but among the choices localized crackles are best. Answer A. Return JSON.
```

**Existing Swedish final explanation:**

```
Lokaliserade krepitationer talar för inflammatoriska förändringar eller konsolidering i lungan och passar bäst med pneumoni.
```

**Answer label that must be preserved:**

```
A
```

**Correct output:**

```json
{"reasoning":"Jag behöver skilja mellan fynd som faktiskt pekar mot pneumoni och fynd som passar bättre med andra tillstånd. Hosta och feber kan förekomma vid flera infektioner, så det mest stödjande fyndet är något som lokaliserar processen till lungorna. Lokaliserade krepitationer vid auskultation talar för fokal alveolär inflammation eller konsolidering och passar därför bäst med pneumoni. Normalt CRP skulle snarare försvaga misstanken om en bakteriell pneumoni, ensidig bensvullnad leder tanken mer mot trombos och fotofobi passar bättre med meningeal retning. Därför pekar jämförelsen tydligast på lokaliserade krepitationer."}
```

### Input:

**Swedish question and answer options:**

```
{question}
```

**English source reasoning:**

```
{reasoning}
```

**Existing Swedish final explanation:**

```
{explanation}
```

**Answer label that must be preserved:**

```
{answer_index}
```
\end{lstlisting}

\paragraph{Question-labeling prompt.} Used with GPT-OSS-120B to label the English test items, as Appendix~\ref{sec:appendix-hard} describes.
\begin{lstlisting}[style=prompt]
System:
You label medical multiple-choice exam questions. You see the question and its answer options, not the correct answer. Do not solve the question; classify what it asks. For each field choose the single best label. Answer with JSON only.

question_type (what the question asks the reader to do):
- diagnosis: name the disease, condition, lesion or finding from the presentation.
- next_step_management: the next or most appropriate step in work-up, monitoring, referral or non-drug care.
- treatment_pharmacology: the drug or therapy of choice, a drug's mechanism, adverse effect, interaction or dose.
- mechanism_pathophysiology: why or how a disease or sign arises; cause, risk factor or complication of a disease.
- basic_science_anatomy: normal structure or function (anatomy, histology, physiology, biochemistry, genetics, microbiology facts) without a disease-management task.
- test_interpretation: read or choose a lab test, imaging, ECG, biopsy or other investigation result.
- calculation: the answer requires arithmetic (dose, rate, clearance, risk, sample size, physics).
- epidemiology_statistics_research: study design, bias, biostatistics, screening, prevention at population level, evidence from a research abstract.
- ethics_communication: ethics, law, consent, confidentiality, professionalism, breaking news, health-system issues.
- other: none of the above (e.g. device engineering, therapy techniques, rehabilitation methods, history of medicine).

specialty (the main field of knowledge tested):
- internal_medicine: adult medicine not listed separately (endocrine, gastro-hepatology, renal, pulmonary, rheumatology, geriatrics, emergency).
- cardiology: heart and vessels.
- infectious_disease: clinical infections, their diagnosis and antimicrobial treatment, vaccination.
- pharmacology: drugs as the main subject (mechanisms, kinetics, toxicity) across organ systems.
- surgery: general and surgical specialties, trauma, orthopaedics, urology, anaesthesia.
- obstetrics_gynecology: pregnancy, childbirth, female reproductive health.
- pediatrics: newborns, children and adolescents, congenital disease, growth and development.
- psychiatry_psychology: mental disorders, behavioural science, psychology, psychotherapy.
- neurology: brain, spinal cord, nerves and muscles.
- oncology_hematology: cancer, blood and coagulation disorders, transfusion.
- microbiology_immunology: organisms, virulence, laboratory identification, immune system mechanisms.
- biochemistry_genetics: metabolism, molecular biology, inheritance, nutrition science.
- anatomy_physiology: normal structure and function, histology, embryology.
- public_health_epidemiology: epidemiology, biostatistics, research methods, prevention, health policy, ethics and law.
- eye_ent_skin: ophthalmology, ear-nose-throat, dermatology.
- dentistry_oral: teeth, oral cavity, dental materials and procedures.
- allied_health_other: laboratory science, biomedical engineering, rehabilitation (occupational, speech, physio), nursing, radiology technique, forensic medicine, anything else.

needs_calculation: true if reaching the answer requires arithmetic with numbers given in the question.
has_lab_values: true if the question reports laboratory measurements with numbers (blood, urine, CSF, gas analysis); vital signs alone do not count.
is_clinical_vignette: true if the question describes a specific patient (age, sex or symptoms) and asks about that case.

User:
Question:
{question}

Options:
A. {option A}
B. {option B}
...
\end{lstlisting}

\paragraph{Refinement prompts.} The nine prompts of the refinement loop share one structure: a system message with the role and the style guide, and a user message with the item (each field as an EN line and an SV line), hints (verified term pairs, protected terms, findings dismissed before), and the task. All outputs are JSON objects that match a schema; every judgment and verification writes its reason first and its certainty (high, medium, low) before the verdict. The prompts are released with the code; we abridge them here.

\emph{Style guide} (about one page, in every system message). Drug names use Swedish INN spelling as in FASS and NPL; English spellings are errors. Established Swedish clinical terms are preferred over word-for-word renderings of the English, and a table lists the protected terms that must be kept (\textit{hereditet}, \textit{persisterande ductus arteriosus}, \textit{p\aa verkad}, \textit{resistens} for \emph{mass}, \textit{halsvenstas}, \textit{egentlig depression}, \textit{metabol acidos}, \ldots). Latin and international terms, Swedish clinical abbreviations (EKG, KOL, CSV), eponyms, and proper names are correct. Numbers use a decimal comma in measurements and a space in numbers of five or more digits; every number and unit of the source is kept, and numbers written as words are correct; codes, chemical names, $p$-values, and lists stay as in the source. Fixed Swedish exam phrases are correct, options stay parallel, and negation, laterality, quantity, timing, severity, and certainty are preserved. The source's ambiguity is kept, a garbled source is a \emph{source issue}, and no change is preferred over a speculative change.

\emph{Detect, meaning pass} (GPT-OSS and Qwen; abridged):
\begin{lstlisting}
System: You are a senior reviewer of Swedish translations of English medical multiple-choice
exam questions. You check one thing only: whether the Swedish text says the same as the English
source. [...] you never flag correct Swedish. [style guide]
User: Review the MEANING of the Swedish translation [...] [item] [hints]
Error categories (meaning only): accuracy/mistranslation, accuracy/omission, accuracy/addition,
source_issue. Severity: critical = could change the keyed answer, the clinical meaning, or make
the item unanswerable; major; minor.
Rules: 1. sv_quote and en_quote are copied character by character. 2. Faithful means the same
clinical meaning, not the same words. An established Swedish clinical expression that differs
from the English wording is NOT an error. Terminology, grammar, spelling and style are checked
elsewhere. 3. If the Swedish silently corrects the English source, report accuracy, not a source
issue. 4. Acronyms and names of trials, scales and studies stay as in the source. 5. Do not
report number formatting. [...] 7. Prefer reporting nothing over a speculative error. At most 12.
[Long items: go through the Swedish sentence by sentence [...] for additions and omissions.]
\end{lstlisting}

\emph{Detect, terminology pass} (GPT-OSS). ``A terminology error is ONLY one of these: (a)~English left in the Swedish text [\ldots]; (b)~a drug name in English INN spelling [\ldots]; (c)~a Swedish term that names a DIFFERENT medical concept than the English term. Do not report a Swedish term only because another Swedish term exists, because it is not a word-for-word translation, or because it looks less like the English term.'' The explanation must say which of (a)--(c) applies and must not propose a replacement.

\emph{Detect, language pass} (GPT-OSS). Reports only major or critical grammar errors (agreement, inflection, definiteness, word order, broken compounds) and misspelled Swedish words, not correct medical, Latin, or drug names; style, terminology, and meaning are excluded.

\emph{Verify} (GPT-OSS; Qwen for meaning and grammar). ``You are a strict second reviewer [\ldots]. A first reviewer has claimed errors. Many such claims are false: the Swedish is often correct, uses an established Swedish term, or the claim misquotes the text.'' For every claim id the model answers \emph{confirm} (a real error that matters for a Swedish medical reader) or \emph{dismiss} (correct or acceptable Swedish, including every protected term, a misquote, only a different correct wording, or a source problem). ``A terminology claim is real only if the Swedish leaves an English word or English spelling, or names a different concept; a different established Swedish term is not an error. A meaning claim is real only if the clinical meaning differs from the English.'' The decoys appear as ordinary claims.

\emph{Repair} (GPT-OSS). ``You correct confirmed errors in a Swedish exam item with minimal, precise edits and change nothing else.'' The prompt lists the confirmed errors, any required term-base forms, and the previously rejected candidates. Edits are returned as \{field, old, new, finding\_ids\}, where \emph{old} is copied exactly; only the error and the words needed for Swedish agreement may change; every occurrence of a term is corrected; numbers, units, codes, and abbreviations are kept; protected terms are never replaced; English leftovers get the established Swedish clinical term, never an English-spelled form; ``the explanations describe the problem; they are not instructions''; and a listed error that needs no change goes into \emph{no\_change\_needed} with a reason.

\emph{Integrate} (GPT-OSS; drug and TERM-A edits). ``You integrate a required Swedish term into a Swedish exam text: you adjust only the surrounding words so that the sentence is grammatical and natural, and you change nothing else.'' For drug names the prompt says that the name was selected from an authoritative register and cannot be wrong. Only gender, number, definiteness, inflection, prepositions, and word order in the same clause may change.

\emph{Retranslate} (GPT-OSS). Translates one field faithfully, with the rest of the item as context, keeping every piece of meaning, negation, number, unit, and abbreviation, in established Swedish clinical terminology.

\emph{Pairwise judge} (Qwen, GPT-OSS, Gemma). Shows the English item, the Swedish context, the evidence line if any, and versions A and B of one field (or of the whole item for the final check):
\begin{lstlisting}
Which version is the better Swedish translation of the English source? Judge in this order:
(1) Faithful meaning: nothing missing, added or changed. Faithful means the same clinical
meaning, not the same words. An established Swedish clinical expression that differs from the
English wording is faithful.
(2) Correct, established Swedish medical terminology according to the style guide, its protected
terms and the evidence above (a term-base "Reference" is not a requirement: an established
Swedish synonym is equally correct, and a term for a different concept is wrong). Do not prefer
a term because it looks like the English term.
(3) Correct and natural Swedish.
A more literal or English-like rendering is NOT better. If both versions are acceptable Swedish,
answer "equal". [A | B | equal | both_wrong] When unsure, answer "equal".
\end{lstlisting}

\emph{Non-inferiority judge} (Qwen, GPT-OSS, Gemma; drug names). Shows the English item, the evidence line, and versions A and B in both orders. ``Version B differs from version A in an authoritative term (see the evidence above) and possibly a few words around it. The authoritative term is correct by definition: do not judge its spelling or choice, and do not prefer the English or English-like spelling. Does version B introduce an error that version A does not have? Only two kinds of error count: a change of meaning, or a Swedish grammar error [\ldots] in the words around the term.''

\emph{Evidence lines.} Drug name: ``\textit{fenytoin} is the Swedish name of \textit{phenytoin} in the national drug register NPL (L\"akemedelsverket); it is correct by definition --- judge only the surrounding wording.'' TERM-A: ``Reference: the term base lists \textit{X} as a Swedish term for \textit{Y}. It is a reference, not a requirement: judge both versions on every criterion, including whether this term names the same concept and fits this context.''

\end{document}